\pdfoutput=1

\documentclass{article}

\usepackage{PRIMEarxiv}

\usepackage[utf8]{inputenc} 
\usepackage[T1]{fontenc}    
\usepackage{hyperref}       
\usepackage{url}            
\usepackage{booktabs}       
\usepackage{amsfonts}       
\usepackage{nicefrac}       
\usepackage{microtype}      
\usepackage{lipsum}
\usepackage{fancyhdr}       
\usepackage{graphicx}       
\graphicspath{{media/}}     

\usepackage{times}
\usepackage{framed} 
\usepackage{amsthm}
\usepackage{amsmath}
\usepackage{amssymb}
\usepackage{algorithm}
\usepackage{algorithmic}
\usepackage{enumerate}
\usepackage{geometry}
\usepackage{subfigure}
\usepackage{float}
\usepackage{fancyhdr}
\usepackage{booktabs}
\usepackage{color,xcolor}
\usepackage{marvosym}

\usepackage{caption}
\usepackage{multirow}
\usepackage[normalem]{ulem}
\useunder{\uline}{\ul}{}

\makeatletter
\DeclareRobustCommand\TGD{\mathop{\operator@font TGD}\nolimits}
\DeclareRobustCommand\GD{\mathop{\operator@font GD}\nolimits}

\makeatother

\title{Applications of Tao General Difference \\in Discrete Domain}

\author{
  Linmi Tao$^{\ast}$\textsuperscript{\Letter}, Ruiyang Liu$^{\ast}$, Donglai Tao, Wu Xia, Feilong Ma, Yu Cheng, Jingmao Cui \\ \\
  Department of Computer Science and Technology, Tsinghua University \\
  Beijing {\rm 100084}, China\\ \\
  \normalsize{$^\ast$Co-first author.}\\
  \normalsize{\textsuperscript{\Letter} Corresponding author. E-mail: linmi@tsinghua.edu.cn} \\
}

\begin{document}
\maketitle

\begin{abstract}

Numerical difference computation is one of the cores and indispensable in the modern digital era. Tao general difference (TGD) is a novel theory and approach to difference computation for discrete sequences and arrays in multidimensional space. Built on the solid theoretical foundation of the general difference in a finite interval, the TGD operators demonstrate exceptional signal processing capabilities in real-world applications. A novel smoothness property of a sequence is defined on the first- and second TGD. This property is used to denoise one-dimensional signals, where the noise is the non-smooth points in the sequence. Meanwhile, the center of the gradient in a finite interval can be accurately location via TGD calculation. This solves a traditional challenge in computer vision, which is the precise localization of image edges with noise robustness.  Furthermore, the power of TGD operators extends to spatio-temporal edge detection in three-dimensional arrays, enabling the identification of kinetic edges in video data. These diverse applications highlight the properties of TGD in discrete domain and the significant promise of TGD for the computation across signal processing, image analysis, and video analytic.

\end{abstract}


\vspace{40pt}

This paper is the application part of the paper "Tao General Differential and Difference: Theory and Application". The theory part of the paper is renamed as "A Theory of General Difference in
Continuous and Discrete Domain", which is Arxived in arXiv:2305.08098v2

\newpage
\tableofcontents
\newpage

\section{Introduction}
In signal processing, sampling is the reduction of a continuous-time signal to a discrete-time signal, in which the inevitable noise overlays complex irregularities and corrupts the signals. Denoising is the process of extracting signals from a combination of signals and noise, thus preserving critical information that is essential in various applications. The smoothing assumption is reasonable for discrete signals\footnote{Mathematically, a discrete signal refers to an ordered sequence of real-valued numbers sampled from a smooth signal, while inevitable noise overlays the signal with complex irregularities that cause discontinuities.} and it enables pattern forecast and provides both flexible and robust data analysis~\cite{simonoff2012smoothing}. In this sense, in the context of one-dimensional data, signal smoothing and signal denoising can be used interchangeably~\cite{chen2019denoising}.

Smooth functions have continuous derivative functions. However, improving the smoothness of denoised signals is a great challenge in the absence of robust noise-resistant methods for calculating discrete signal “derivative”. Fortunately, TGD~\cite{TGD2023theory} can serve as a derivative representation of discrete signals, which extends the computational interval and offers some noise immunity. Moreover, the continuity of the General Difference results in sequence smoothness. Therefore, we can enhance the smoothness of the denoised signal by constraining the continuity of TGD. 

In computer vision and image processing area, edge refers to places where there appears to be a jump in brightness value in the image. These transitions correspond to peaks in the first-order derivative or zero-crossings in the second-order derivative of intensity~\cite{marr1976early,marr1979computational,marr1980theory}. Thus, determining an image's first- or second-order derivative and identifying the precise location of extremas or zero-crossings constitute the fundamental basis for image edge detection. The prevalent techniques for edge detection fall into two categories~\cite{dharampal2015methods}: gradient-based and Laplacian-based approaches, which correspond to the computation of an image's first- and second-order derivatives, respectively.

\begin{figure}[htb]
    \centering
    \begin{minipage}[b]{1.0\linewidth}
        \centering
        \centerline{\includegraphics[width=0.8\linewidth]{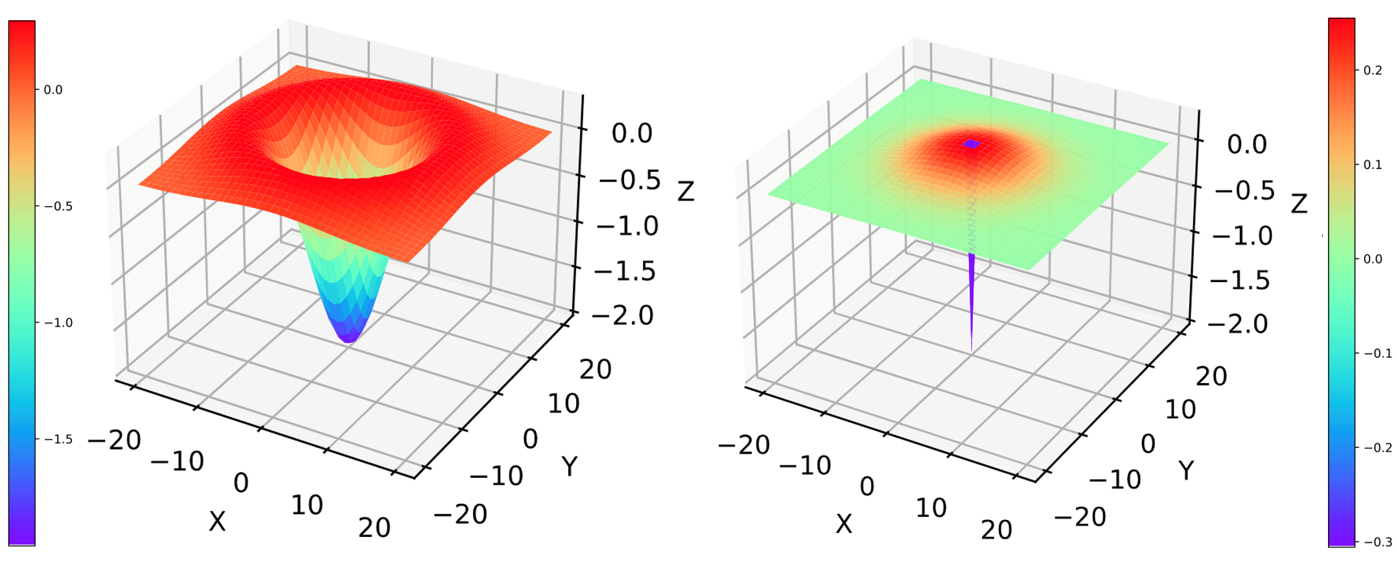}}
    \end{minipage}
    \caption{
        LoG (Laplace of Gaussian) operator (Left) and LoT (Laplace of TGD) operator (Right) constructed with Gaussian kernel function.
    }
    \label{fig:laplacian}
\end{figure}

The Laplacian of Gaussian (LoG) (Figure~\ref{fig:laplacian} Left), is a well known and widely used edge detection tool in computer vision ~\cite{DBLP:journals/pami/TorreP86,wang2007laplacian,mlsna2009gradient}. The kernel of LoG is the second-order derivative of Gaussian, which brings highlighted regions of rapid intensity changes. As we have mentioned in the Theory of TGD, Gaussian does not fit the Monotonic Convexity Constraint (C3), therefore, LoG is not a difference operator. 

Figure~\ref{fig:laplacian} shows the 2D LoG as well as the LoT operator constructed using a Gaussian kernel function. Both operators are normalized to have a minimum value of $-2$. Although they share a commonality in being created using Gaussian functions, they differ in the distribution of negative values. With the LoG operator, the negative values are concentrated around the center, and the maximum positive values are far from the center. In contrast, with the LoT operator, negative values are only in the center, and the maximum positive values are next to the center. This difference embodies the Monotonic Constraint, which, in the theory of TGD, ensures that the zero-crossing positions of difference values (where the original signal changes fastest locally) obtained using the LoT operator are unbiased.

\begin{figure}[!htbp]    	
    \centering    	
    \subfigure[First-order Gaussian Derivative Operator]{  			 
        \includegraphics[width=0.45\linewidth]{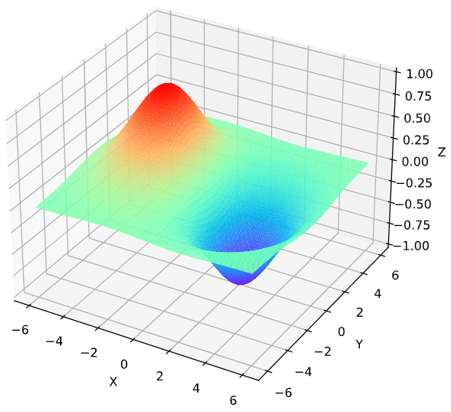}
    }
    \subfigure[First-order TGD Operator]{  			 
        \includegraphics[width=0.45\linewidth]{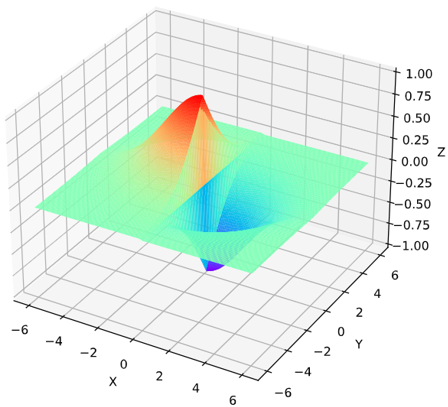}
    }
    \subfigure[First-order Gaussian Derivative Operator]{  			 
        \includegraphics[width=0.45\linewidth]{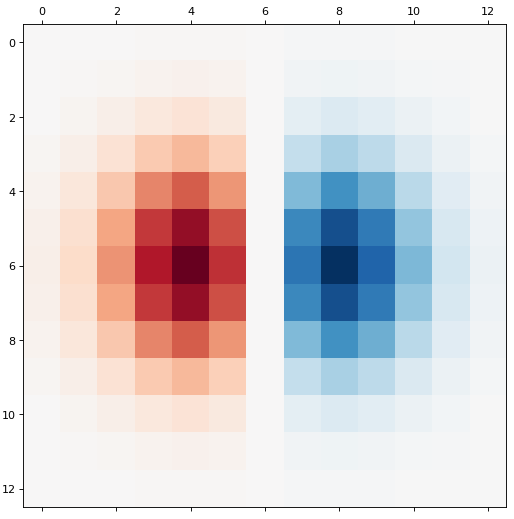}
    }
    \subfigure[First-order TGD Operator]{  			 
        \includegraphics[width=0.45\linewidth]{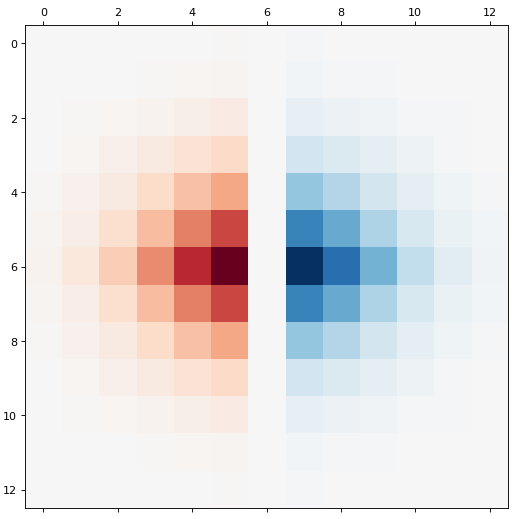}
    }
    \caption{The first-order Gaussian derivative operator and TGD operator, whose kernel function is Gaussian.}  
    \label{fig:GaussVsOur1} 
\end{figure}

The difference between the kernels of LoT and LoG leads to the difference in the proximity of their zero-crossings. The calculated zero-crossing locates at the center of intensity change and responses well to the fastest change in LoT, while the zero-crossing is at the vicinity of the center and responses to the slow change only in LoG (see Figure~\ref{fig:laplacian}). A similar behavior can be observed in first-order difference operators constructed by the first-order derivative of Gaussian and TGD with Gaussian kernel function (see Figure~\ref{fig:GaussVsOur1}). 

Based on the Theory of TGD, 3D kernels can be constructed to compute gradients in 3D discrete sequences. Thus, the TGD-based edge detection method can be extended naturally to the 3D case, and we expand the definition of an edge from 2D images to 3D image sequences including videos, where there is a jump change in brightness, corresponding to local extrema of the gradient in spatiotemporal space. 

This paper is organized in four sections: Firstly, we give a novel definition of smoothness of a sequence based on TGD, and demonstrate the application of TGD in 1D signal denoising. Secondly, we show a property of locating the center of gradient and the application in image edge detection. Thirdly, we introduce the application of TGD in video analytics. Finally, a short conclusion and outlook are given.

\clearpage
\newpage

\section{1D TGD-Based Signal Denoise}

\subsection{TGD-Based Signal Denoise Algorithm}

Among a plethora of data smoothing techniques, the most intuitive method is moving average~\cite{DBLP:reference/stat/Hyndman11b} or convolution smoothing~\cite{clark1977non} with a smoothing kernel such as the Gaussian~\cite{o1997pragmatic}. From a frequency point of view, convolution smoothing is low-pass filtering, and the result is similar to the Fourier thresholding technique~\cite{huang1974noise}, which eliminates the high-frequency components that fall below a certain threshold. Alongside the Fourier transform, the wavelet shrinkage~\cite{o1997pragmatic,taswell2000and} modifies wavelet coefficients based on specific criteria to achieve noise suppression. Singular-value decomposition (SVD) based denoise ~\cite{chen2019denoising} maps the noisy signal to a partial circulant matrix and reconstructs smoothed sequence on the low-rank approximation of the matrix containing only the signal components.

Unlike the aforementioned methods, a noteworthy mainstream technique is the total variation (TV) denoising~\cite{rudin1992nonlinear,condat2013direct}, which directly accounts for the continuity of the discrete signal, i.e., the difference between adjacent points. Specifically, given a noisy signal $X=(x_1, \ldots, x_N) \in \mathbb{R}^N$ of size $N \geq 1$, and we want to efficiently compute the denoised signal $Y^* \in \mathbb{R}^N$, defined implicitly as the solution to the minimization problem:
\begin{equation}
   \begin{aligned}
       \underset{Y \in \mathbb{R}^N}{\operatorname{minimize}} \sum_{i=1}^N\left(Y[i]-X[i]\right)^2+\lambda \left(\sum_{i=1}^{N-1}\left|Y[i+1]-Y[i]\right|^p\right)^{\frac{1}{p}}
   \end{aligned}
   \label{eq:TVDoriginal}
\end{equation}
where the regularization parameter $\lambda \geq 0$, and determines the similarity between the output signal and the input signal. Notice, when using $\ell_1$ norm with $p = 1$, the function to minimize is strongly convex, and the solution $Y^*$ to the problem exists and is unique whatever the noisy signal $X$. 

Actually, Formula~\eqref{eq:TVDoriginal} imposes a constraint on the continuity of the sequence without any stipulations on its “derivative”. Consequently, TV denoising only approximates the output signal to a $C^0$ function\footnote{Consider an open set $U$ on the real line and a function $f$ defined on $U$ with real values. $k$ is a non-negative integer. The function $f$ is said to be a $C^{k}$ function if the derivatives $f^{\prime},\dots,f^{(k)}$ exist and are continuous on $U$. If $f$ is a $C^{k}$ function, then $f$ must be a $C^{k-1}$ function~\cite{warner1983foundations}.}, which is continuous but may lack smoothness due to its "derivative" not being calculated. 

Mathematically, a function must be smooth if its derivative is continuous. Therefore, intuitively, imposing a continuity constraint on the “derivative” of a discrete signal would result in a smoother denoised signal:
\begin{equation}
   \begin{aligned}
       \underset{Y \in \mathbb{R}^N}{\operatorname{minimize}} \sum_{i=1}^N\left(Y[i]-X[i]\right)^2+\lambda \left(\sum_{i=1}^{N-1}\left|Y^{\prime}[i+1]-Y^{\prime}[i]\right|^p\right)^{\frac{1}{p}}
   \end{aligned}
   \label{eq:TVDImprove}
\end{equation}

However, the traditional forward, backward, and center difference schemes are not the derivative of a sequence. The Formula~\eqref{eq:TVDImprove} can not be solved in this traditional difference schemes. In practise, the resulting sequence driven by Formula~\eqref{eq:TVDImprove} with the center difference has an even lower signal-to-noise ratio (SNR). 

\begin{figure}[!htb]
  \begin{minipage}[b]{1.0\linewidth}
    \centering
    \centerline{\includegraphics[width=\linewidth]{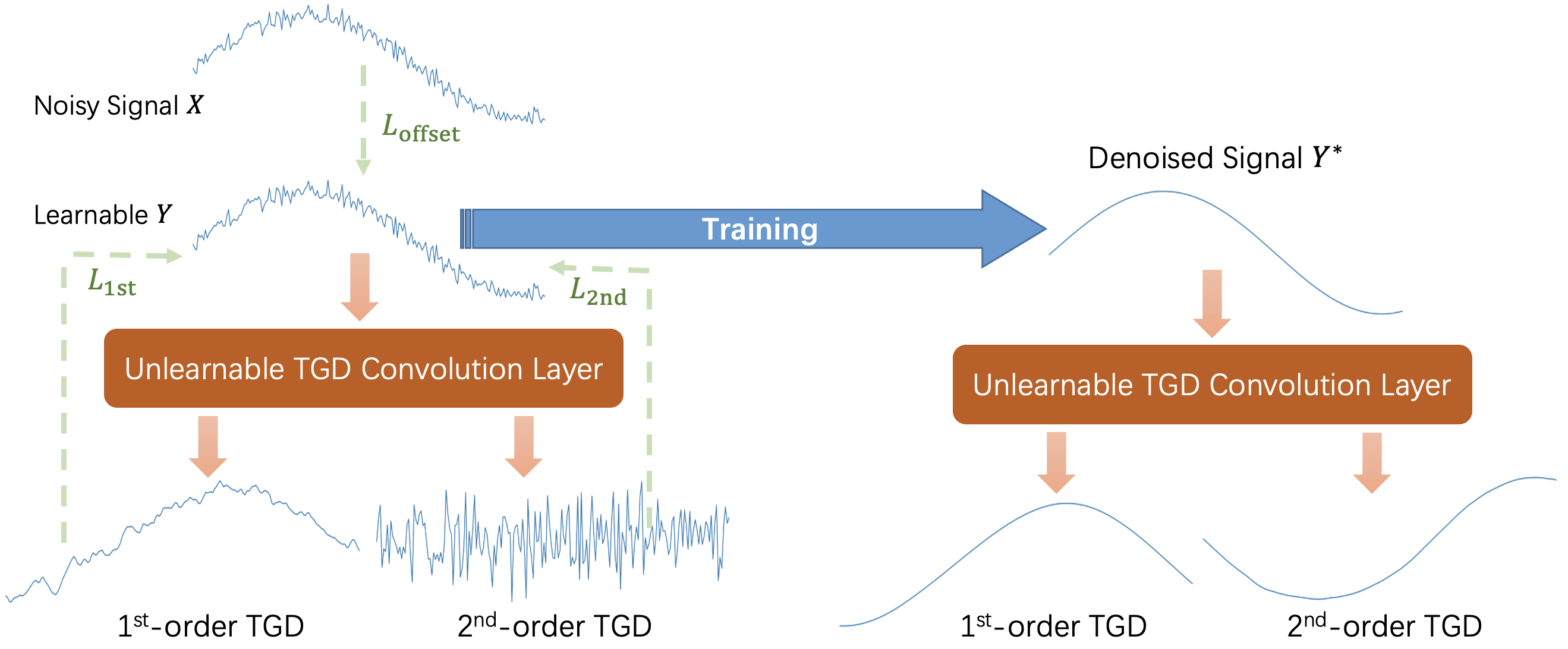}}
  \end{minipage}
  \caption{
    Schematic diagram of TGD-based signal denoise algorithm.
  }
  \label{fig:TGDDenoiseAlgorithm} 
\end{figure}

Naturally, the challenge in Formula~\eqref{eq:TVDImprove} can be solved in TGD theory by enhancing the smoothness of the denoised signal through constraining the continuity of a sequence via TGD. The novel TGD based minimization algorithm is expressed as follows:

\begin{equation}
   \begin{aligned}
       \underset{Y \in \mathbb{R}^N}{\operatorname{minimize}} \sum_{i=1}^N\left(Y[i]-X[i]\right)^2&+\lambda_{\text{1st}} \left(\sum_{i=1}^{N-1}\left|Y^{\prime}_{\text{TGD}}[i+1]-Y^{\prime}_{\text{TGD}}[i]\right|^p\right)^{\frac{1}{p}} \\
       &+\lambda_{\text{2nd}} \left(\sum_{i=1}^{N-1}\left|Y^{\prime\prime}_{\text{TGD}}[i+1]-Y^{\prime\prime}_{\text{TGD}}[i]\right|^p\right)^{\frac{1}{p}}
   \end{aligned}
   \label{eq:TVDnew}
\end{equation}

where the regularization parameter $\lambda_{\text{1st}} \geq 0$ and $\lambda_{\text{2nd}} \geq 0$. When $\lambda_{\text{1st}} > 0$ and $\lambda_{\text{2nd}} = 0$, the first-order TGD of the output signal is required continuous and the denoised signal approximates a $C^1$ function. When $\lambda_{\text{2nd}} > 0$, the first- and second-order TGDs of the output signal are required continuous and the denoised signal approximates a $C^2$ function.

Although Formula~\eqref{eq:TVDnew} is non-convex, a neural network with only a single convolution layer is used to learn the optimization. Figure~\ref{fig:TGDDenoiseAlgorithm} shows the schematic diagram of the TGD-based denoising algorithm. In terms of implementation, we first set the given noisy signal $X$ of size $N \geq 1$ as a learnable parameter matrix $Y$ in the neural network.
Then we give $N_{\text{1st}}$ 1D first-order discrete TGD operators and $N_{\text{2nd}}$ second-order discrete TGD operators (examples in  Formula~\eqref{eq:DTGD_Example 7}), and set them as non-learnable weights to a convolutional layer $Conv$ with input channel of $1$ and output channel of $N_{\text{1st}} + N_{\text{2nd}}$. 

\begin{equation}
    \setcounter{MaxMatrixCols}{35}
    \begin{aligned}
        \widehat{T}_{\text{Gaussian}} &= \frac{1}{131}\!\begin{bmatrix} 1&3&8&18&32&50&64&0&-64&-50&-32&-18&-8&-3&-1 \end{bmatrix} \\
        \widehat{R}_{\text{Gaussian}} &= \frac{1}{178}\begin{bmatrix} 1&3&8&18&32&50&64&-356&64&50&32&18&8&3&1 \end{bmatrix} \\
    \end{aligned}
    \label{eq:DTGD_Example 7}
\end{equation}

TGD results are calculated by feeding the learnable matrix $Y$ to the convolutional layer\footnote{The batch dimension is ignored here to simplify the representation.}:
\begin{equation}
    \begin{aligned}
        Z &= Conv\left(Y\right) \\
        Y^{\prime}_{\text{TGD}} &= Z[1:N_{\text{1st}}] \\
        Y^{\prime\prime}_{\text{TGD}} &= Z[N_{\text{1st}}+1:N_{\text{1st}} + N_{\text{2nd}}]
    \end{aligned}
    \label{eq:TGDinterpolation_conv}
\end{equation}

\begin{figure}[!htb]
  \begin{minipage}[b]{1.0\linewidth}
    \centering
    \centerline{\includegraphics[width=\linewidth]{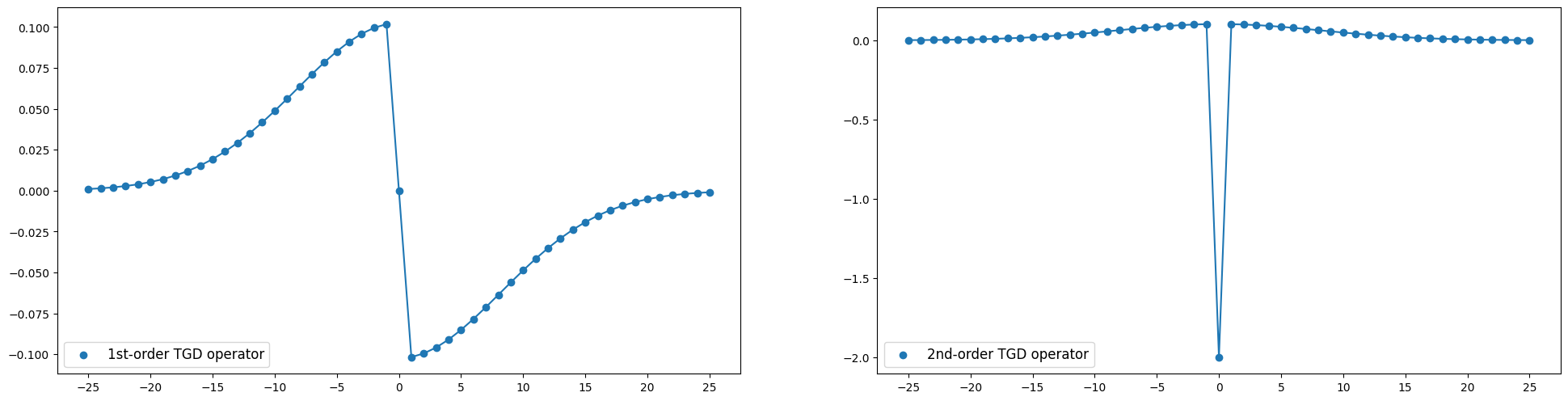}}
  \end{minipage}
  \caption{
    The discrete TGD operators used in the denoising experiments with size of $51$, which are constructed based on the Gaussian kernel function.
  }
  \label{fig:TGDDenoiseOperator} 
\end{figure}

Next, we convert Formula~\eqref{eq:TVDnew} into loss functions for training the learnable denoised signal $Y$, which includes three parts: first-order TGD continuity loss $L_{\text{1st}}$, second-order TGD continuity loss $L_{\text{2nd}}$, and sampling point offset loss $L_{\text{offset}}$:
\begin{equation}
    \begin{aligned}
        L_{\text{1st}} &= \sum_{i = 1}^{N_{\text{1st}}}\left(\sum_{j = 1}^{N-1} \left|Y^{\prime}_{\text{TGD}}[i,j] - Y^{\prime}_{\text{TGD}}[i,j+1]\right|^p\right)^{\frac{1}{p}} \\
        L_{\text{2nd}} &= \sum_{i = 1}^{N_{\text{2nd}}}\left(\sum_{j = 1}^{N-1} \left|Y^{\prime\prime}_{\text{TGD}}[i,j] - Y^{\prime\prime}_{\text{TGD}}[i,j+1]\right|^p\right)^{\frac{1}{p}} \\
        L_{\text{offset}} &= \sum_{i = 1}^{N} \left(Y[i] -  X[i]\right)^2 \\
    \end{aligned}
    \label{eq:TGDinterpolation_loss}
\end{equation}

And the final loss function is expressed as follows:
\begin{equation}
    \begin{aligned}
        L_{\text{final}} = \lambda_{\text{1st}}L_{\text{1st}} + \lambda_{\text{2nd}}L_{\text{2nd}} + \lambda_{\text{offset}}L_{\text{offset}}
    \end{aligned}
    \label{eq:TGDinterpolation_loss_final}
\end{equation}
where $\lambda_{\text{1st}}$, $\lambda_{\text{2nd}}$ and $\lambda_{\text{offset}}$ are loss factors. $L_{\text{1st}}$ and $L_{\text{2nd}}$ contribute to the continuity of the TGD results for the denoised signal, resulting in a smoother denoised signal. Besides, $L_{\text{offset}}$ facilitates the retention of similarity between the original and denoised signals. Iteratively optimizing the loss function yields the final denoised signal.

\subsection{Experiments and Analysis}

\subsubsection{Qualitative Experiments}

Let's take two examples to demonstrate the training process and the effectiveness of the TGD-based denoise algorithm. In a noisy environment, two discrete sequences were sampled from two smooth functions, respectively:
\begin{equation}
    \begin{aligned}
        X_1\left(n\right) &= 20\sin\left(\frac{n}{40}\right) + \frac{n^2}{20000} + \delta\left(n\right), \quad n = 0,1,\dots,1000
    \end{aligned}
    \label{eq:noisysignal1}
\end{equation}
\begin{equation}
    \begin{aligned}
        X_2(n) &= \arctan\left(\frac{n - 500}{40}\right) + \frac{n^2}{20000} + \delta\left(n\right), \quad n = 0,1,\dots,1000
    \end{aligned}
    \label{eq:noisysignal2}
\end{equation}
where $\delta$ is the noise.

\begin{figure}[!htb]    	
    \centering    	
    \subfigure[Epoch = 0]{  			 
        \includegraphics[width=\linewidth]{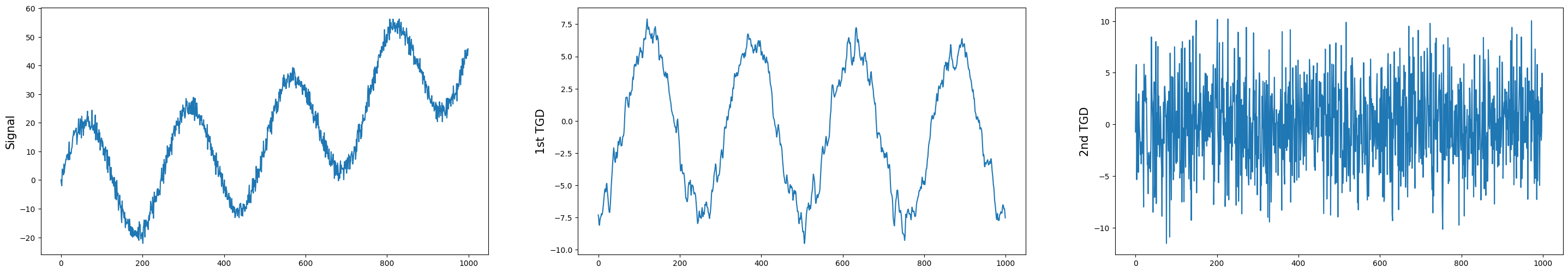}
    } 
    \subfigure[Epoch = 100]{   	 		 
        \includegraphics[width=\linewidth]{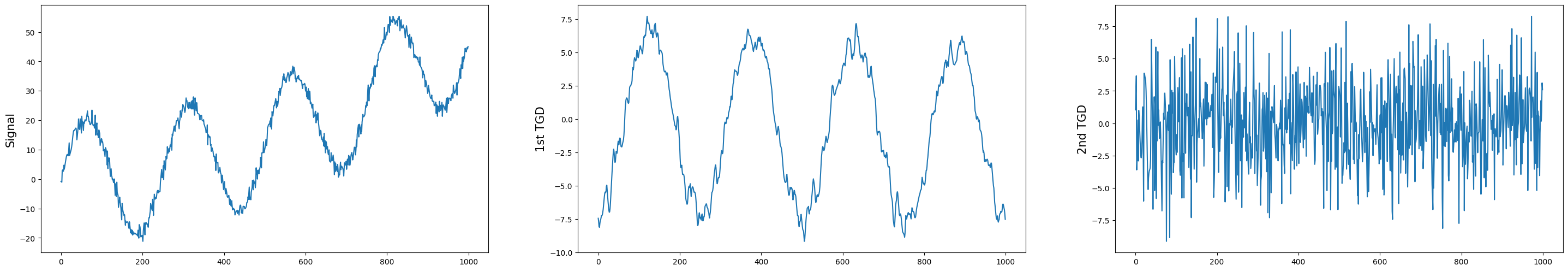}
    }  
    \subfigure[Epoch = 1000]{   	 		 
        \includegraphics[width=\linewidth]{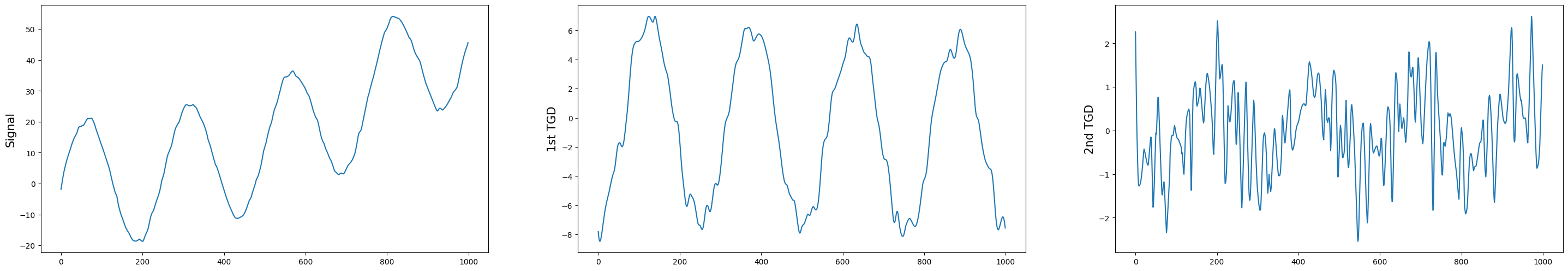}
    }     	 
   \subfigure[Epoch = 10000]{  			 
       \includegraphics[width=\linewidth]{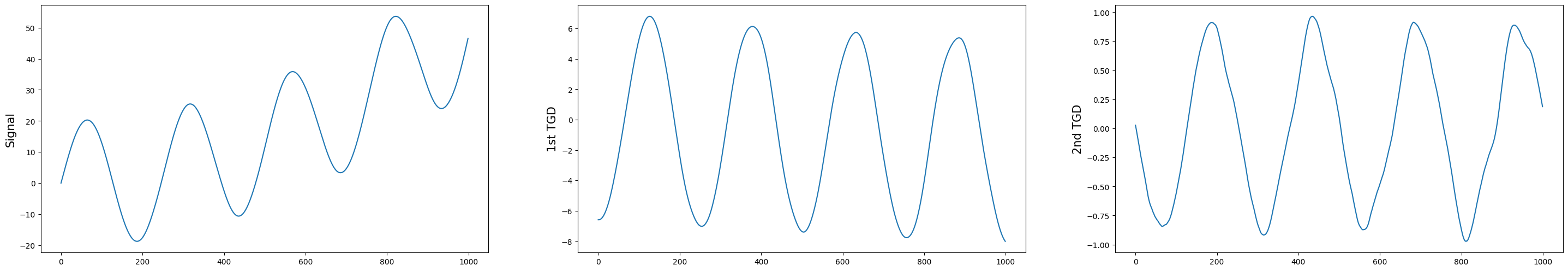}
   } 
    \subfigure[Epoch = 100000]{   	 		 
        \includegraphics[width=\linewidth]{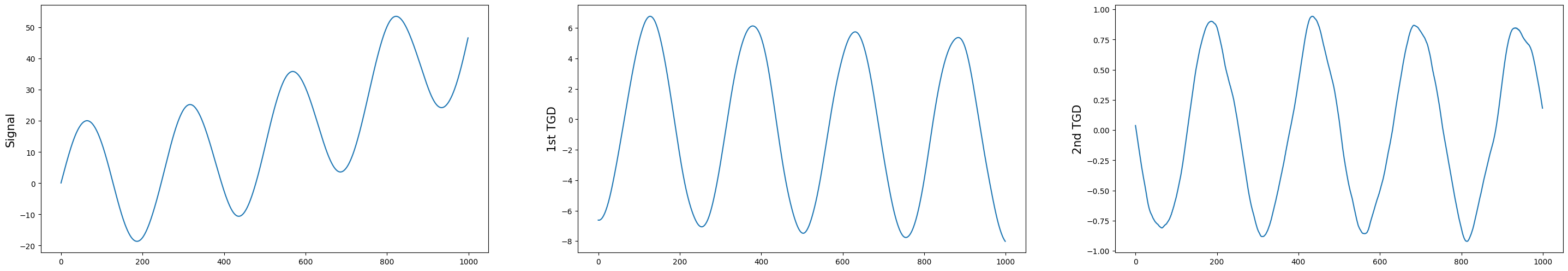}
    } 	
    \caption{The training process for noisy signal $X_1$. The left column shows the current denoised signal. The middle column shows the first-order TGD results of the current denoised signal, and the rightmost column shows the second-order TGD results of the current denoised signal.}  
    \label{fig:TDGDenoiseExample1} 
\end{figure}

\begin{figure}[!htb]    	
    \centering    	
    \subfigure[Epoch = 0]{  			 
        \includegraphics[width=\linewidth]{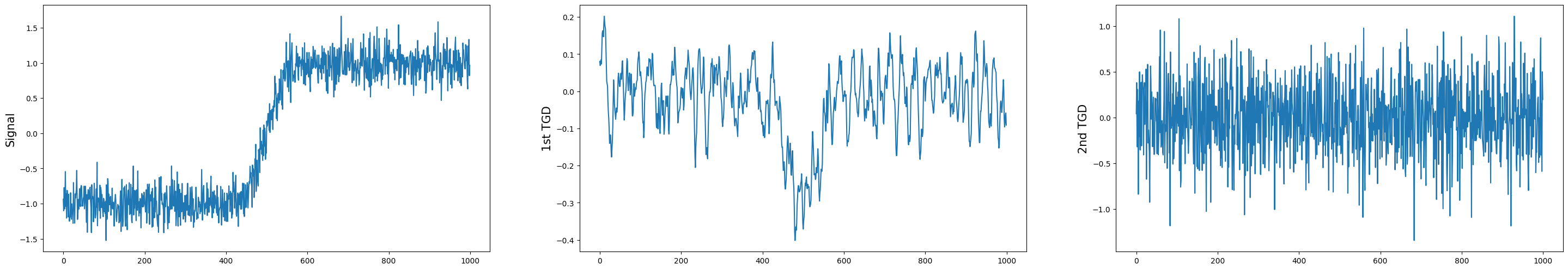}
    } 
    \subfigure[Epoch = 100]{   	 		 
        \includegraphics[width=\linewidth]{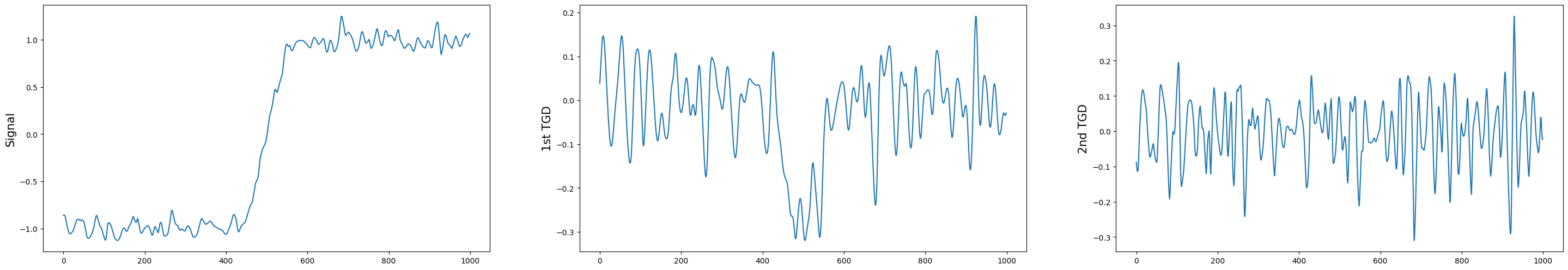}
    }  
    \subfigure[Epoch = 1000]{   	 		 
        \includegraphics[width=\linewidth]{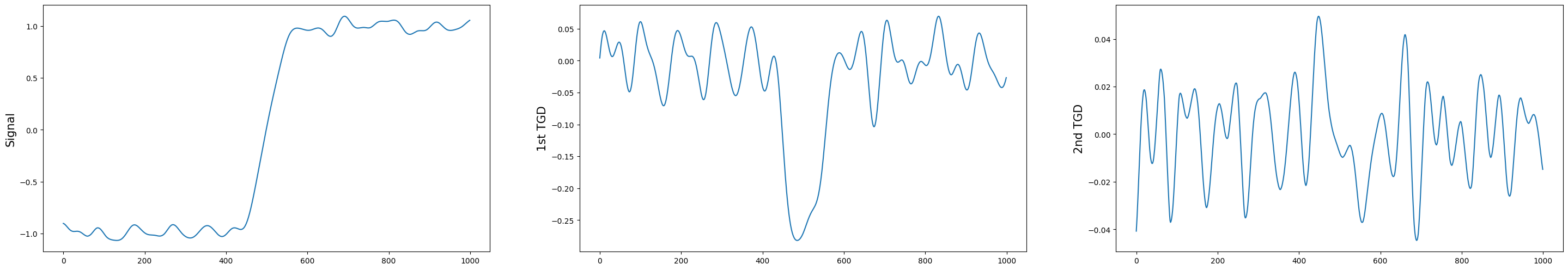}
    }     	 
   \subfigure[Epoch = 10000]{  			 
       \includegraphics[width=\linewidth]{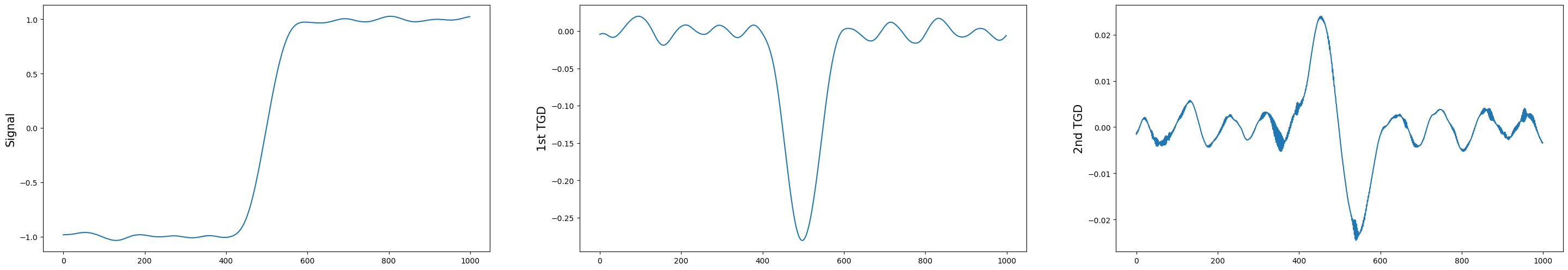}
   } 
    \subfigure[Epoch = 100000]{   	 		 
        \includegraphics[width=\linewidth]{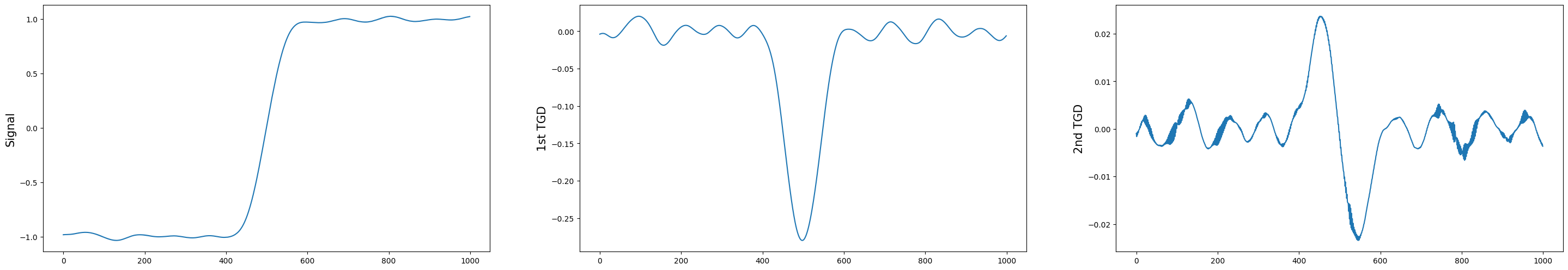}
    } 	
    \caption{The training process for noisy signal $X_2$. The left column shows the current denoised signal. The middle column shows the first-order TGD results of the current denoised signal, and the rightmost column shows the second-order TGD results of the current denoised signal.}  
    \label{fig:TDGDenoiseExample2} 
\end{figure}

Since the value ranges of $X_1$ and $X_2$ are different, we add Gaussian noise with mean $\mu = 0$ and different variances, $\sigma = 2$ to $X_1$, and $\sigma = 0.2$ to $X_2$. The $\ell_2$ norm is chosen in loss $L_{\text{1st}}$ and $L_{\text{2nd}}$. We eliminate square root calculations and use the sum of squares directly as the loss to speed up the training process. This allows $L_{\text{offset}}$, $L_{\text{1st}}$ and $L_{\text{2nd}}$ to be formally consistent. As for loss factors, we choose $\lambda_{\text{1st}} = 1$, $\lambda_{\text{2nd}} = 10$ and $\lambda_{\text{offset}} = 0.01$. In training, we use the Adam optimizer with a learning rate of $0.01$. The learnable denoised signal is trained for $100,000$ epochs. Using the StepLR strategy, the learning rate decays to 10\% of the current whenever the number of training epochs reaches $10,000$. One first-order TGD operator and a second-order TGD operator of sizes $51$ are used (Figure~\ref{fig:TGDDenoiseOperator}). Due to the effect of padding in the convolution, we actually sampled $1,200$ data points and discarded $100$ points each before and after, keeping only the remaining part in the middle. The training process is shown in Figure~\ref{fig:TDGDenoiseExample1} and~\ref{fig:TDGDenoiseExample2}. When the result converges, we get a smooth denoised signal that approximates a $C^2$ function.

\clearpage

\clearpage

\begin{figure}[!htb]    	
    \centering    	
    \subfigure[Signal $X_1$ with uniform noise]{  			 
        \includegraphics[width=0.48\linewidth]{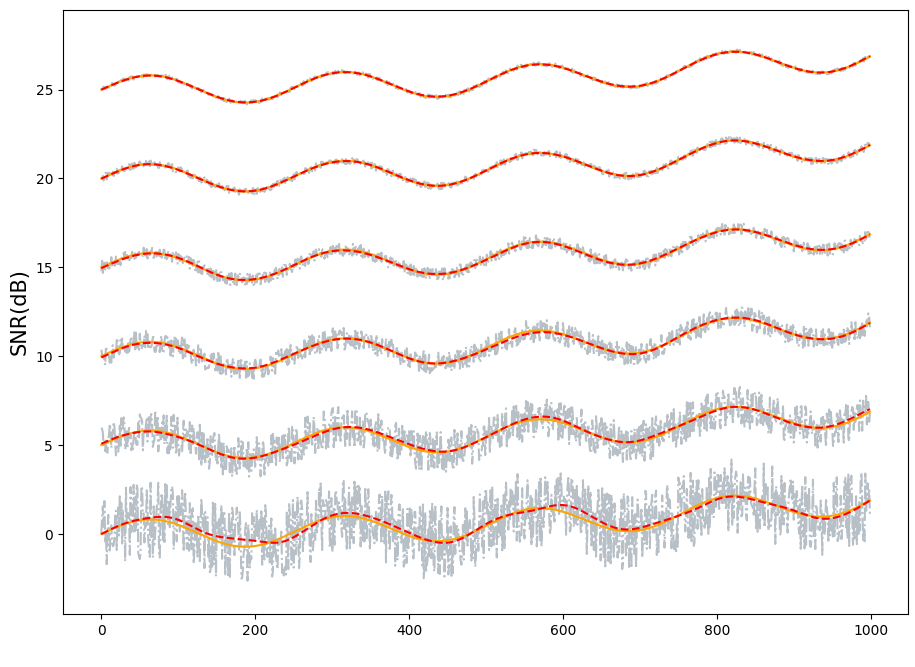}
    } 
    \subfigure[Signal $X_1$ with Gaussian noise]{   	 		 
        \includegraphics[width=0.48\linewidth]{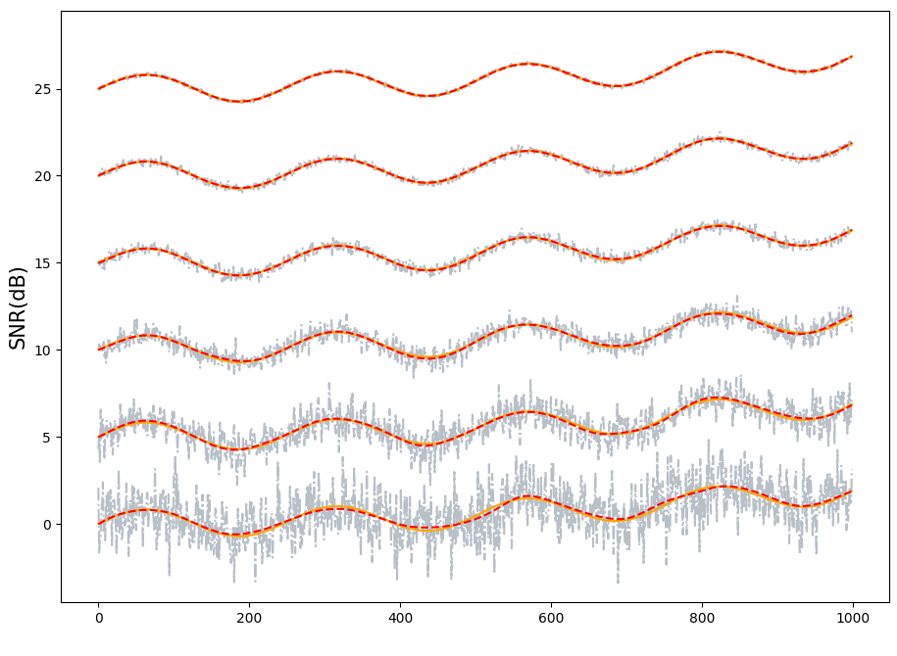}
    }  
    \subfigure[Signal $X_2$ with uniform noise]{   	 		 
        \includegraphics[width=0.48\linewidth]{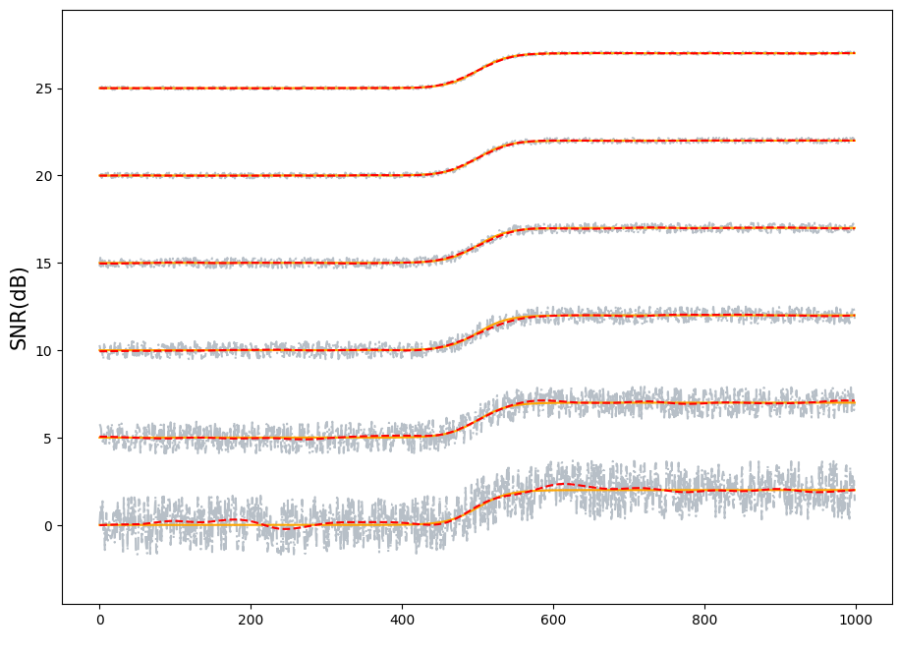}
    }  
    \subfigure[Signal $X_2$ with Gaussian noise]{   	 
        \includegraphics[width=0.48\linewidth]{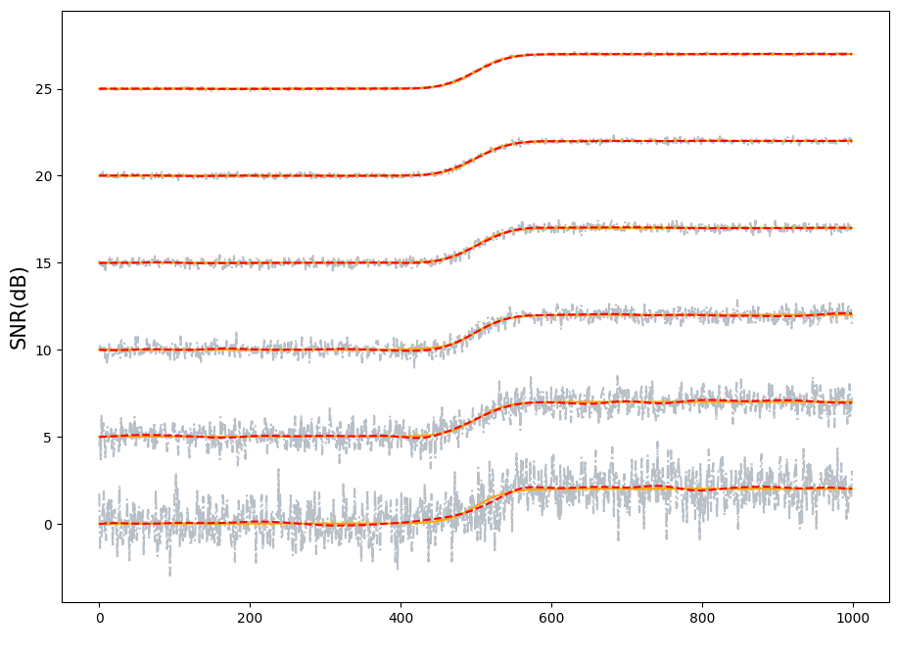}
    }  
    \caption{Illustration of the smoothing effect on the noise-corrupted signals with various SNRs. The noise obeys a uniform and normal distribution, respectively. The gray dashed line is the noisy signal, the orange line is the ground truth, and the red dashed line is the denoised approximation.}  
    \label{fig:TDGDenoiseExample3} 
\end{figure}

Further, we investigate the denoising ability of the TGD-based denoise algorithm for different noise levels. The TGD-based denoise algorithm only attaches smoothness assumptions to the original signal and does not rely on prior knowledge of the noise. Here, the signals are contaminated by adding respectively two types of noise, namely uniform and normal, to produce noisy sequences. The amount of the contamination is controlled by the signal-to-noise ratio (SNR) ranging from $0$ to $25$ dB, which is defined as:
\begin{equation}
    \begin{aligned}
        \text{SNR}_{\text{db}}=10 \log_{10} \frac{P(X_{\text{signal}})}{P(X_{\text{noise}})}
    \end{aligned}
\end{equation}
where the power $P$ of a signal $X$ is measure via:
\begin{equation}
    \begin{aligned}
        P(X) = \frac{1}{N}  \sum_{i=1}^{N} X[i]^2
    \end{aligned}
\end{equation}

Figure~\ref{fig:TDGDenoiseExample3} displays the denoising results, using the same training settings as previously mentioned. The TGD-based algorithm is nonparametric and capable of removing any additive noise. The results indicate that the smooth signal recovery is effective and little distortion occurs in different noise environments, even at an SNR of $0$ dB.

\subsubsection{Quantitative Experiments}
Following a qualitative analysis of the effectiveness of the TGD-based denoise algorithm, we then provide a quantitative comparison of the proposed TGD-based algorithm with previous methods. For evaluation metrics, we use mainstream measures, including RMSE, PSNR, and SSIM.

The root-mean-square error (RMSE) is a common metric used to quantify the differences between predicted and true values. RMSE is always non-negative, and a value of $0$ would indicate a perfect fit. The RMSE of the original signal with respect to a denoised signal is determined as the square root of the mean squared error:
\begin{equation}
    \begin{aligned}
        \text{RMSE} (X, Y) = \sqrt{\frac{1}{N} \sum_{i = 1}^{N}(X[i] - Y[i])^2}
    \end{aligned}
\end{equation}

The peak signal-to-noise ratio (PSNR) is a metric used to measure the level of noise present in a signal. Specifically, it is the ratio of the maximum possible signal power to the power of corrupting noise that may affect its fidelity. Typically, a higher PSNR value indicates less signal distortion from the noise. The PSNR is defined as:
\begin{equation}
    \begin{aligned}
        \text{PSNR} (X, Y) = 20 \log_{10}\left(\frac{\text{MAX}}{\text{RMSE} (X, Y)}\right)
    \end{aligned}
\end{equation}
where $\text{MAX}$ is the maximum possible value of the signal.

The structural similarity index measure (SSIM) is employed to determine the similarity between two signals~\cite{moore2014denoising}. The formula below can be used to compute the SSIM, where $\mu$ represents the mean value and $\sigma^2$ is the variance, $\sigma_{X Y}$ is the covariance of the two signals:
\begin{equation}
    \begin{aligned}
        \operatorname{SSIM}(X, Y)=\frac{\left(2 \mu_X \mu_Y+c_1\right)\left(2 \sigma_{X Y} +c_2\right)}{\left(\mu_X^2+\mu_Y^2+c_1\right)\left(\sigma_X^2+\sigma_Y^2+c_2\right)}
    \end{aligned}
\end{equation}
where $c_1=\left(k_1 L\right)^2, c_2=\left(k_2 L\right)^2$, and $k_1 = 0.01, k_2 = 0.03$ by default. $L$ is the dynamic range of signal values.

Four classical denoising methods are compared with the proposed TGD-based denoise algorithm, including Gaussian smoothing, wavelet shrinkage, SVD, and TV denoising, respectively. The noisy signals used in experiments are given by Formula~\eqref{eq:noisysignal1} and~\eqref{eq:noisysignal2}, where Gaussian noise is added. The experimental setup used in the proposed algorithm keeps the same as that used in the qualitative analysis. To be consistent with the TGD-based algorithm, we also use size $51$ for the Gaussian smoothing kernel. When applying the wavelet shrinkage, the \emph{db8} is adopted to perform the decomposition and the level is $5$. The wavelet transform is carried out with \emph{Pywavelets} package~\cite{lee2019pywavelets}. The TV is based on the algorithm for one-dimensional data in Ref.~\cite{condat2013direct}\footnote{\url{https://github.com/MrCredulous/1D-MCTV-Denoising}}, and the regularization parameter is selected to be $4.47$\footnote{Suggested in the reference implementation, the regularization parameter $\lambda = \sqrt{0.5 N}/5$, where $N$ is the length of signals.}. The SVD is based on the algorithm in Ref.~\cite{chen2019denoising}\footnote{\url{https://github.com/nerdull/denoise}}, where the rows of the constructed matrix are $300$.

Denoise algorithms are evaluated and compared quantitatively in Table~\ref{tab:TGDDenoiseCompare1}. The best results are bolded while the second-best results are underlined. The lowest RMSE and highest PSNR are achieved by the TGD-based denoise algorithm, while also maintaining a satisfactory SSIM score. The results indicate that the TGD-based algorithm performs better than other mainstream methods, resulting in almost no signal distortion between the denoised and original signals.

\begin{table}[htb]
\centering
\caption{
    Performance comparison of different denoise methods. We added Gaussian noise with different variance, $\sigma = 2$ to $X_1$, and $\sigma = 0.2$ to $X_2$. The best results are bolded while the second-best results are underlined.
  }
\setlength{\tabcolsep}{2.5mm}{
\begin{tabular}{@{}cc|cccc|c@{}}
\toprule
\multicolumn{2}{c|}{Metric} & Gaussian~\cite{o1997pragmatic} & Wavelet~\cite{o1997pragmatic} & TV~\cite{condat2013direct} & SVD~\cite{chen2019denoising} & TGD (ours)      \\ \midrule
\multicolumn{1}{c|}{\multirow{3}{*}{{\begin{tabular}[c]{@{}c@{}} Signal\\$X_1$\end{tabular}}}}  & RMSE & 0.3926   & 0.3568  & 0.8432 & {\ul 0.2715}    & \textbf{0.2480} \\
\multicolumn{1}{c|}{}                         & PSNR & 42.71    & 43.54   & 36.07  & {\ul 45.91}     & \textbf{46.70}  \\
\multicolumn{1}{c|}{}                         & SSIM & 0.9420   & 0.9743  & 0.6325 & \textbf{0.9926} & {\ul 0.9900}    \\ \midrule
\multicolumn{1}{c|}{\multirow{3}{*}{{\begin{tabular}[c]{@{}c@{}} Signal\\$X_2$\end{tabular}}}}  & RMSE & 0.0368   & 0.0285  & 0.0258          & {\ul 0.0242}    & \textbf{0.0228} \\
\multicolumn{1}{c|}{}                         & PSNR & 28.69    & 30.92   & 31.76           & {\ul 32.33}     & \textbf{32.82}  \\
\multicolumn{1}{c|}{}                         & SSIM & 0.9714   & 0.9907  & 0.9745          & {\ul 0.9962}    & \textbf{0.9963} \\
\bottomrule
\end{tabular}}
\label{tab:TGDDenoiseCompare1}
\end{table}

\begin{figure}[!htb]
  \begin{minipage}[b]{1.0\linewidth}
    \centering
    \centerline{\includegraphics[width=0.85\linewidth]{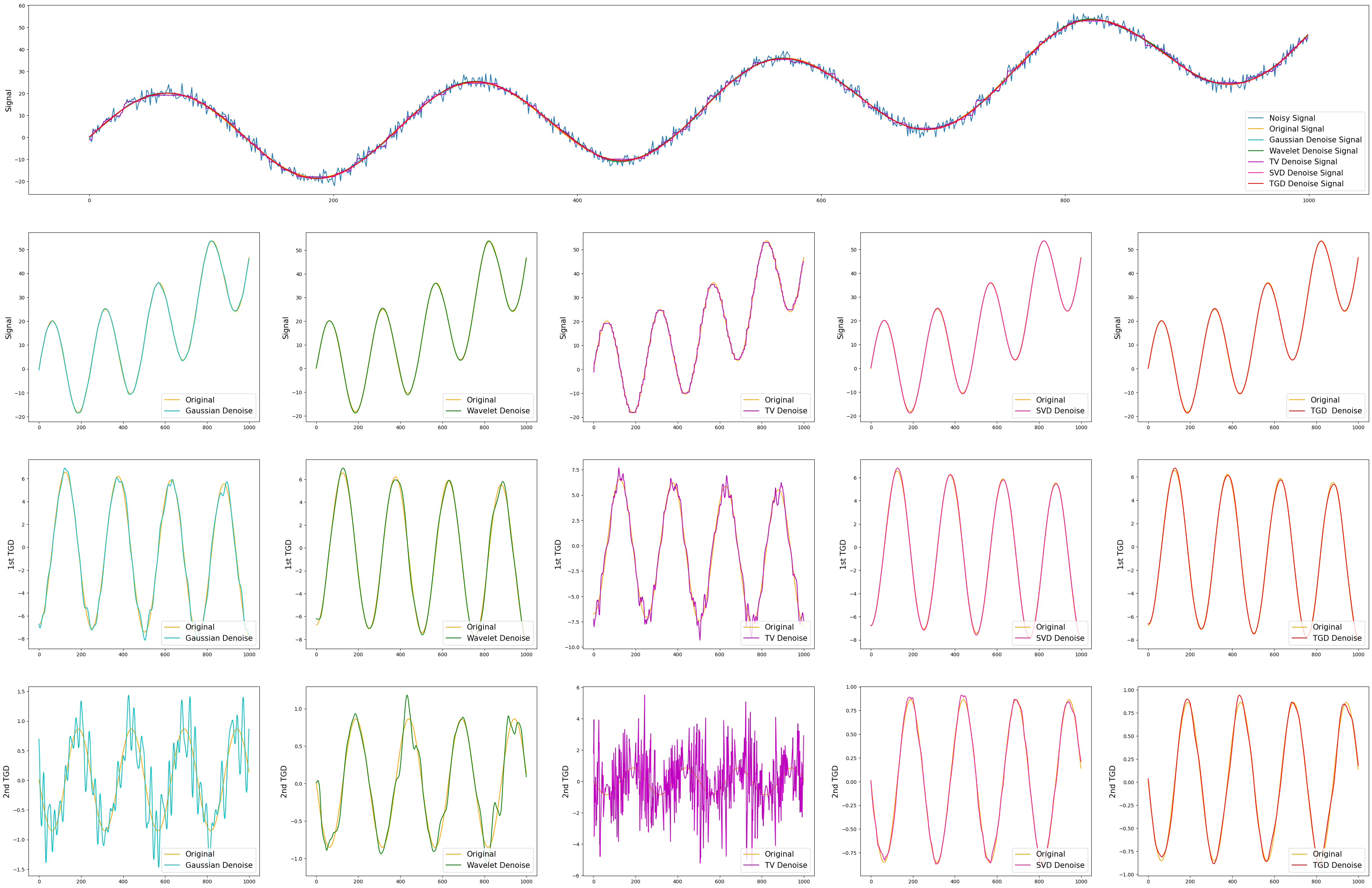}}
  \end{minipage}
  \caption{
    Performance comparison of different denoise methods. We added Gaussian noise with variance $\sigma = 2$ to $X_1$. Denoising methods include: Gaussian convolutional smoothing, wavelet shrinkage, total variation (TV), singular-value decomposition (SVD), and the proposed TGD-based denoise method. The original smooth signal as well as the denoised signal, and their TGD results are compared separately.
  }
  \label{fig:TDGDenoiseCompare1} 
\end{figure}

\begin{figure}[!htb]
  \begin{minipage}[b]{1.0\linewidth}
    \centering
    \centerline{\includegraphics[width=0.85\linewidth]{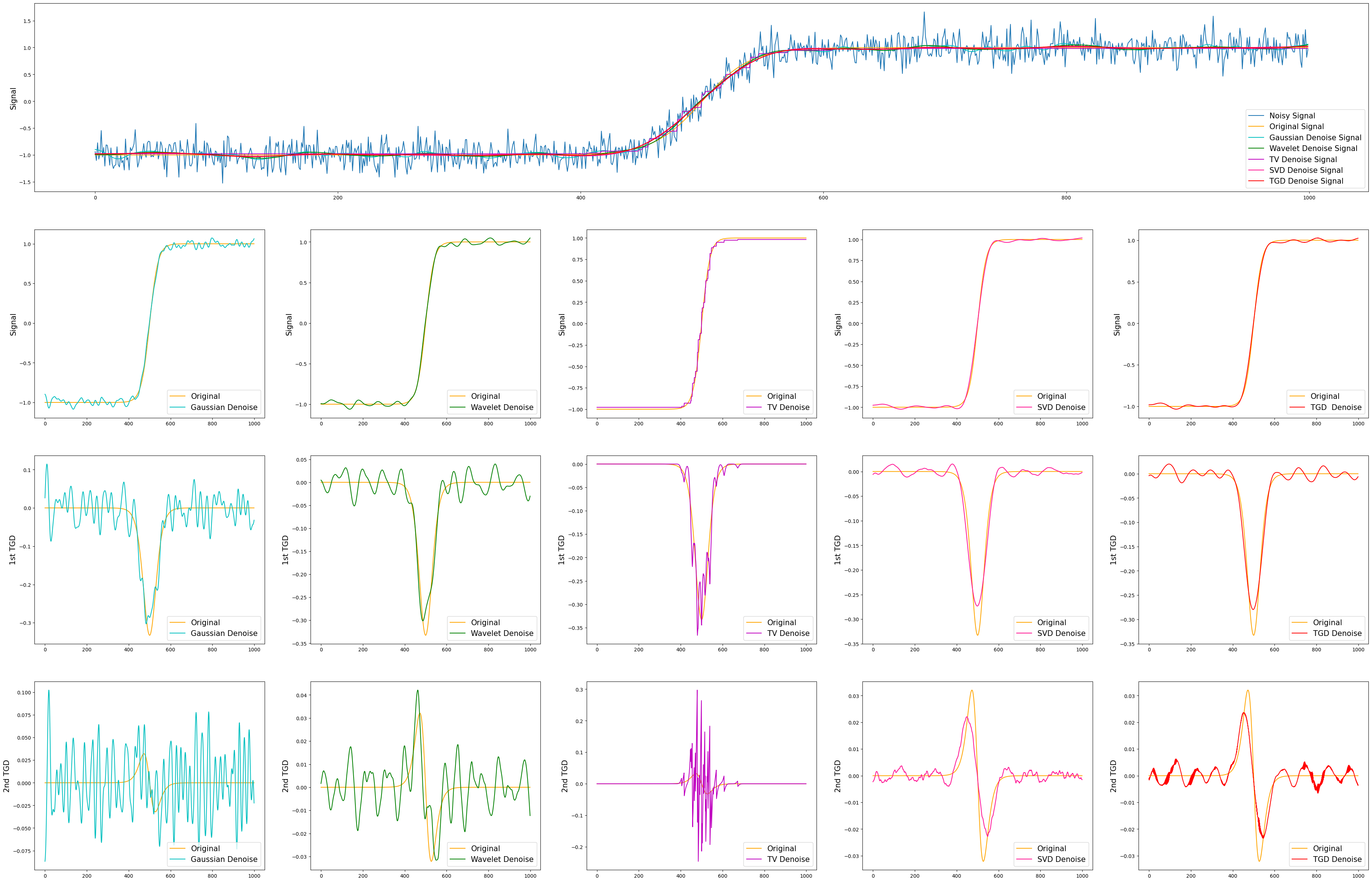}}
  \end{minipage}
  \caption{
    Performance comparison of different denoise methods. We added Gaussian noise with variance $\sigma = 0.2$ to $X_2$. Denoising methods include: Gaussian convolutional smoothing, wavelet shrinkage, total variation (TV), singular-value decomposition (SVD), and the proposed TGD-based denoise method. The original smooth signal as well as the denoised signal, and their TGD results are compared separately.
  }
  \label{fig:TDGDenoiseCompare2} 
\end{figure}

\begin{figure}[!htb]    	
    \centering    	
    \subfigure[Noisy signal $X_1$]{  			 
        \includegraphics[width=0.98\linewidth]{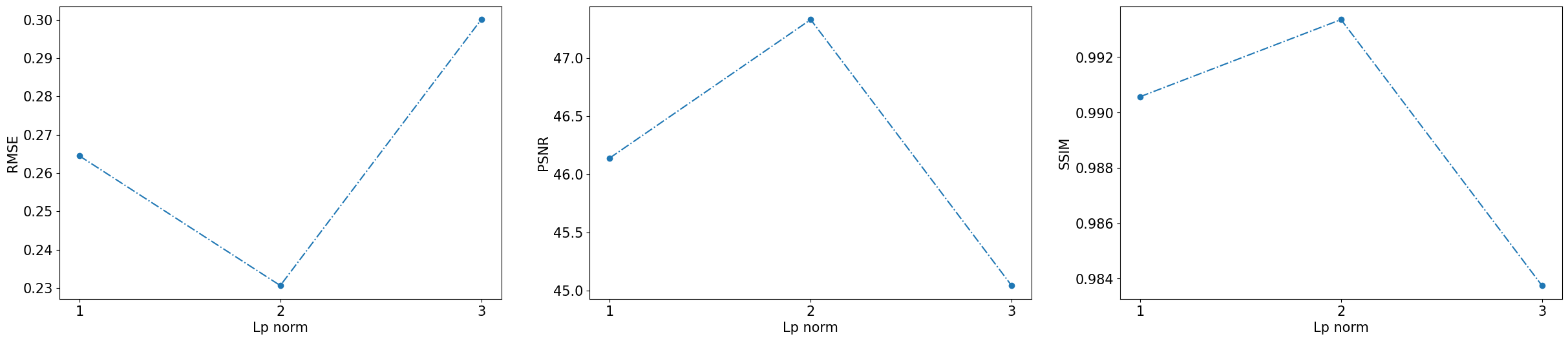}
    } 
    \subfigure[Noisy signal $X_2$]{   	 		 
        \includegraphics[width=0.98\linewidth]{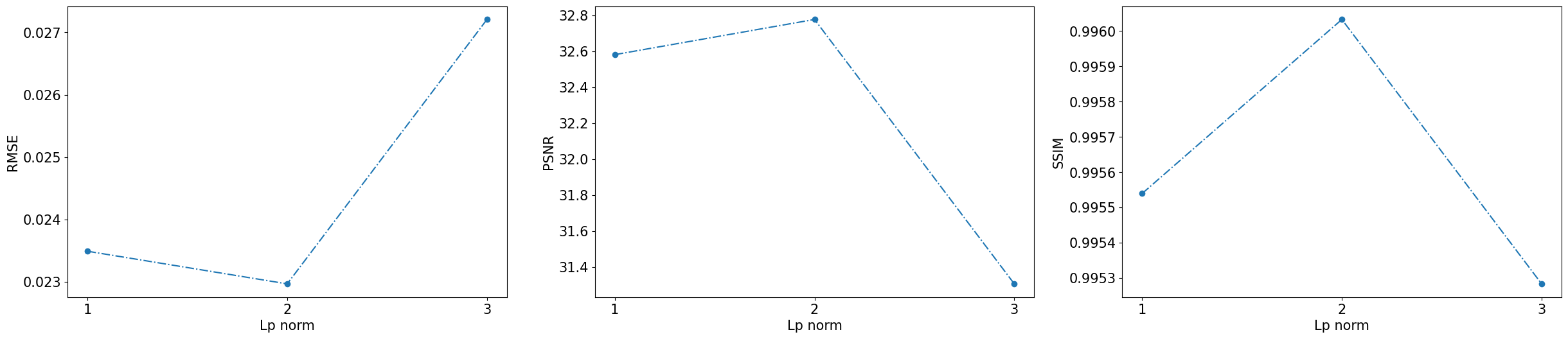}
    }  
    \caption{Comparison of the smoothing effect with different $\ell_p$ norm. Since the loss value may change, we adjusted the loss factor accordingly. The experimental results suggest that the $\ell_2$ norm would be a good choice.} 
    \label{fig:TDGDenoiseLP} 
\end{figure}

Figure~\ref{fig:TDGDenoiseCompare1} and~\ref{fig:TDGDenoiseCompare2} further visualize the comparison between the output denoised signals and original signals. The continuity of the denoised signals is achieved by all methods, but the TV denoise algorithm fails to maintain the continuity of the first-order TGD. This aligns with theoretical analysis showing that a continuous function does not necessarily need to be derivable/smooth. 
Due to the exchangeability of the convolution, calculating the TGD after Gaussian smoothing of the noisy signal is equivalent to Gaussian smoothing of the TGD for the noisy signal. As the SNR of the derivative is lower than that of the noisy signal, Gaussian smoothing does not guarantee continuity in the TGD, particularly second-order TGD, causing the denoised signal to not approximate a $C^2$ function. The continuity of the first- and second-order TGDs is better with wavelet shrinkage denoising than with Gaussian, but still lagged behind the proposed TGD-based algorithm. The closest performance to the proposed algorithm is the SVD denoising method, where they both obtain a signal that approximates a $C^2$ function. However, the TGD-based algorithm is optimal regarding quantitative metrics.

\subsubsection{Ablation Experiments}

In the ablation experiments, the noisy signals are obtained from Formula~\eqref{eq:noisysignal1} and~\eqref{eq:noisysignal2}, where Gaussian noise with $\sigma = 2$ is added to $X_1$, and $\sigma = 0.2$ to $X_2$. First, we compare the smoothing effects of different $\ell_p$ norms in the loss function, including $p = 1$, $2$, and $3$. The experimental setup employed in the TGD-based algorithm remains consistent with that used in the qualitative analysis. It is worth pointing out that, the difference in the $\ell_p$ norms results in variations in the loss magnitude, hence the need to adjust the loss factor for each $p$ choice. The outcomes of RMSE, PSNR, and SSIM metrics are depicted in Figure~\ref{fig:TDGDenoiseLP}. Based on the quantitative results, the $\ell_2$ norm seems to be a better choice.

\begin{figure}[!htb]    	
    \centering    	
    \subfigure[$\lambda_{\text{1st}} = 1$, $\lambda_{\text{2nd}} = 10$ and $\lambda_{\text{offset}} = 0.01$ (Signal RMSE = 0.2480)]{  			 
        \includegraphics[width=0.98\linewidth]{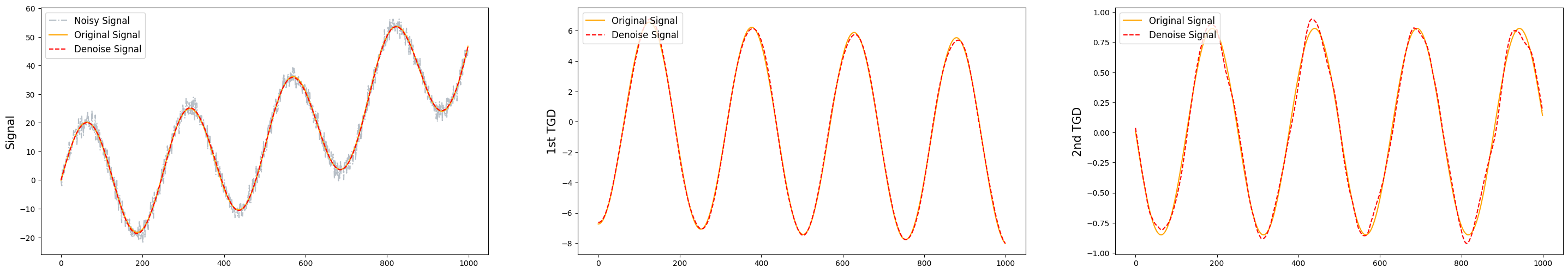}
    } 
    \subfigure[$\lambda_{\text{1st}} = 0$, $\lambda_{\text{2nd}} = 10$ and $\lambda_{\text{offset}} = 0.01$ (Signal RMSE = 0.2832)]{  			 
        \includegraphics[width=0.98\linewidth]{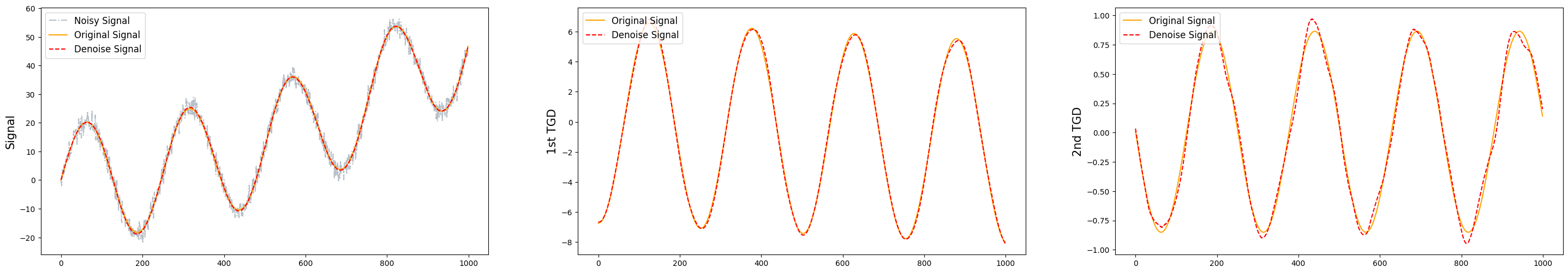}
    } 
    \subfigure[$\lambda_{\text{1st}} = 1$, $\lambda_{\text{2nd}} = 0$ and $\lambda_{\text{offset}} = 0.01$ (Signal RMSE = 1.0508)]{  			 
        \includegraphics[width=0.98\linewidth]{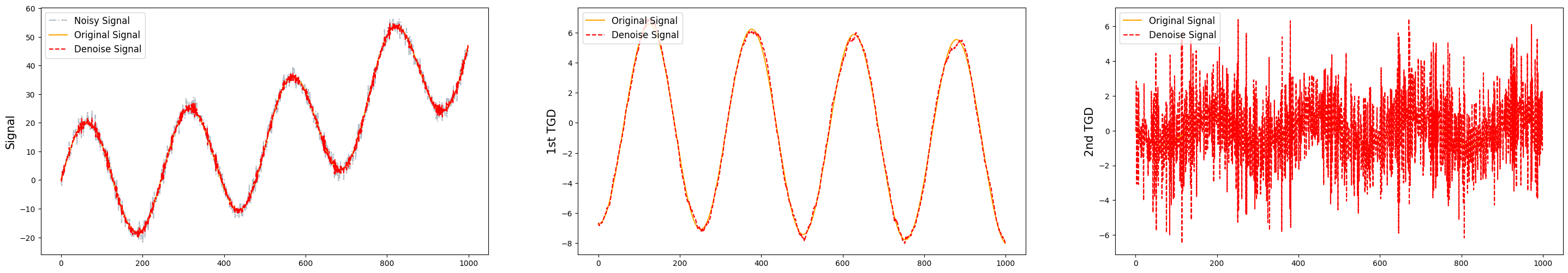}
    } 
    \subfigure[$\lambda_{\text{1st}} = 1$, $\lambda_{\text{2nd}} = 10$ and $\lambda_{\text{offset}} = 0$ (Signal RMSE = 10.7970)]{  			 
        \includegraphics[width=0.98\linewidth]{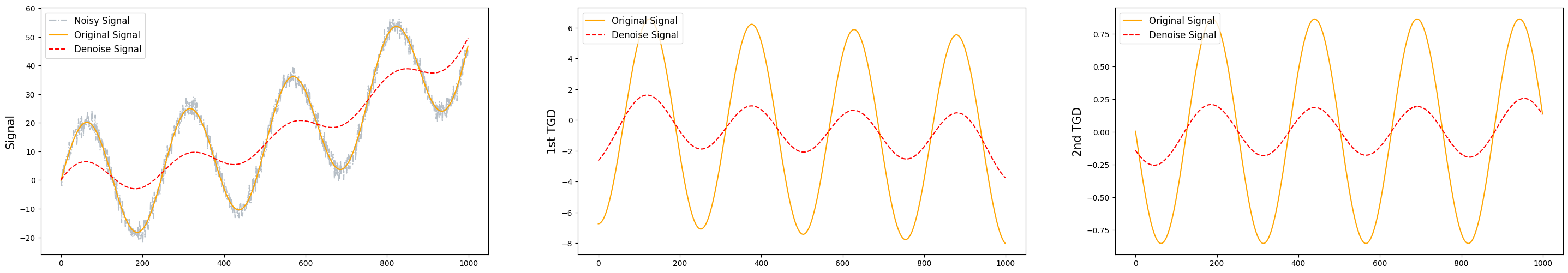}
    } 
    \subfigure[$\lambda_{\text{1st}} = 1$, $\lambda_{\text{2nd}} = 10$ and $\lambda_{\text{offset}} = 1$ (Signal RMSE = 0.4377)]{   	 		 
        \includegraphics[width=0.98\linewidth]{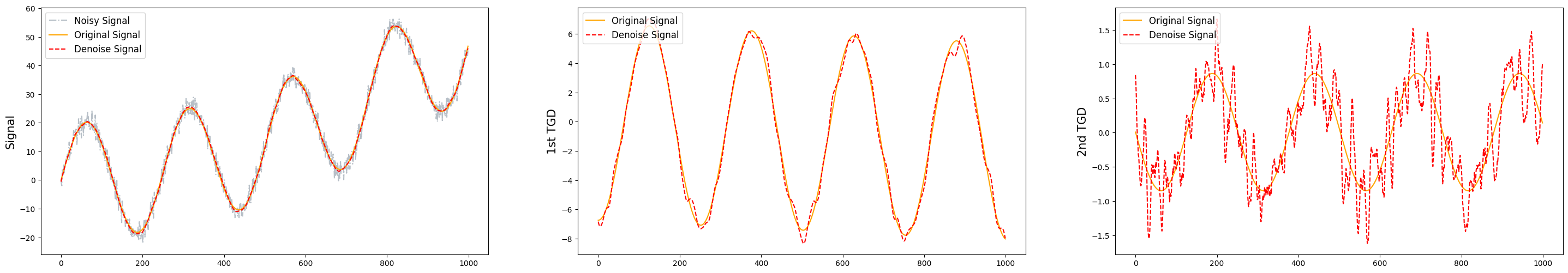}
    }  
    \caption{Comparison of the smoothing effect with different loss factors. The results show that simultaneous consideration of the complete loss is beneficial for denoising.}  
    \label{fig:TDGDenoiseLossFactor} 
\end{figure}

Next, we do the ablation analysis for loss factors. Figure~\ref{fig:TDGDenoiseLossFactor} illustrates the denoising effect of various loss factors applied to noisy signal $X_1$. The first three rows compare the degree of continuity constraints. For $\lambda_{\text{1st}} > 0$ and $\lambda_{\text{2nd}} = 0$, the first-order TGD of the output signal is continuous, but the second-order TGD fluctuates drastically. Conversely, for $\lambda_{\text{2nd}} > 0$, the output signals' second-order TGDs are continuous, and resulting denoised signals resemble a $C^2$ function. In such cases, it is preferable to apply continuity constraints to both first- and second-order TGDs, rather than just the latter. The final two lines demonstrate the impact of varying $\lambda_{\text{offset}}$. For $\lambda_{\text{offset}} = 0$, no correlation is needed between the denoised signal and the noisy signal, resulting in severe distortion. At this stage, the optimal solution is for both first- and second-order TGD to be constant values, leading to a constant or linear denoised signal. When $\lambda_{\text{offset}}$ is relatively large, the denoised signal fits the noisy signal as closely as possible, making it hard to maintain derivative continuity in a high-noise setting. Hence, $\lambda_{\text{offset}}$ should be decreased accordingly when the SNR is low.

\begin{figure}[!htb]    	
    \centering    	
    \subfigure[Noisy signal $X_1$]{  			 
        \includegraphics[width=0.98\linewidth]{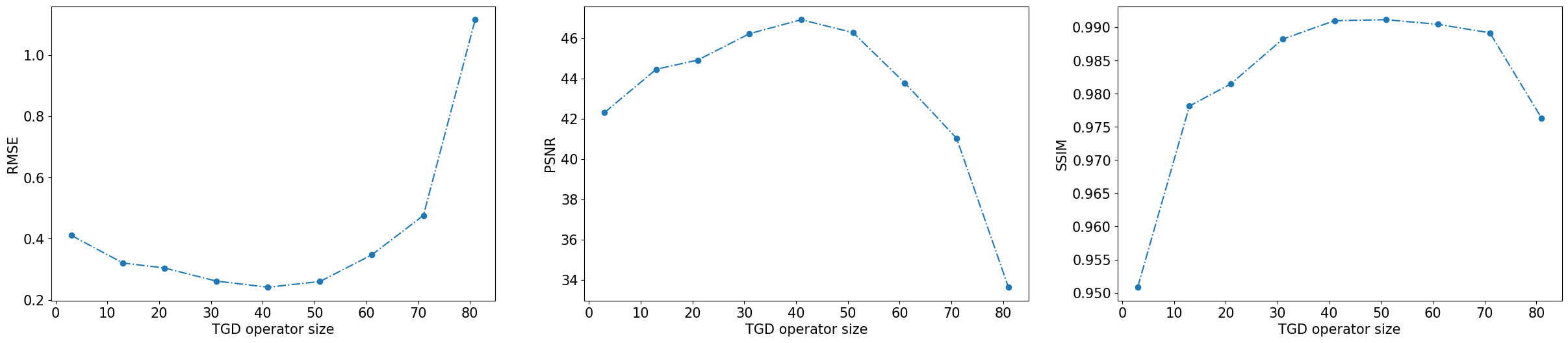}
    } 
    \subfigure[Noisy signal $X_2$]{   	 		 
        \includegraphics[width=0.98\linewidth]{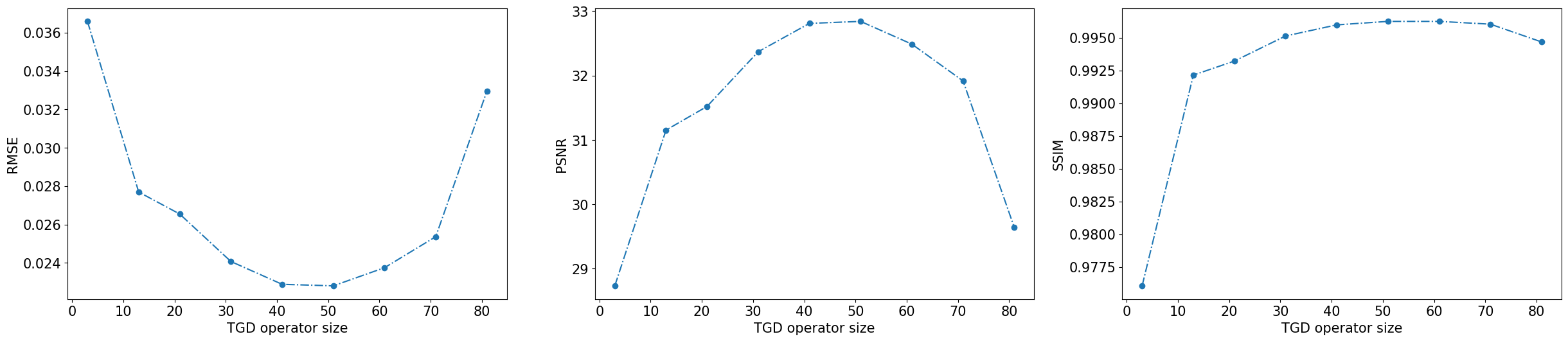}
    }  
    \caption{Comparison of the smoothing effect with various TGD operator sizes, from $3$ to $81$. It can be seen that the denoising effect gets better and then worse as the operator size increases, and the smoothing effect is satisfactory when the operator size is around $51$.}  
    \label{fig:TDGDenoiseKernelsize} 
\end{figure}

Lastly, we examine the influence of TGD operator size on smoothing performance. When the operator size is relatively small, the difference computation becomes highly sensitive to noise. The corresponding difference sequence is severely distorted, resulting in poor smoothing. Moreover, increasing the operator size is not always feasible since it tends to lead to a global operation. This, in turn, can distort the signal if it results in excessive smoothness. Additionally, long padding causes spurious patterns of variation at both ends of the signal, which is undesirable. These align with the experimental results depicted in Figure~\ref{fig:TDGDenoiseKernelsize}, where the TGD-based denoise algorithm's performance initially improves as the operator size increases but then deteriorates. Optimal overall performance is achieved when the operator size is approximately $51$. If the signal sequence length is short, the TGD convolution kernel size should be correspondingly reduced.

\clearpage
\newpage

\section{2D TGD-Based image edge detection}

In computer vision, the edge in an image is defined as the center of local intensity gradient. The classic edge detection method is proposed by John Canny~\cite{DBLP:journals/pami/Canny86a}, which consists of four stages in OpenCV\footnote{\url{https://docs.opencv.org/master/d7/de1/tutorial_js_canny.html}}: Noise reduction via Gaussian smoothing, Finding the intensity gradient of the image via Sobel operators, Non-maximum Suppression, and Hysteresis Thresholding (or called Double-Threshold Selection). The TGD theory reveals that the Gaussian smoothing operator violates the Monotonic Convexity Constraint (C3), thereby the first-order Gaussian operator, which also violates the Monotonic Constraint (C2), is not the general difference operator for discrete sequences or arrays. The detection results show the drift phenomena in edge location~\cite{gunn1998edge,gunn1999discrete}, particularly with large kernel sizes ($\geq 13\times13$). This drift becomes more pronounced as the kernel size increases.  
To address this issue, we introduce TGD into image gradient computation and propose first- and second-order TGD-based edge detection algorithms, with reference to classical approaches.


\subsection{First-order TGD-based Edge Detection}

\begin{figure}[htb]
    \centering
    \begin{minipage}[b]{0.9\linewidth}
        \centering
        \centerline{\includegraphics[width=\linewidth]{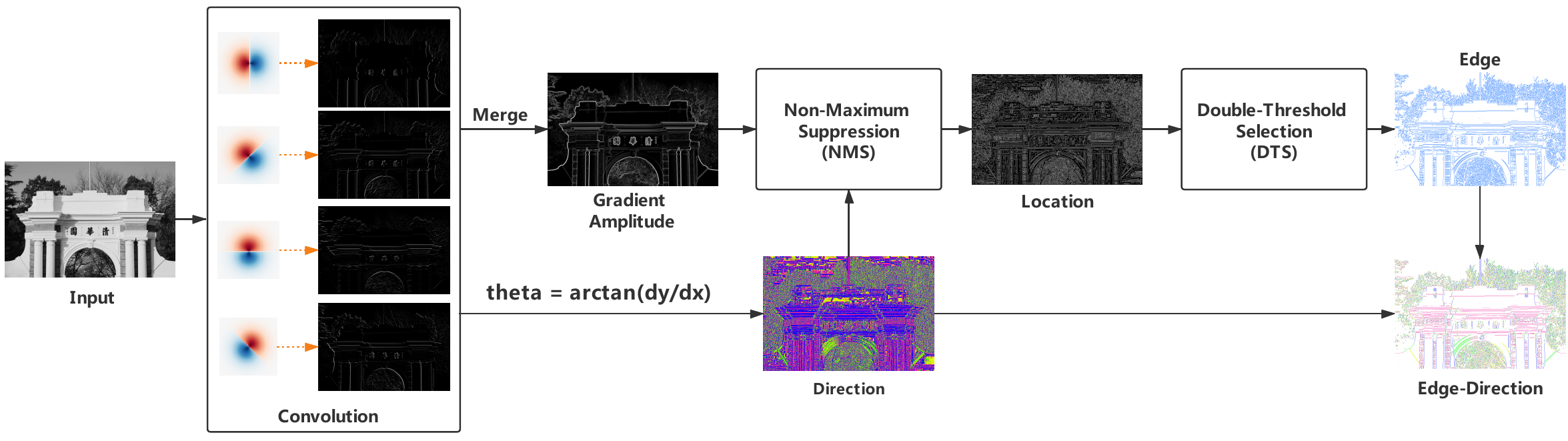}}
    \end{minipage}
    \caption{
    Flowchart of edge detection algorithm based on first-order TGD.
  }
  \label{fig:Algorithm1}
\end{figure}

As the first-order TGD operators are naturally robust to noise in the difference calculation, we replace the first two stages of the Canny algorithm with a single TGD convolution stage to form the three-stage first-order TGD-based edge detection algorithm. Figure~\ref{fig:Algorithm1} shows the whole algorithm flowchart, and the pseudo-code is presented in Algorithm~\ref{algorithm:Algorithm1}. 

In the subsequent experiments, we utilize the 2D first-order discrete TGD operators obtained by the rotational construction method (Section 2.3.1 in TGD Theory). 
The kernel function employs the exponential function (Section 4.1.2 in TGD Theory) and the cosine function is used for the rotation weights. For a fair comparison, we maintain a consistent size for the convolution kernel in both the Canny and TGD-based edge detection algorithms, while leaving the other aspects of the algorithms unaltered.

\begin{algorithm}[htb]
  \caption{First-order TGD-based Edge Detection Algorithm}
  \hspace*{0.02in} {\bf Input:} 
    Gray image $I$, first-order discrete TGD operators in four directions $\widehat{T}_{0'}, \widehat{T}_{45'}, \widehat{T}_{90'}, \widehat{T}_{135'}$, high threshold $highThr$ and low threshold $lowThr$ for Double-threshold Selection.\\
  \hspace*{0.02in} {\bf Output:} 
    Edge position $E$ and edge orientation $D$
  \begin{algorithmic}[1]
    \STATE $dx = I * {\widehat{T}}_{0'}$
    \STATE $d_{45'} = I * {\widehat{T}}_{45'}$
    \STATE $dy = I * {\widehat{T}}_{90'}$
    \STATE $d_{135'} = I * {\widehat{T}}_{135'}$
    \STATE $Grad = \left(\sqrt{dx^2 + dy^2}+\sqrt{d_{45'}^2 + d_{135'}^2}\right) / 2$
    \STATE $\Theta = \arctan(dy / dx)$
    \STATE $NMS = \text{Non-Maximum Suppression}(Grad, \Theta)$
    \STATE $E = \text{Double-Threshold Selection}(NMS, lowThr, highThr)$
    \STATE $D = \text{Matrix Element Multiplication}(E, \Theta)$
    \RETURN $E$, $D$
  \end{algorithmic}
  \label{algorithm:Algorithm1}
\end{algorithm}

\begin{figure}[htbp]    	
    \centering 
    \subfigure[Nature Image]{  			 
         \centerline{\includegraphics[width=0.9\linewidth]{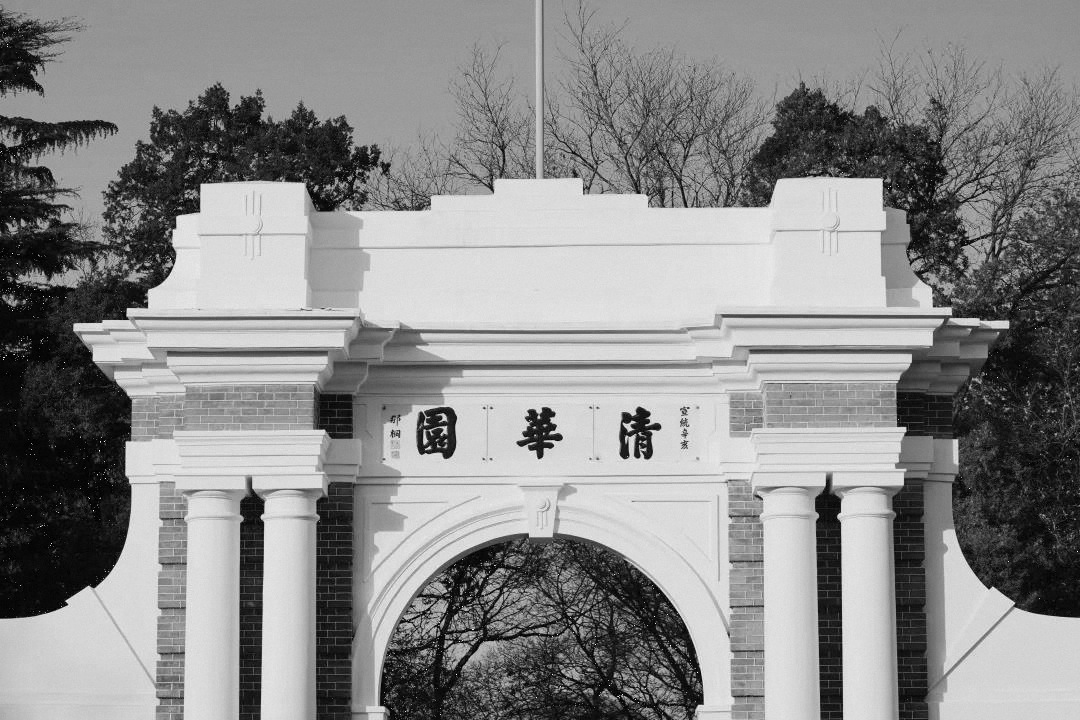}
        }
    }
    \subfigure[Canny $13\times13$]{  			 
        \includegraphics[width=0.44\linewidth]{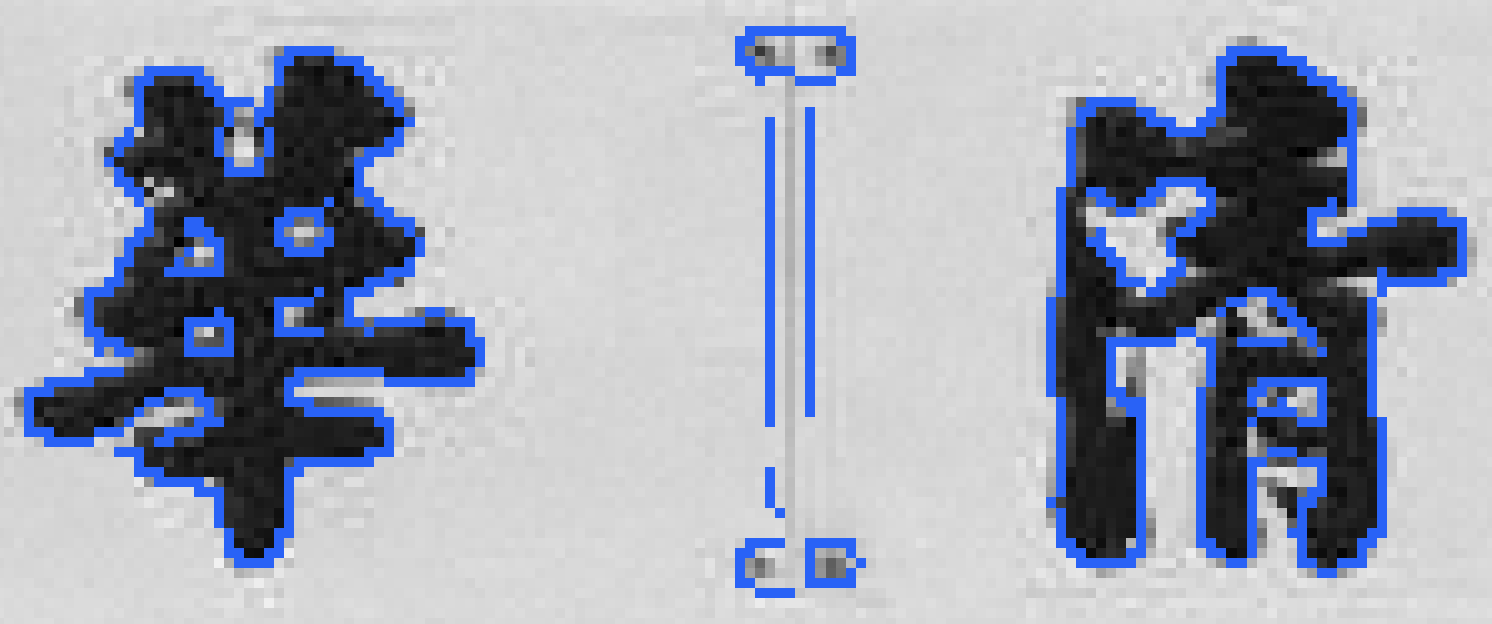}
    }    	 
    \subfigure[Canny $17\times17$]{   	 		 
        \includegraphics[width=0.44\linewidth]{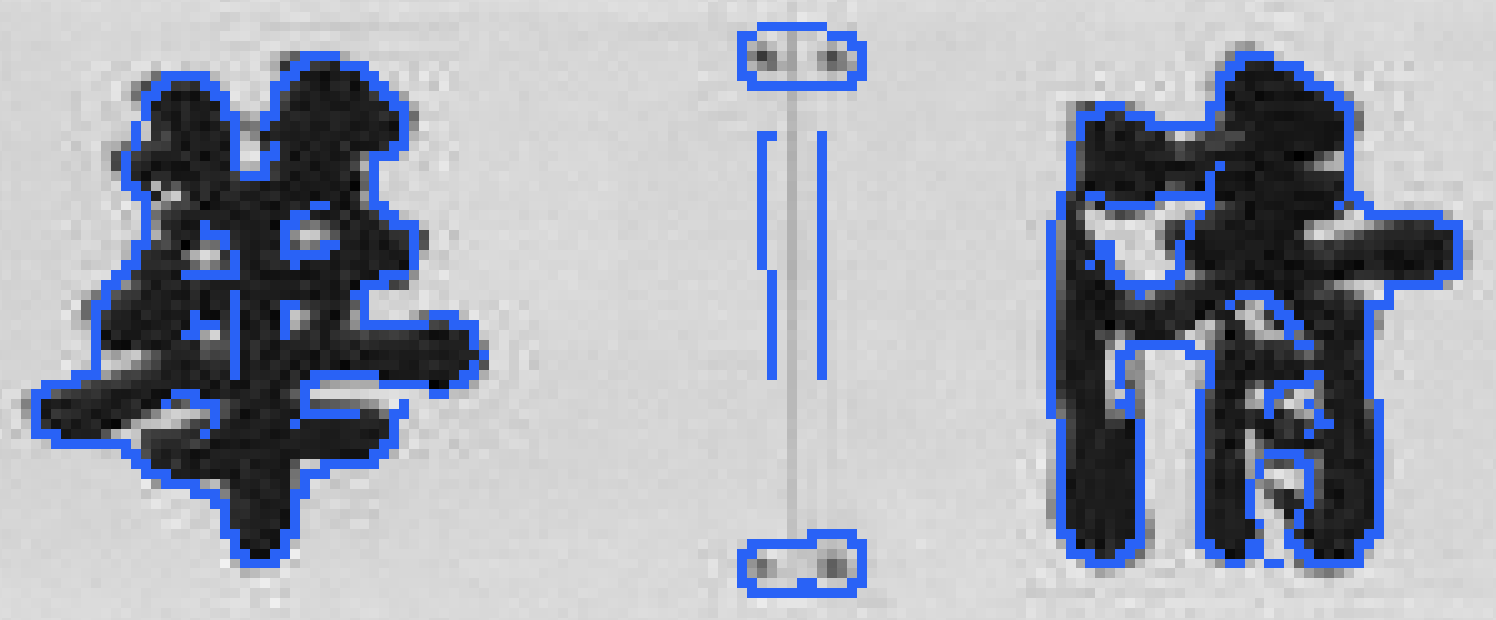}
    }
    \subfigure[TGD $13\times13$]{  			 
        \includegraphics[width=0.44\linewidth]{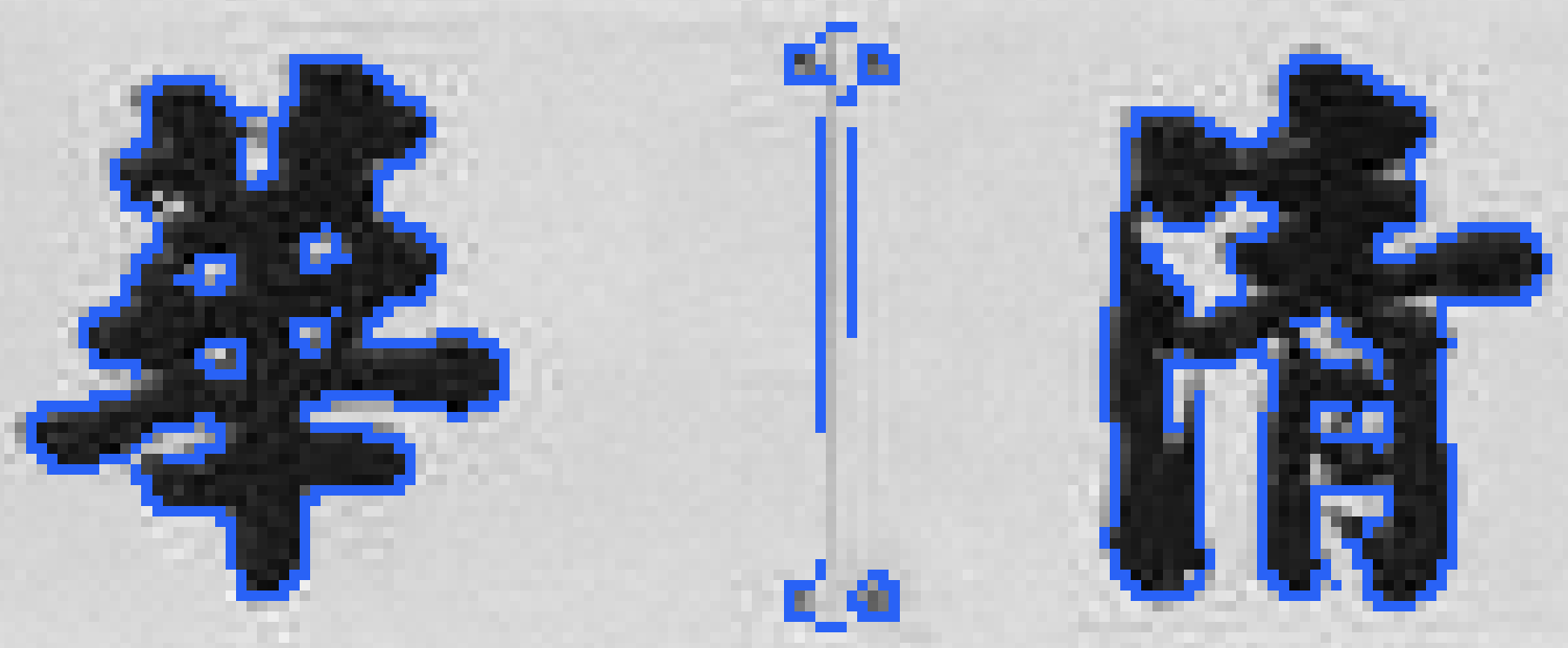}
    }
    \subfigure[TGD $17\times17$]{  			 
        \includegraphics[width=0.44\linewidth]{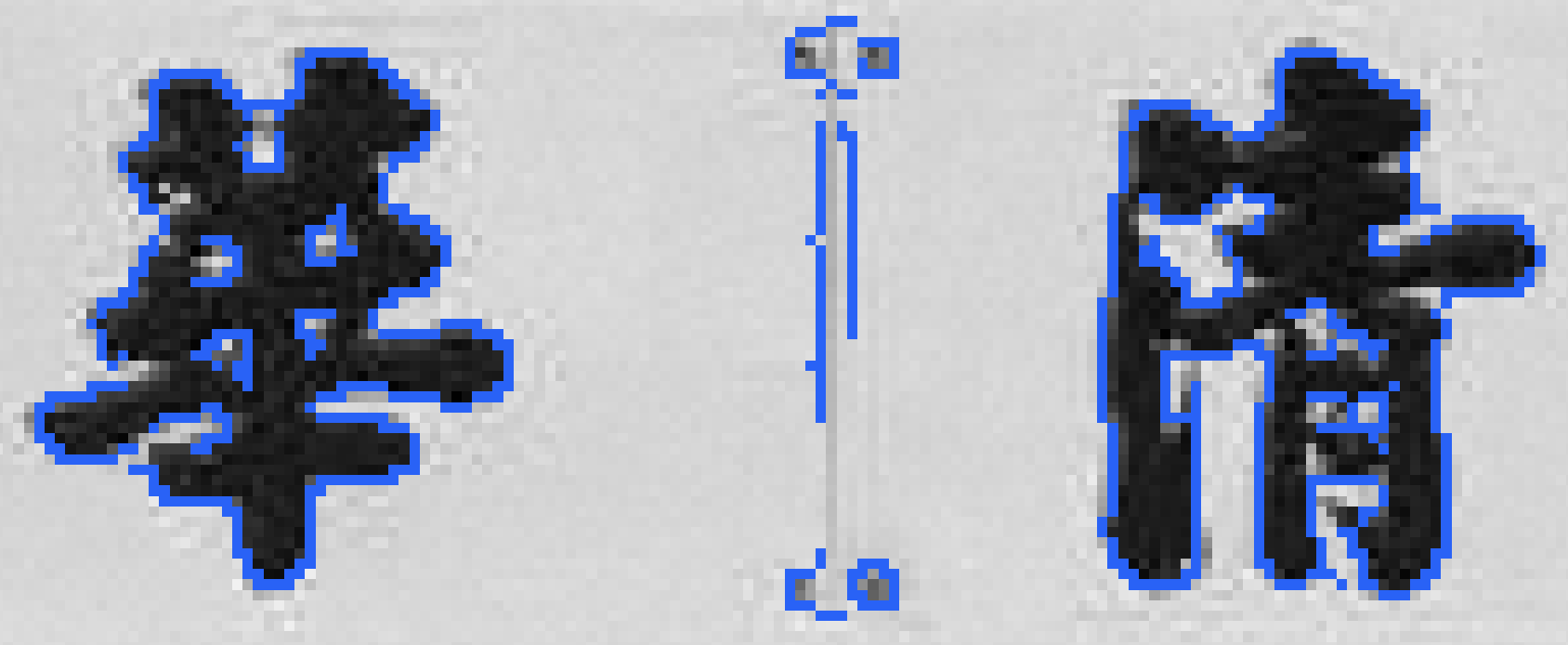}
    }
    \caption{Edge detection results: (a) original nature image; (b - e) detected edges by Canny and TGD algorithms, in which the edges detected by Canny algorithm have one-pixel drift at the kernel size $13\times13$, and two pixels drift at the kernel size $17\times17$.}  
    \label{fig:example-tsinghuaResult1} 
\end{figure}

\begin{figure}[htbp]    	
    \centering    	
    \subfigure[Texture Image]{  			 
        \centerline{\includegraphics[width=0.9\linewidth]{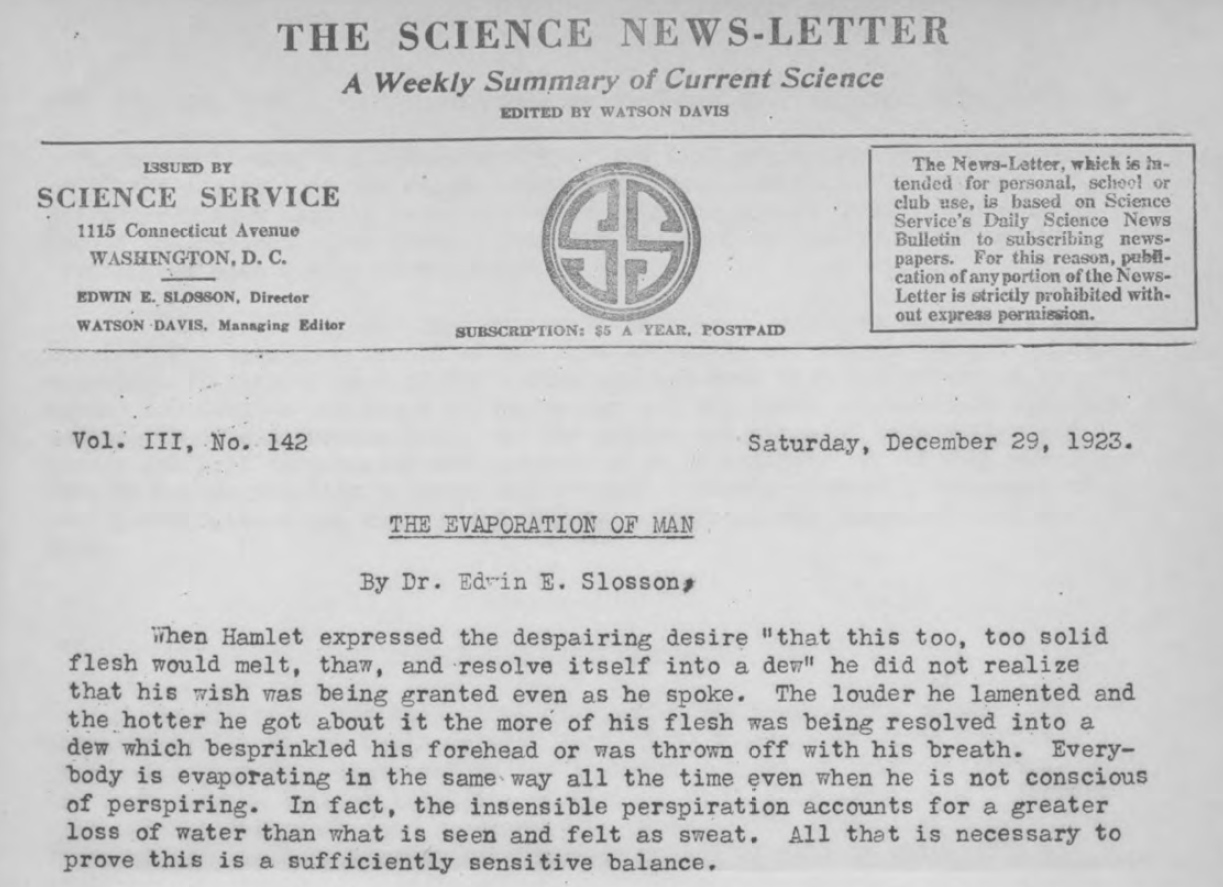}
        }
    }
    \subfigure[Canny $13\times13$]{  			 
        \includegraphics[width=0.45\linewidth]{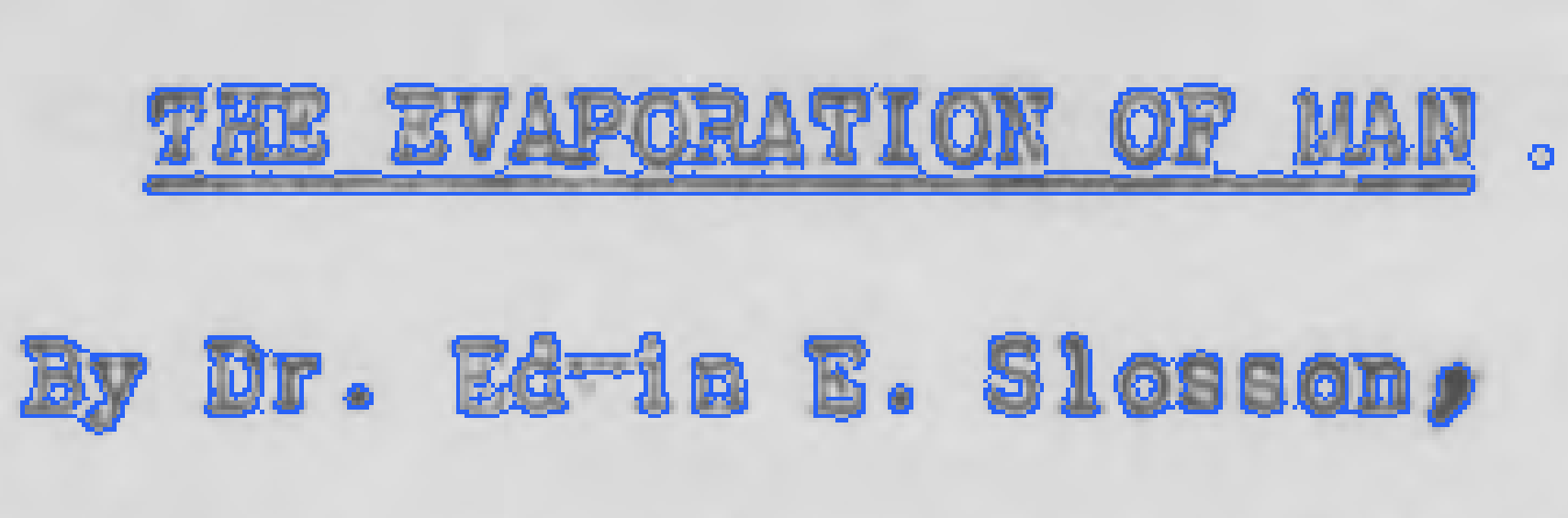}
    }    
    \subfigure[Canny $17\times17$]{  			 
        \includegraphics[width=0.45\linewidth]{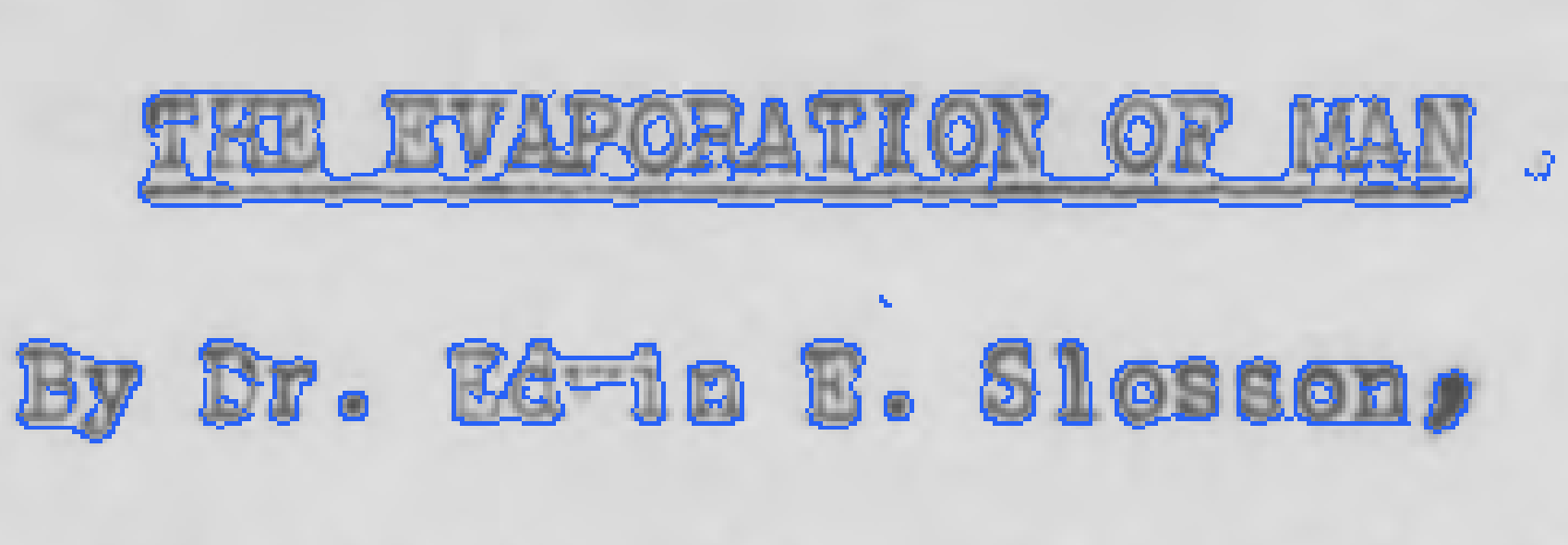}
    }  
    \subfigure[TGD $13\times13$]{   	 		 
        \includegraphics[width=0.45\linewidth]{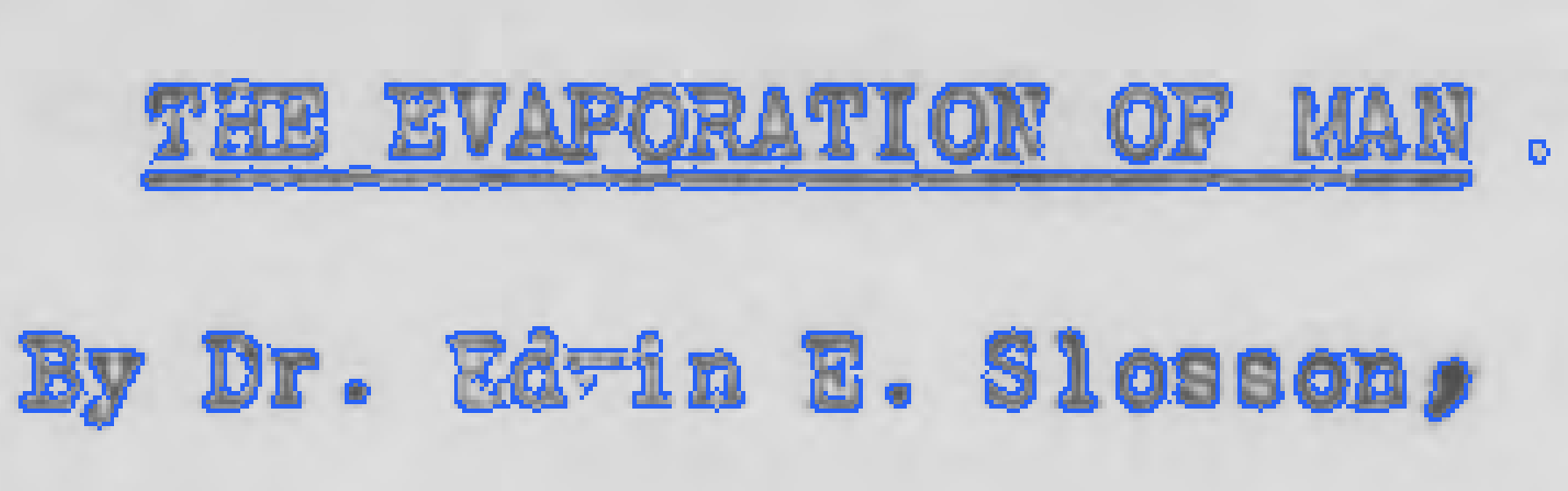}
    }
    \subfigure[TGD $17\times17$]{   	 		 
        \includegraphics[width=0.45\linewidth]{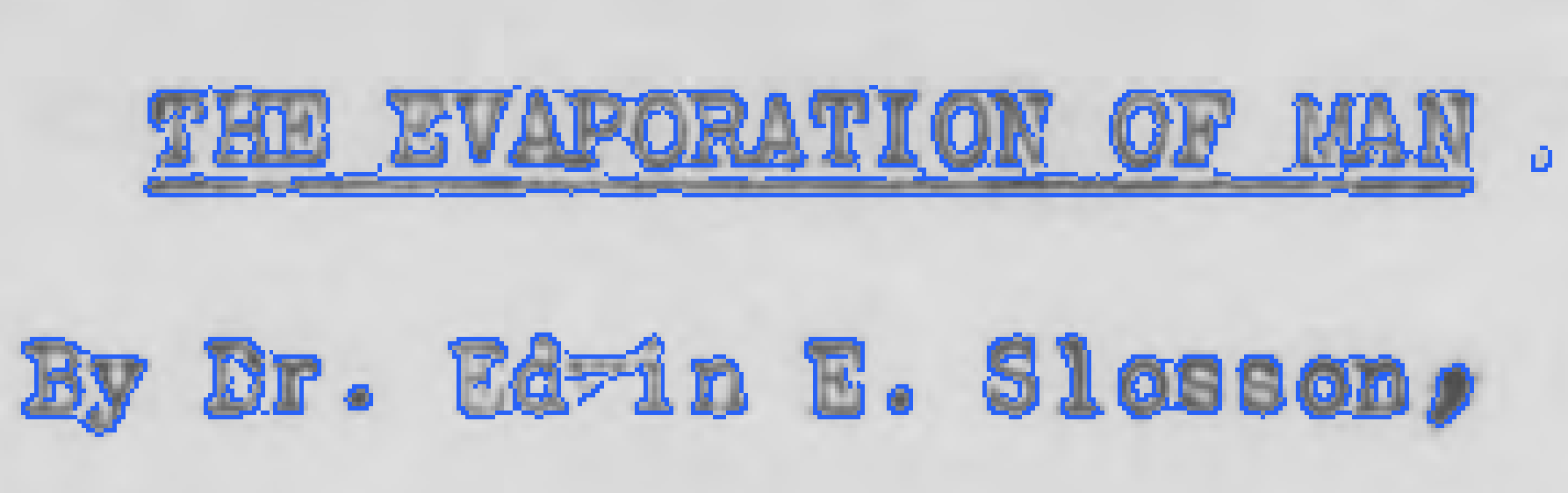}
    }
    \caption{Edge detection results: (a) original texture image; (b - e) detected edges by Canny and TGD algorithms, in which the edges detected by Canny algorithm have one-pixel drift at the kernel size $13\times13$, and two pixels drift at the kernel size $17\times17$.}  
    \label{fig:example-englishResult1} 
\end{figure}
\clearpage

In the theoretical analysis, 
we point out that Gaussian smooth filter fails to satisfy \emph{C3: Monotonic Convexity Constraint}, and produces drifting localization of detected edges with increasing kernel size. This is consistent with the experimental results shown in Figure~\ref{fig:example-tsinghuaResult1} and~\ref{fig:example-englishResult1}, in which the local details of edge localization detected by Canny operator have drifted with one to two pixels. In contrast, the first-order TGD operators, employing the same kernel size as that of Canny, enable accurate localization with good noise reduction performance.

\begin{figure}[htbp]    	
    \centering    	
    \subfigure[Gaussian-Sobel]{  			 
        \includegraphics[width=0.48\linewidth]{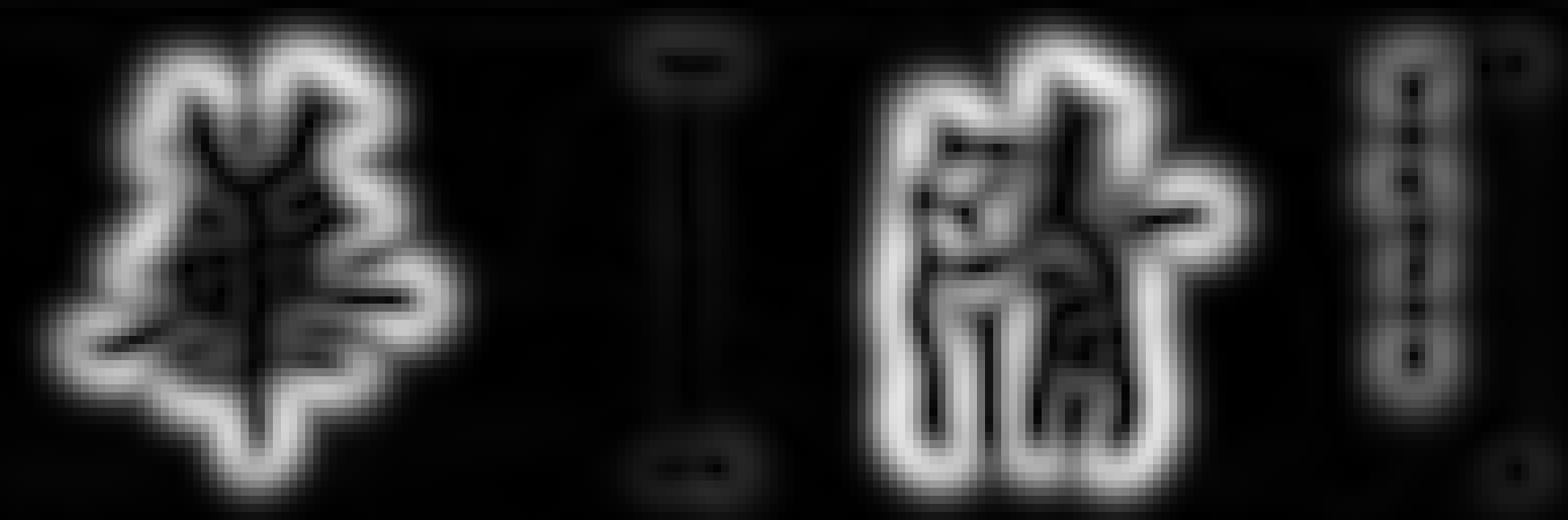}
    }
    \subfigure[TGD]{  			 
        \includegraphics[width=0.48\linewidth]{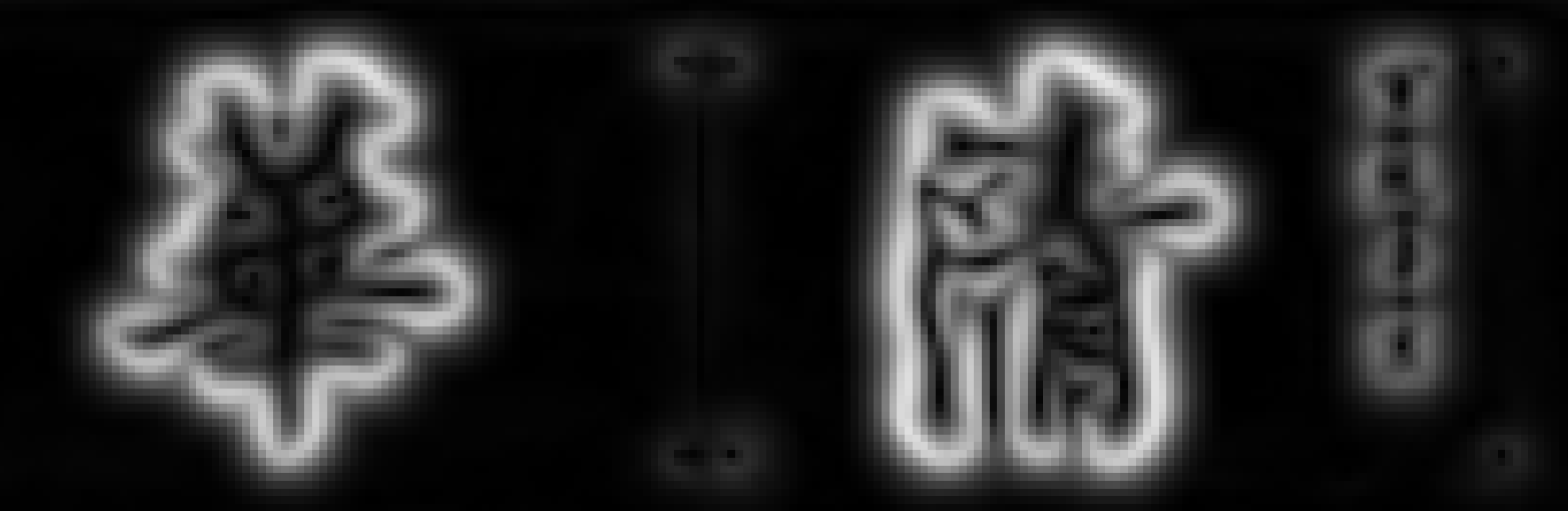}
    } 
    \subfigure[Gaussian-Sobel]{   	 		 
        \includegraphics[width=0.48\linewidth]{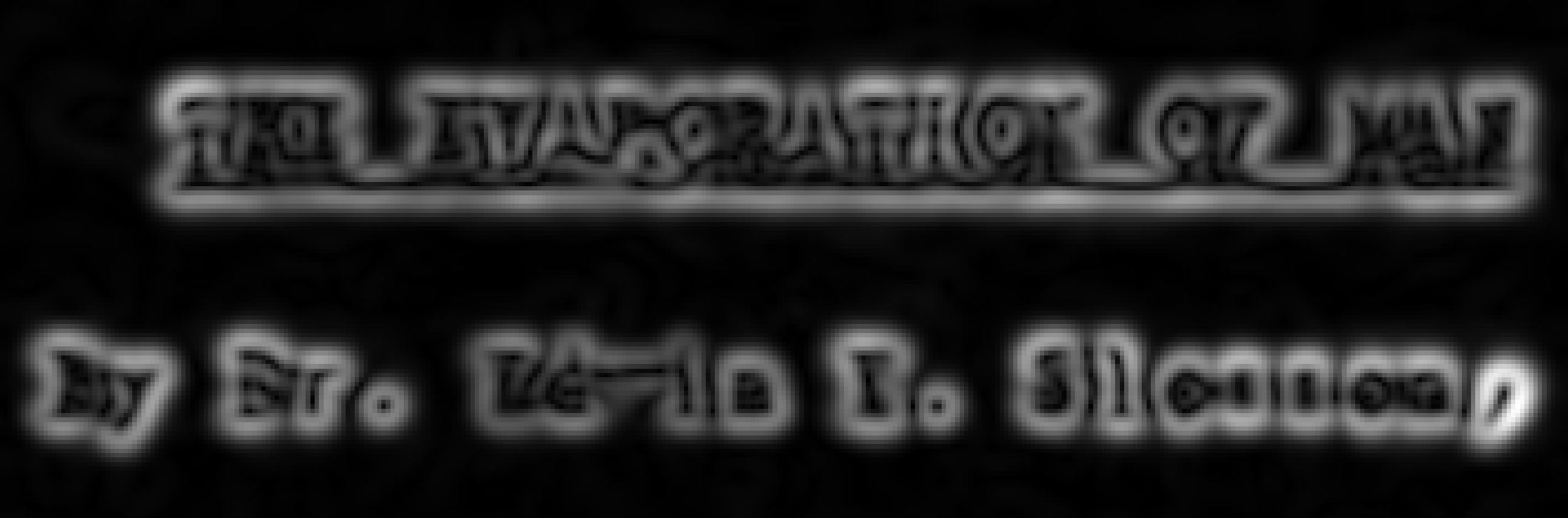}
    }   
    \subfigure[TGD]{   	 		 
        \includegraphics[width=0.48\linewidth]{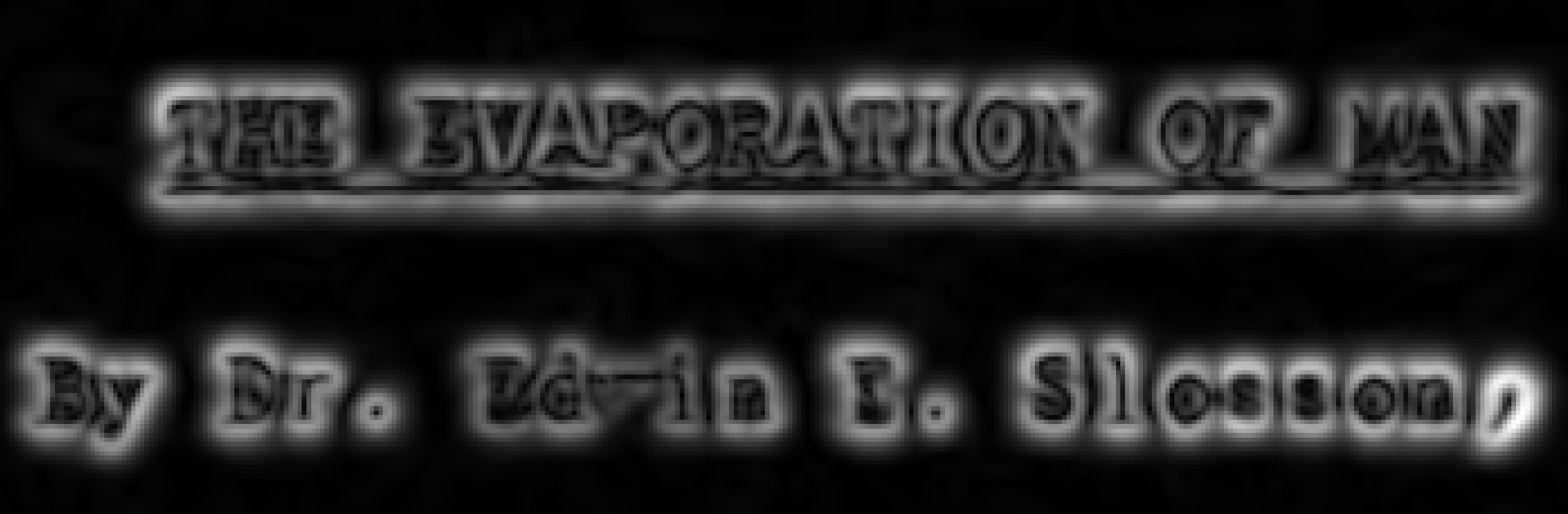}
    }
    \caption{Comparison of gradient magnitudes in detail: (a,c) plot of gradient magnitude computed by $17\times17$ first-order Gaussian-Sobel derivative operators; (b,d) plot of gradient magnitude computed by $17\times17$ first-order TGD operators. The gradient magnitude values are positive and normalized, and the whiter the color in the visualization represents the larger the value.} 
    \label{fig:example-convResult1} 
\end{figure}

\begin{figure}[htbp]
    \centering    			 
    \includegraphics[width=0.98\linewidth]{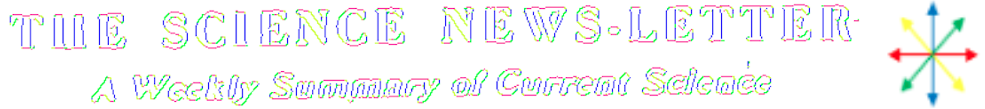}
    \caption{Edge orientation map obtained by $17\times17$ first-order TGD operators in four directions, which are indicated in color arrows.}  
    \label{fig:example-direction} 
\end{figure}

The edge refers to the location map of the local maximum gradients. Figure~\ref{fig:example-convResult1} provides a comparison of the gradient amplitudes obtained through first-order Gaussian derivative operators in Canny and first-order TGD operators. Notably, there is significant outward expansion of the local extremes in the gradient amplitude map calculated by the first-order Gaussian derivative operators. This expansion is detrimental to accurate recognition of Chinese characters (Figure~\ref{fig:example-convResult1}.a) and English letters (Figure~\ref{fig:example-convResult1}.c). In contrast, the results yielded via the first-order TGD operators are reliable, and the local gradient extremes are concentrated toward the edges and precisely located. Such reliable and accurate gradient information is crucial for downstream tasks, as it can serve as an essential image feature beyond edge detection. Benefiting from the orientation sensitivity of the TGD operator, our algorithm also outputs the edge orientation map of the image (Figure~\ref{fig:example-direction}).

\subsection{Second-order TGD-based Edge Detection}
\begin{figure}[!htb]    	
    \centering    	
    \subfigure[$\widehat{T}_{x}$]{  			 
        \includegraphics[width=0.36\linewidth]{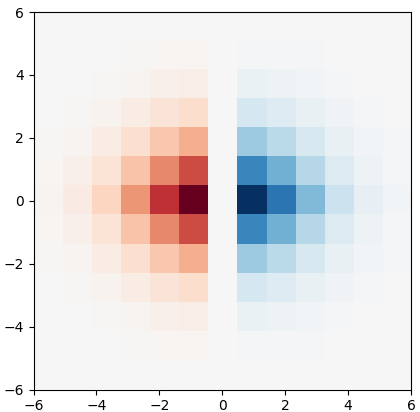}
    }\hspace{10mm}    	 
    \subfigure[$\widehat{R}_{x}$]{   	 		 
        \includegraphics[width=0.36\linewidth]{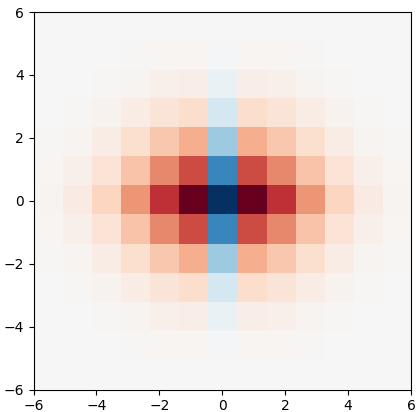}
    }    	 
    \subfigure[$\widehat{T}_{45'}$]{   	 		 
        \includegraphics[width=0.36\linewidth]{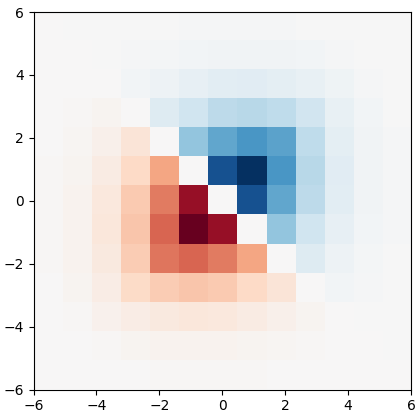}
    }\hspace{10mm}  
    \subfigure[$\widehat{R}_{45'}$]{  			 
        \includegraphics[width=0.36\linewidth]{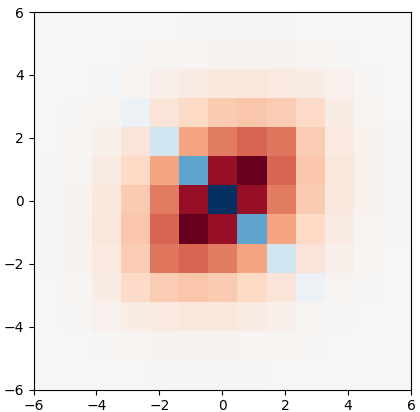}
    }   
    \subfigure[$\widehat{\text{LoT}}$]{  			 
        \includegraphics[width=0.36\linewidth]{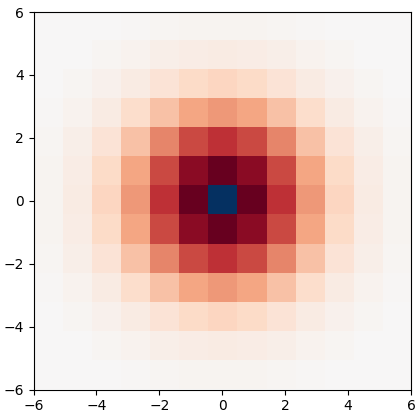}
    } 
    \caption{Visualization of a 2D discrete TGD operators. The operator values can be found in Formula in the theory part. 
    }  
    \label{fig:DTGD_2D_examples} 
\end{figure}

\begin{figure}[htb]
  \begin{minipage}[b]{\linewidth}
    \centering
    \centerline{\includegraphics[width=0.9\linewidth]{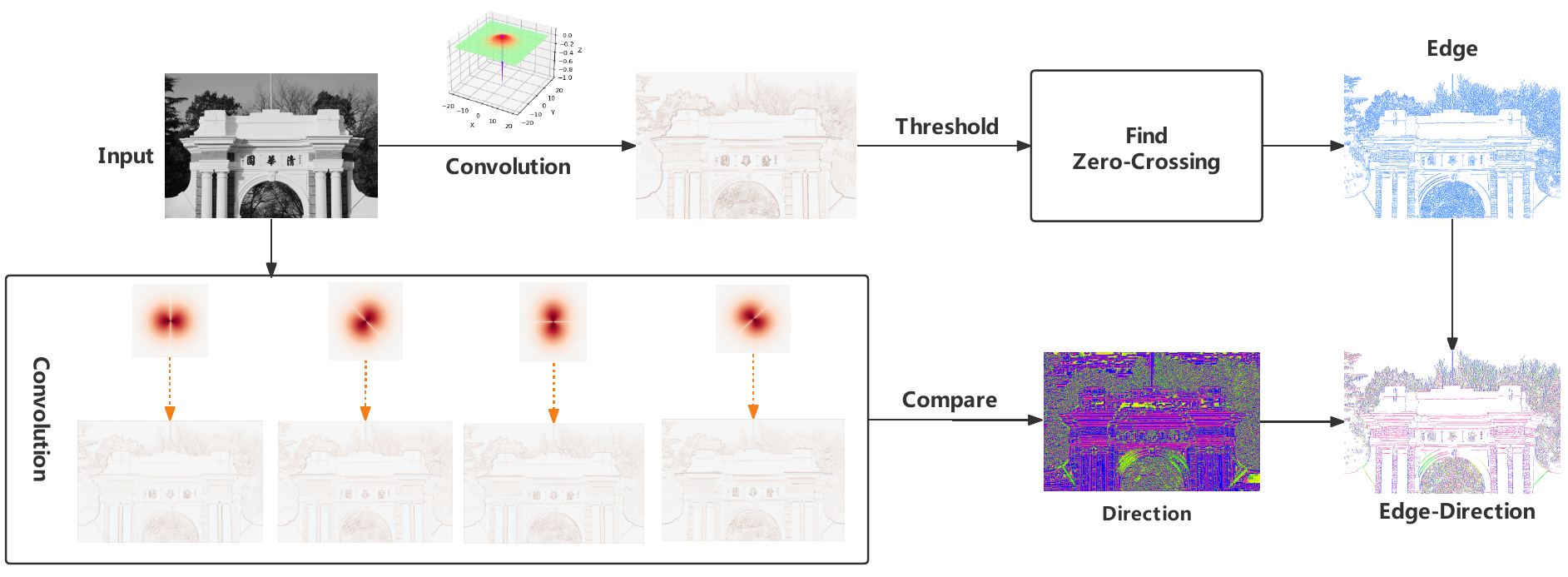}}
  \end{minipage}
  \caption{
    Flowchart of edge detection algorithm based on second-order TGD.
  }
  \label{fig:Algorithm2}
\end{figure}

The edge detection algorithm based on the LoT operator follows the traditional two-stage LoG-based method, where the LoG operator is replaced by the directional LoT operator (Section 2.3.1 in TGD Theory).  
Since the edge orientation is orthogonal to the gradient direction, which has the largest response, the edge orientation can be determined by comparing the response of multiple second-order directional TGD operators (Figure~\ref{fig:DTGD_2D_examples}). However, due to the isotropism of LoG, the edge orientation cannot be obtained as in the gradient-based method (Figure~\ref{fig:example-direction}). Algorithm~\ref{algorithm:Algorithm2} presents the pseudo-code of the LoT-based edge detection algorithm, and Figure~\ref{fig:Algorithm2} shows the whole algorithm flowchart. In the subsequent experiments, the Gaussian kernel function is used to construct the LoT operator.

\clearpage

\begin{algorithm}[htbp]
  \caption{Second-order TGD-based Edge Detection Algorithm}
  \hspace*{0.02in} {\bf Input:} 
    Gray image $I$, second-order discrete TGD operators in four directions $\widehat{R}_{0'}, \widehat{R}_{45'}, \widehat{R}_{90'}, \widehat{R}_{135'}$, LoT operator $\widehat{R}_{\text{Laplacian}}$, Laplacian threshold $LoTThr$.\\
  \hspace*{0.02in} {\bf Output:} 
    Edge position $E$ and edge orientation $D$
  \begin{algorithmic}[1]
  \STATE $Grad_L = I * {\widehat{R}}_{\text{Laplacian}}$
  \STATE $Grad_L[abs(Grad_L) < LoTThr] = 0$
  \STATE $d_{0'} = I * {\widehat{R}}_{0'}$
  \STATE $d_{45'} = I * {\widehat{R}}_{45'}$
  \STATE $d_{90'} = I * {\widehat{R}}_{90'}$
  \STATE $d_{135'} = I * {\widehat{R}}_{135'}$
  \STATE $\Theta(x,y) = \text{argmax}(d_{0'}, d_{45'}, d_{90'}, d_{135'})$
  \STATE $E = \text{Find Zero-Crossing}(Grad_L)$
  \STATE $D = \text{Matrix Element Multiplication}(E, \Theta)$
  \RETURN $E$, $D$
  \end{algorithmic}
  \label{algorithm:Algorithm2}
\end{algorithm}

\begin{figure}[htbp]
    \centering    	
    \subfigure[LoG $13\times13$]{  			 
        \includegraphics[width=0.47\linewidth]{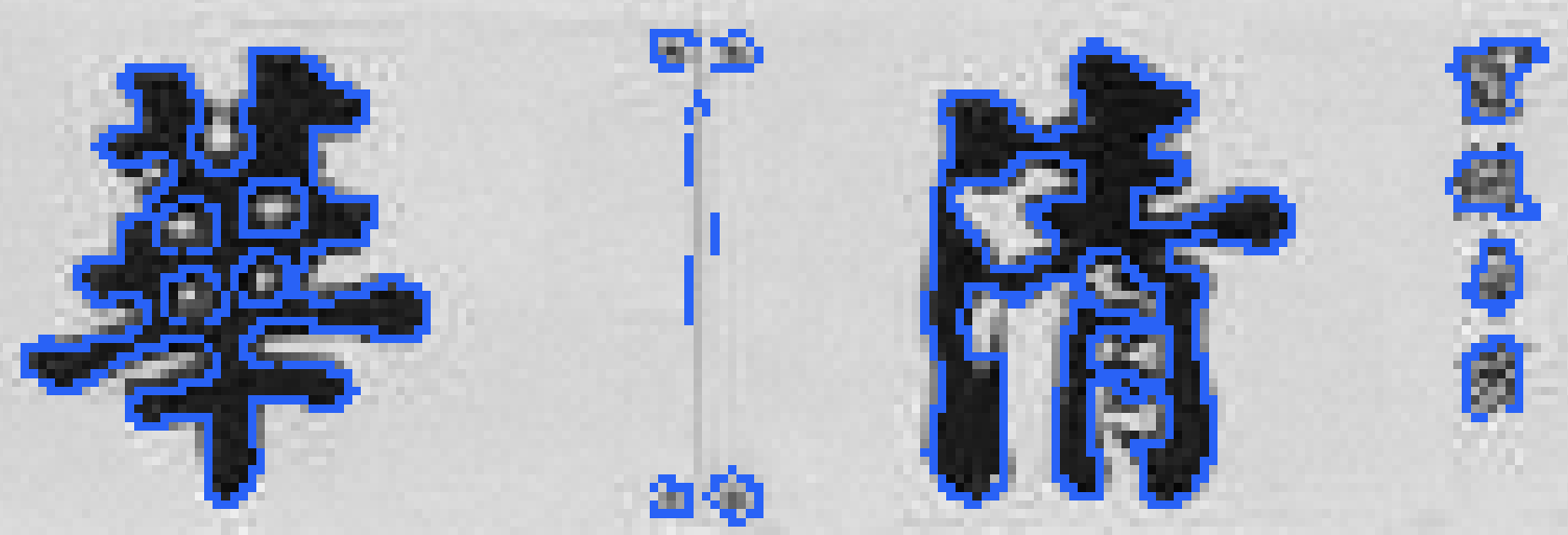}
    }    
    \subfigure[LoG $17\times17$]{  			 
        \includegraphics[width=0.47\linewidth]{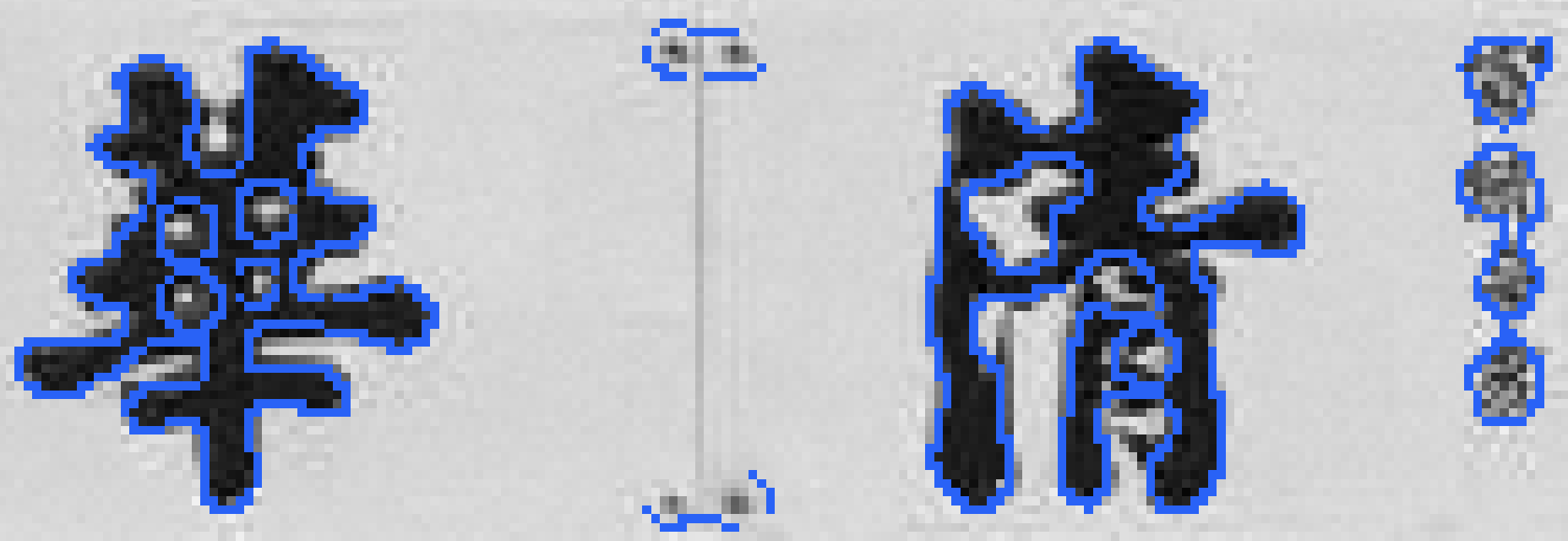}
    }  
    \subfigure[LoT $13\times13$]{   	 		 
        \includegraphics[width=0.47\linewidth]{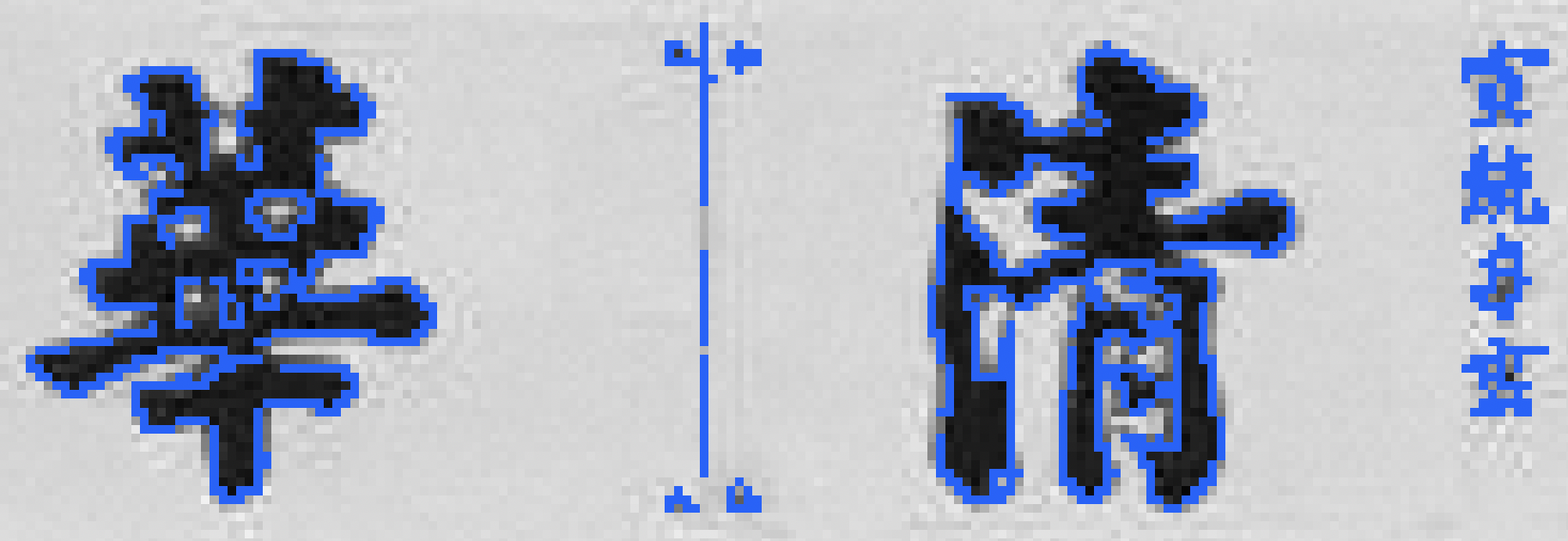}
    }
    \subfigure[LoT $17\times17$]{   	 		 
        \includegraphics[width=0.47\linewidth]{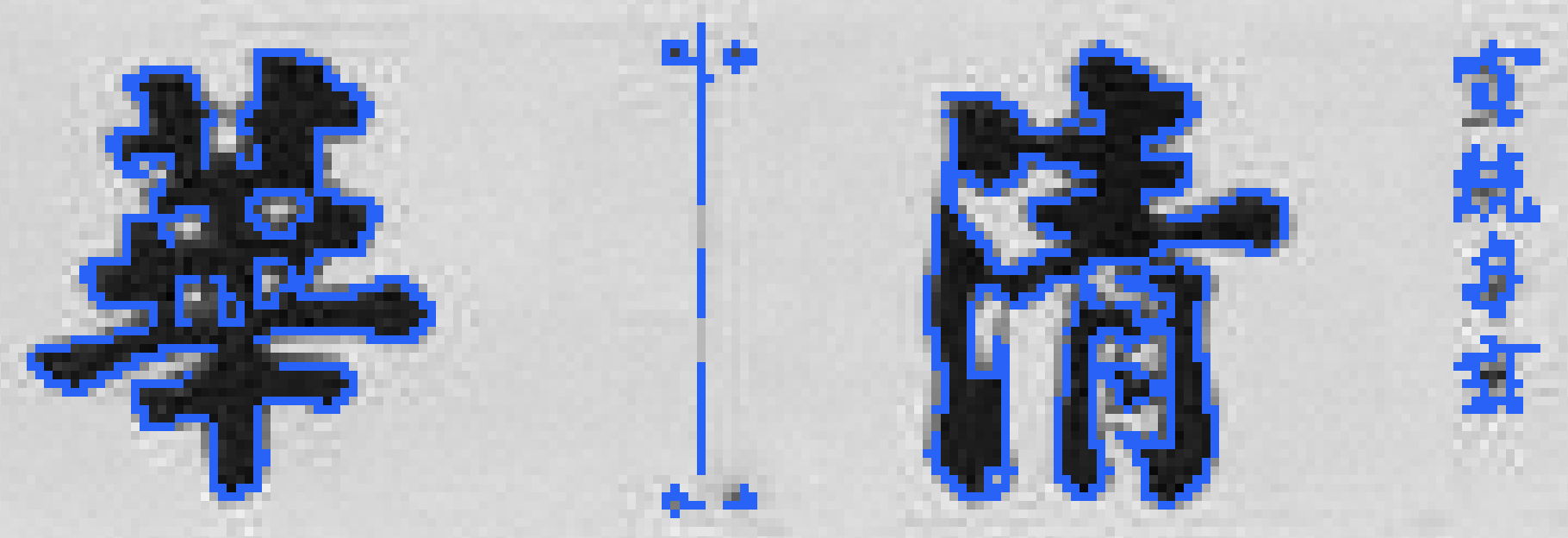}
    }
    \caption{Edge detection: the local detection details of the LoG and LoT operators. The original natural image is shown in Figure \ref{fig:example-tsinghuaResult1}.}  
    \label{fig:example-tsinghuaResult2} 
\end{figure}

Figure~\ref{fig:example-tsinghuaResult2} presents the detection results by the $13\times13$ and $17\times17$ LoG and LoT operators for a natural image with localized edge details. Obviously, the LoG operator exhibits localization drift, whereas the LoT operator produces accurate localization for very fine details.

\begin{figure}[htbp]    	
    \centering    	
    \subfigure[LoG $13\times13$]{  			 
        \includegraphics[width=0.47\linewidth]{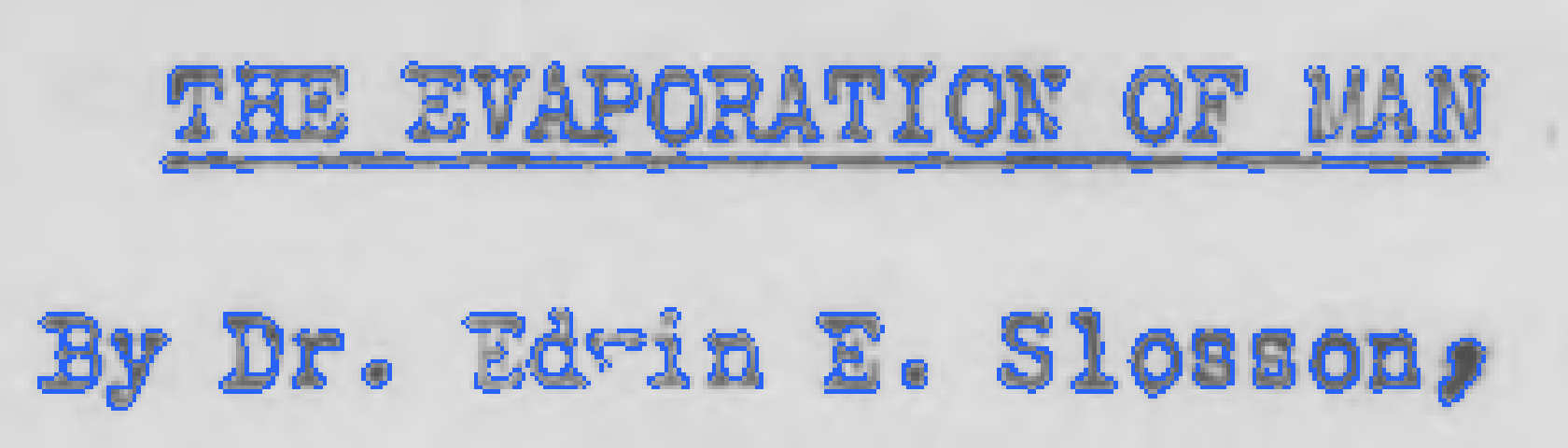}
    }    
    \subfigure[LoG $17\times17$]{  			 
        \includegraphics[width=0.47\linewidth]{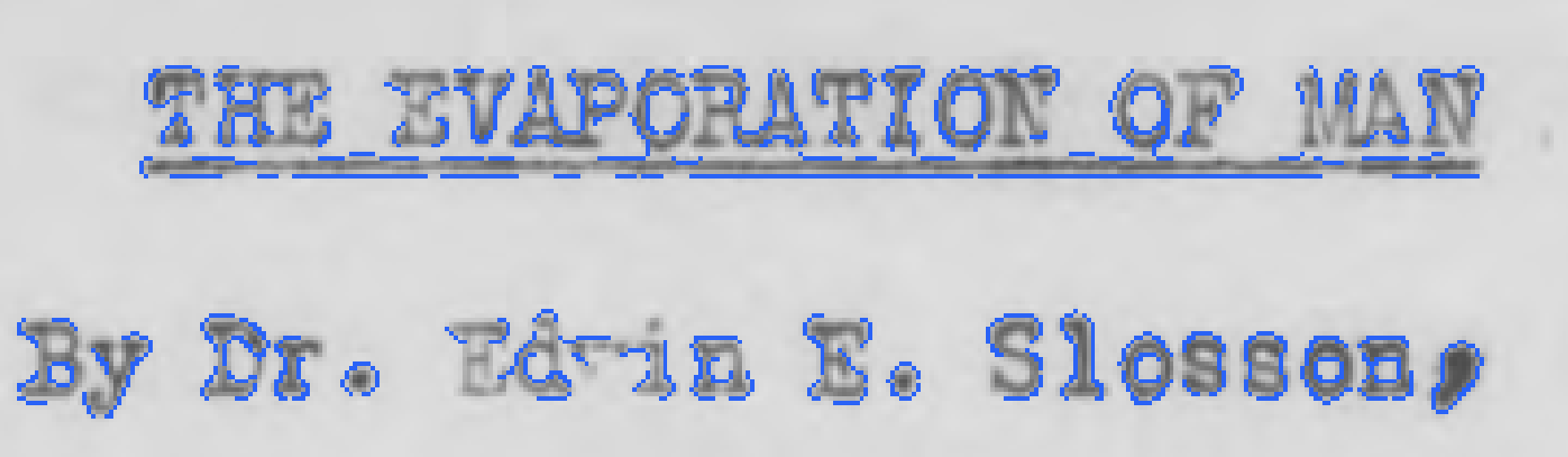}
    }  
    \subfigure[LoT $13\times13$]{   	 		 
        \includegraphics[width=0.47\linewidth]{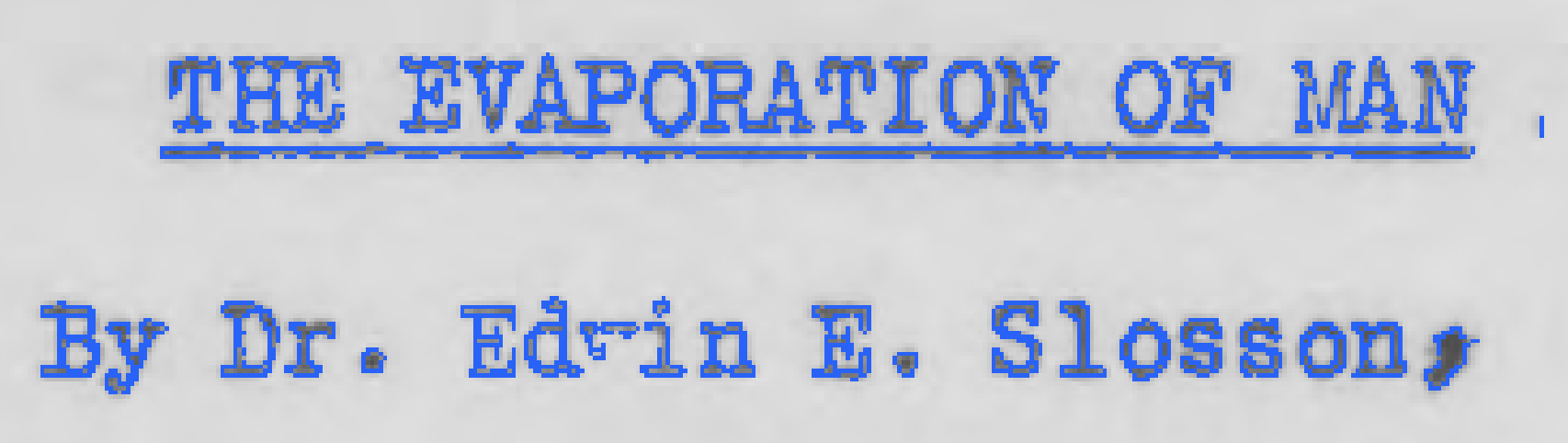}
    }
    \subfigure[LoT $17\times17$]{   	 		 
        \includegraphics[width=0.47\linewidth]{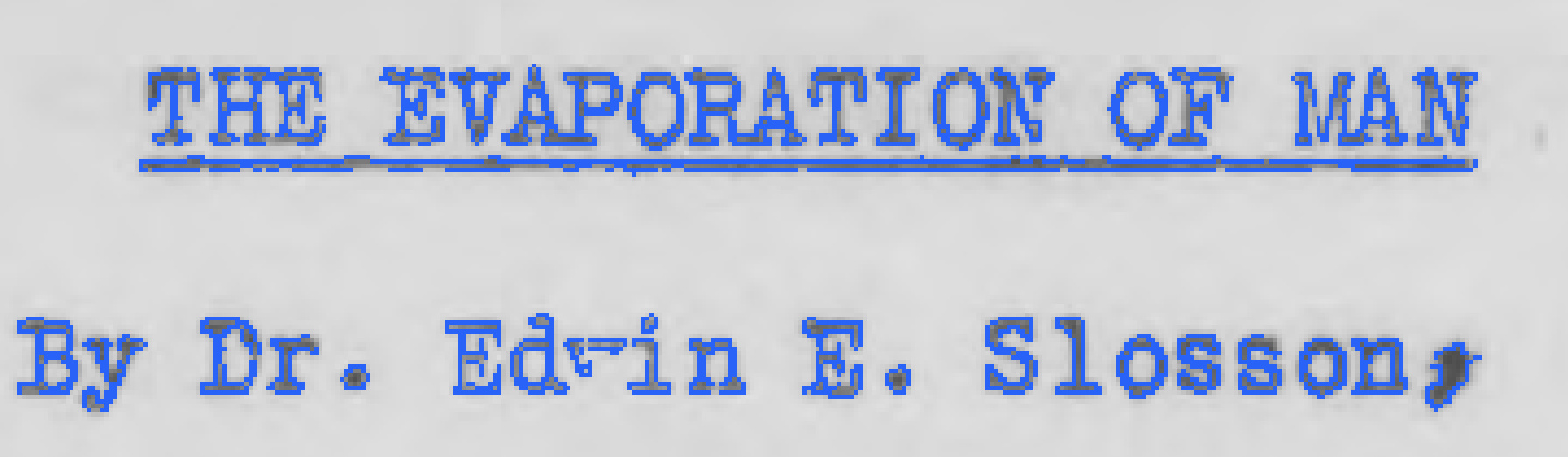}
    }
    \caption{Edge detection results: (a-d) the local detection details of the LoG and LoT operators. The original texture image is shown in Figure \ref{fig:example-englishResult1}.}  
    \label{fig:example-englishResult2} 
\end{figure}

The LoT operator's advantages are more pronounced in text. The detection details illustrate that the edges detected by the LoG operator severely shift, such as the punctuation mark “\textbf{.}” and the letter “\textbf{T}”, due to the latent Gaussian smooth operation. The LoT operator is not sensitive to the kernel size during convolution since only the center is negative. For images with black text on a white background, edges are either black-and-white or white-and-black adjacent pixels. The first-order TGD-based edge detection algorithm typically detects the text outline (white pixels) as the edges (Figure~\ref{fig:example-englishResult1}.e), while the LoT-based edge detection algorithm locates the text itself (black pixels) as the edges (Figure~\ref{fig:example-englishResult2}.d). Therefore, the LoT operator is ideal for detecting text edges. 

\begin{figure}[htbp]
    \centering    	 	 		 
    \includegraphics[width=\linewidth]{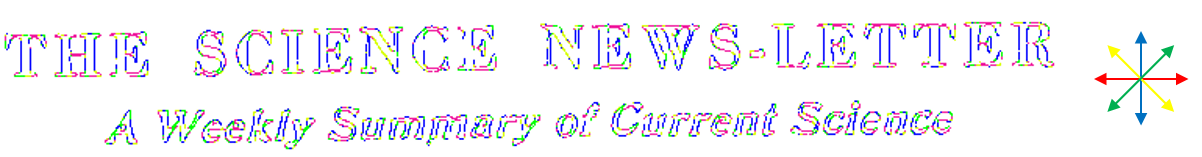}
    \caption{Edge orientation map obtained by $17\times17$ second-order TGD operators in four directions, which are indicated in color arrows.}  
    \label{fig:example-direction2} 
\end{figure}

\begin{figure}[htbp]    	
    \centering    	 
    \subfigure[LoG $17\times17$]{  			 
        \includegraphics[width=0.45\linewidth]{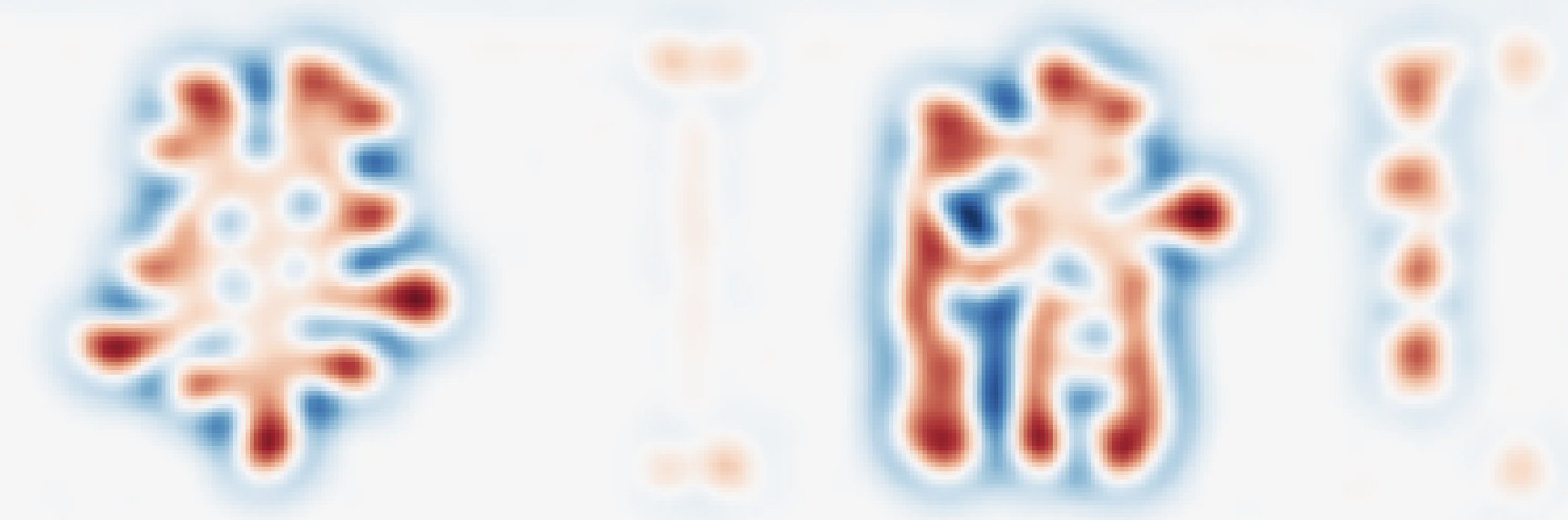}
    }
    \subfigure[LoT $17\times17$]{  			 
        \includegraphics[width=0.49\linewidth]{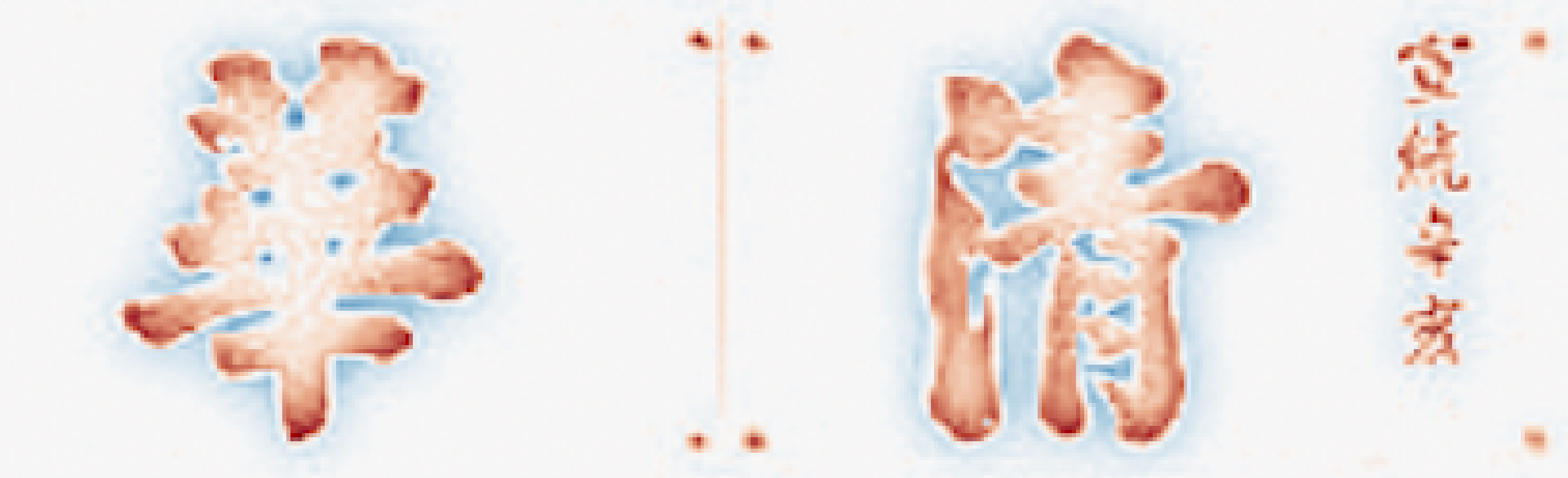}
    } 
    \subfigure[LoG $17\times17$]{   	 		 
        \includegraphics[width=0.45\linewidth]{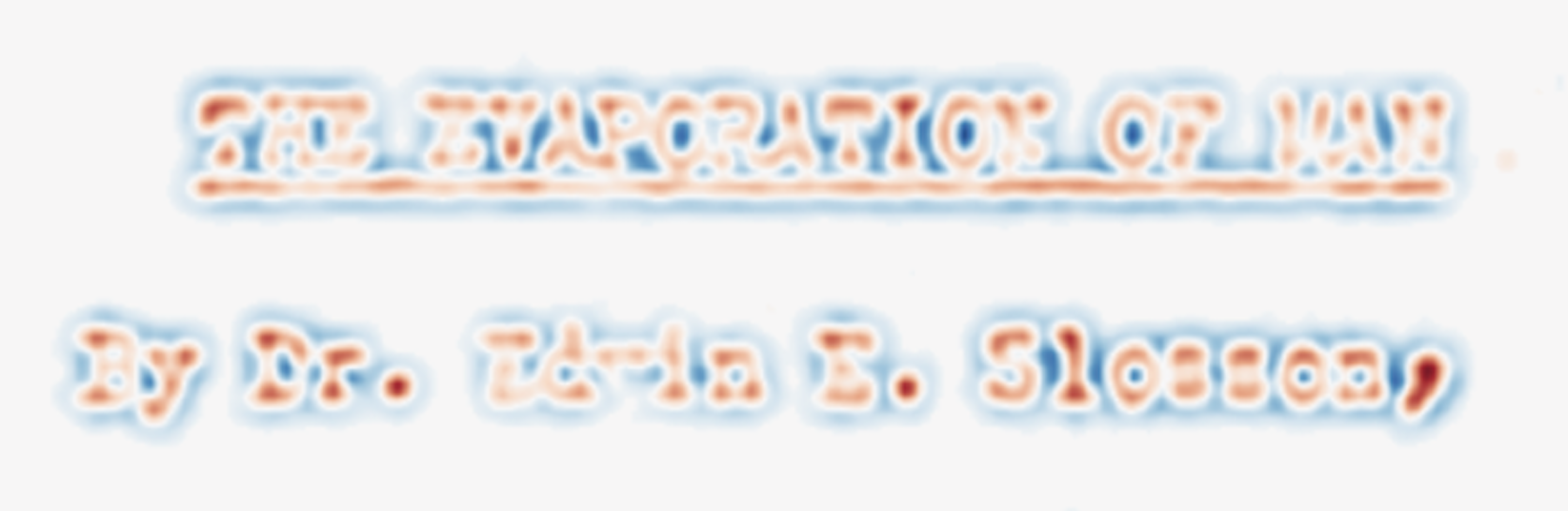}
    }   
    \subfigure[LoT $17\times17$]{   	 		 
        \includegraphics[width=0.47\linewidth]{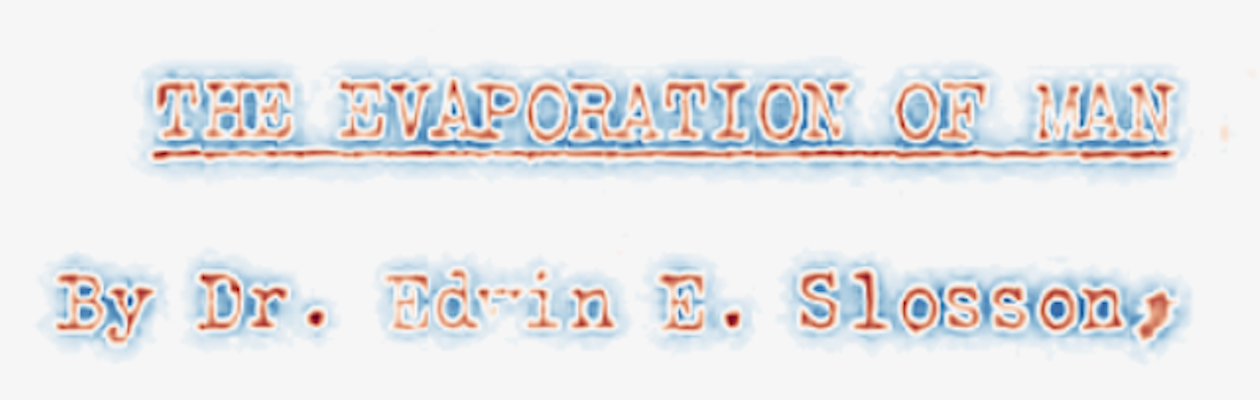}
    }
    
    \caption{Comparison of convolution results: (a,c) convolution results of $17\times17$ LoG operator; (b,d) convolution results of $17\times17$ LoT operator.}  
    \label{fig:example-convResult2} 
\end{figure}

Figure~\ref{fig:example-direction2} illustrates the edge orientation obtained through four-directional orthogonal constructed second-order TGD operators. The detected orientation is the same as the physical edge orientation. Additionally, we observe that both the rotational and orthogonal constructed second-order TGD operators produce similar results.

Figure~\ref{fig:example-convResult2} shows the convolution results of the $17\times17$ LoG and LoT that have been normalized (divided by the maximum positive value), where positive values are red and negative values are blue. The image edges are located at the zero-crossings (white) between red and blue regions. The results indicate that the convolution energy obtained from the LoT operator is precisely concentrated at the edges. In contrast, the convolution energy obtained from the LoG operator spreads away from the edges, causing edge localization drift. This disparity stems fundamentally from the different convolution kernels used. The Monotonic Constraint, which states that only the center weights should be negative (Figure~\ref{fig:laplacian}), ensures the accuracy of edge localization in the LoT operator. Conversely, the LoG operator contains a large number of negative values near the center of the convolution kernel, which triggers the edge localization drift observed in previous studies~\cite{gunn1998edge,gunn1999discrete}.

\clearpage
\newpage

\section{3D TGD-Based Spatio-temporal Edge Detection}

Traditionally, edge is defined as the center of local intensity changes in a single image, which is a 2D space. However, image sequence is three-dimensional in space or spatio-temporal space. \textbf{Here, we define the center of 3D local intensity changes in an image sequence as the \textit{3D edge}.} Based on TGD theory, we construct 3D TGD operators for 3D edge detection in video or image sequences. 

\begin{figure}[htp]
\centering
 \includegraphics[width=0.95\linewidth]{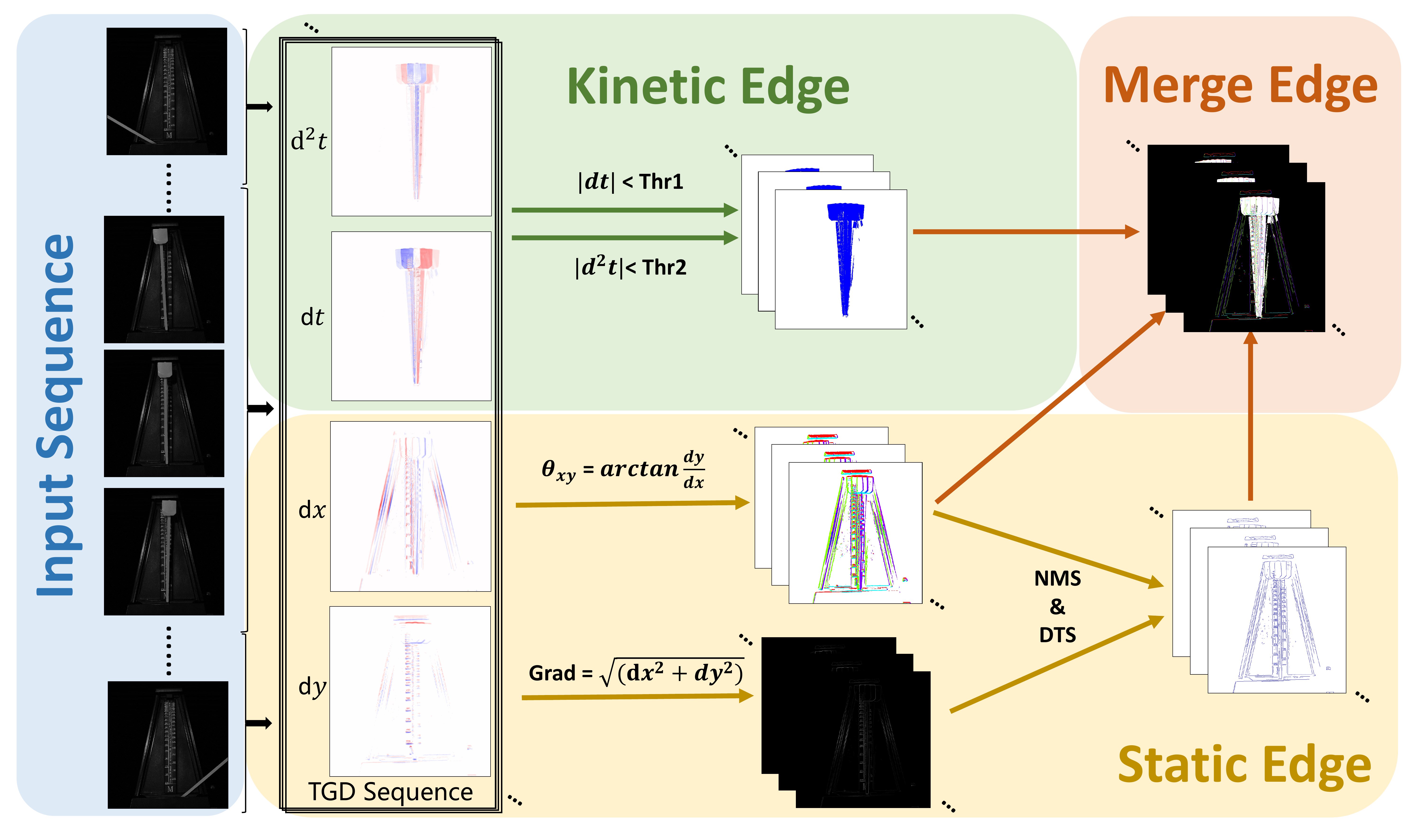}
 \caption{
   Flowchart of TGD-based 3D edge detection algorithm.
 }
 \label{fig:Algorithm3D}
\end{figure}

The TGD-based 3D edge detection algorithm is presented in Figure~\ref{fig:Algorithm3D}, and its pseudo-code is described in Algorithm~\ref{algorithm:Algorithm3D}. Our algorithm comprises three parts. The first part detects local extremes of gradients on the $xy$ plane, using the first-order TGD-based image edge detection algorithm. Our approach incorporates 3D TGD operators to obtain partial TGD in the $x$- and $y$-directions, which enables the algorithm to capture background details shielded by moving objects. We refer to this type of edge as \emph{Static edge}, where a jump in brightness value occurs in the spatial domain. 

The second part is edge detection in temporal space, whereby pixels exhibiting 1D first- or second-order TGD larger than the corresponding threshold along the $t$-axis are marked out\footnote{Kinetic edge detection needs to operate at the pixel level. 3D TGD operators, with the effect of spatial smoothing, may result in wrongly identifying static pixels as motion pixels. Hence, we suggest using 1D TGD operators.}. Although these pixels differ from \emph{static edges}, a substantial brightness change along the $t$- axis implies the occurrence of motion. We refer to these motion sensitive pixels marked out in this part as \emph{kinetic edge}, where is the center of intensity changes occurring over time.

It is worth noting that the central weight value of the first-order TGD operator is $0$, indicating that it is only aware of previous and subsequent states but not the current state. Thus, we combine it with second-order TGD to complete kinetic edge detection. Additionally, we use the HSV space to visualize edge directions, which we combine with gradient directions.

\begin{algorithm}[htp]
  \caption{TGD-based 3D Edge Detection Algorithm}
  \hspace*{0.1in} {\bf Input:} 
    Gray image sequence $I$, 3D first-order discrete TGD operators in $x$- and $y$-directions $\widehat{T}_{x}, \widehat{T}_{y}$, 1D first- and second-order discrete TGD operator in $t$-direction $\widehat{T}_{t}$,$\widehat{R}_{t}$, motion threshold $Thr_1$ and $Thr_2$, high and low threshold $highThr$, $lowThr$ for Double-threshold Selection.\\
  \hspace*{0.05in} {\bf Output:} 
  Motion edge locations $M$, Static edge locations $S$, and merge visualization $V$.
  \begin{algorithmic}[1]
  \STATE $dx = I * {\widehat{T}}_{x}$
  \STATE $dy = I * {\widehat{T}}_{y}$
  \STATE $dt = I * {\widehat{T}}_{t}$
  \STATE $d^2t = I * {\widehat{R}}_{t}$
  \STATE $Grad = \sqrt{dx^2 + dy^2}$
  \STATE $\Theta_{xy} = \arctan(dy / dx)$
  \STATE $NMS = \text{Non-Maximum Suppression}\left(Grad, \Theta_{xy}\right)$
  \STATE $S = \text{Double-Threshold Selection}(NMS, lowThr, highThr)$
  \STATE $M = |dt| > Thr_1 \ \text{or} \  |d^2t| > Thr_2$
  \STATE $V = \text{HSV Visualization}\left(\Theta_{xy}, M, S\right)$
  \RETURN $M$, $S$, $V$
  \end{algorithmic}
  \label{algorithm:Algorithm3D}
\end{algorithm}

In our experiments, we adopt the TGD operators acquired using the orthogonal construction method. 
The 3D TGD operators in the $x$- and $y$-directions employ the Gaussian kernel function, 
while the 1D TGD operators in the $t$-direction use the linear kernel function. 
Their kernel sizes are $15\times15\times15$ and $1\times1\times15$, respectively. Additionally, we scale the time dimension by either copying or skipping frames.

\begin{figure}[htp]    	
    \centering    	
    \subfigure[Input Frame Sequence]{  			 
        \includegraphics[width=0.96\linewidth]{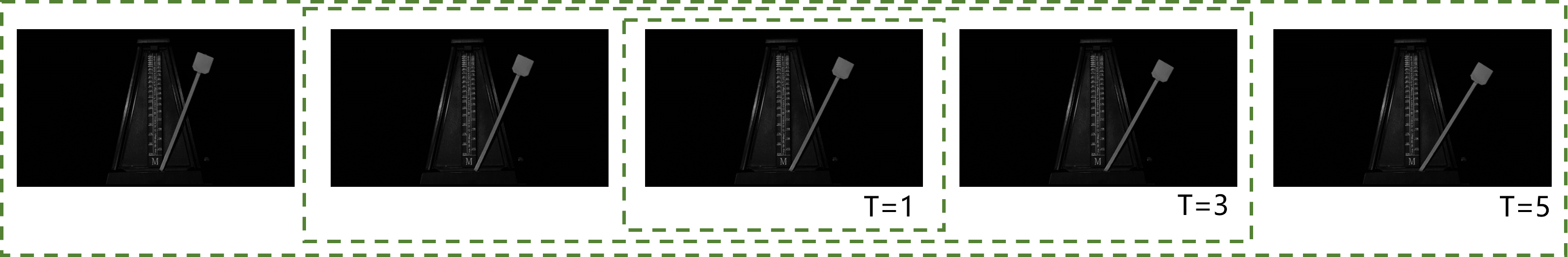}
    }
    \subfigure[Merge (T = 1)]{  			 
        \includegraphics[width=0.31\linewidth]{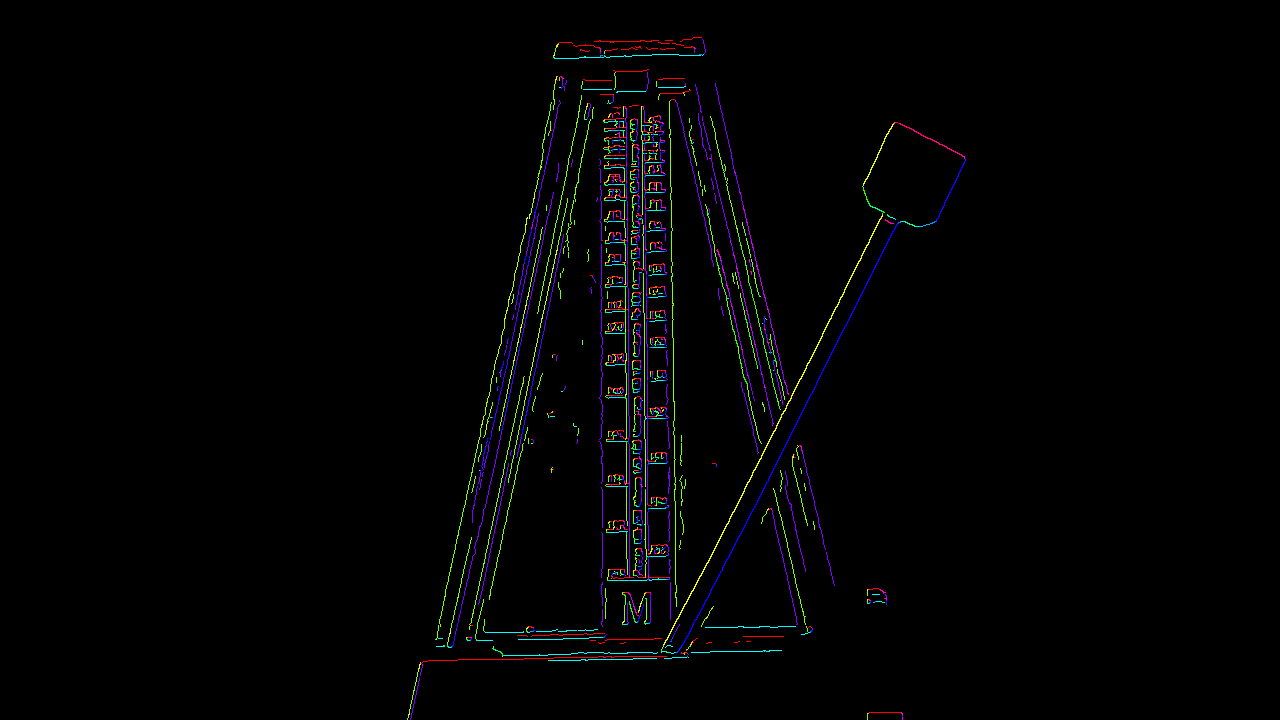}
    } 
    \subfigure[Merge (T = 3)]{  			 
        \includegraphics[width=0.31\linewidth]{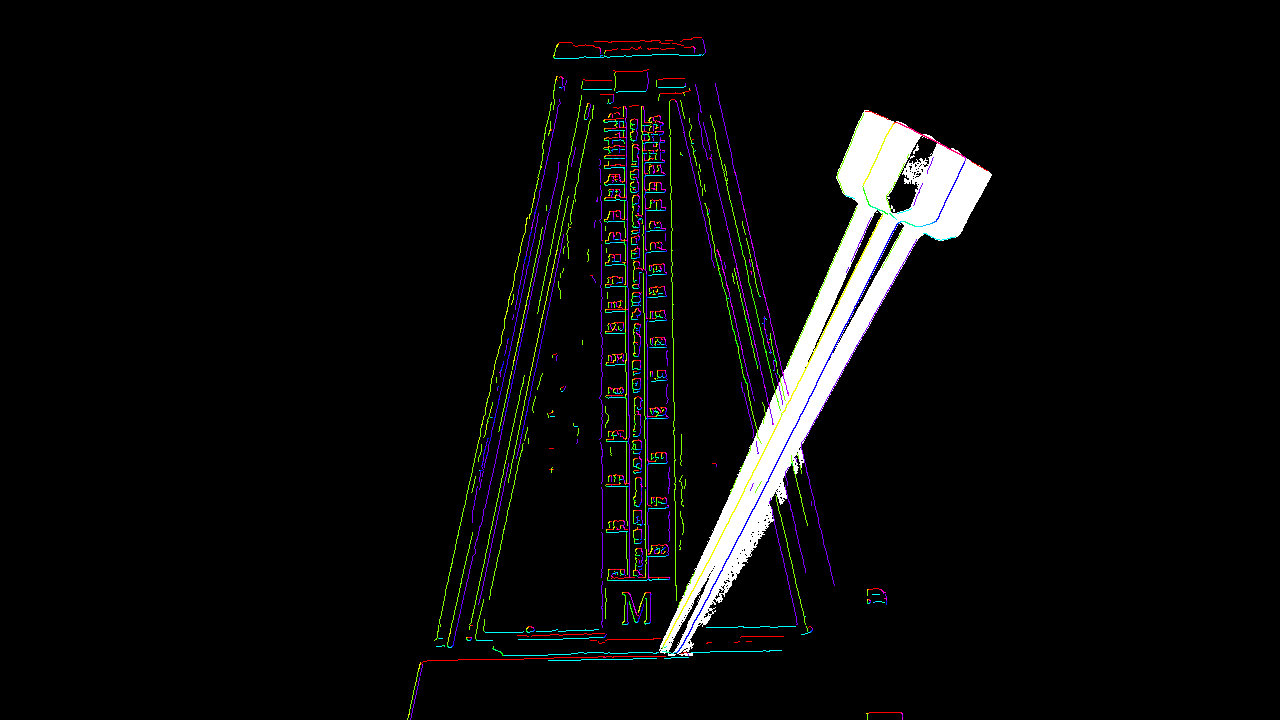}
    }
    \subfigure[Merge (T = 5)]{  			 
        \includegraphics[width=0.31\linewidth]{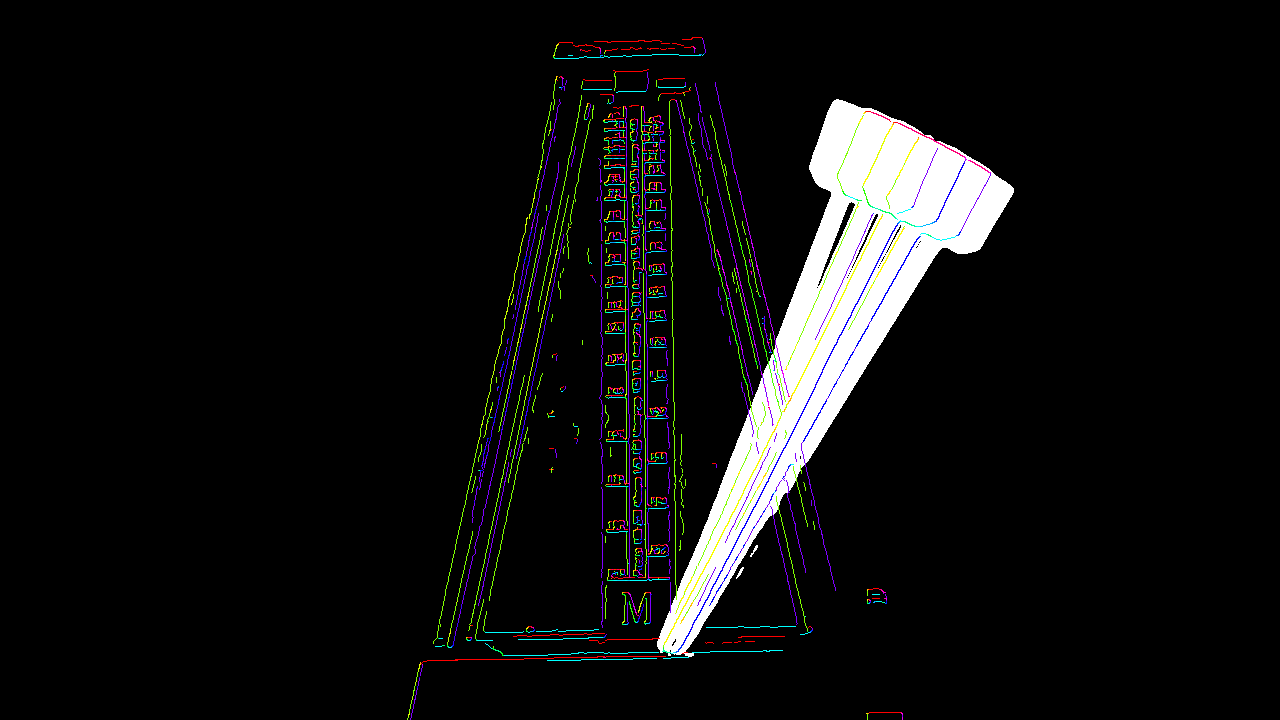}
    } 
    \subfigure[Static Edges (T = 1)]{  			 
        \includegraphics[width=0.31\linewidth]{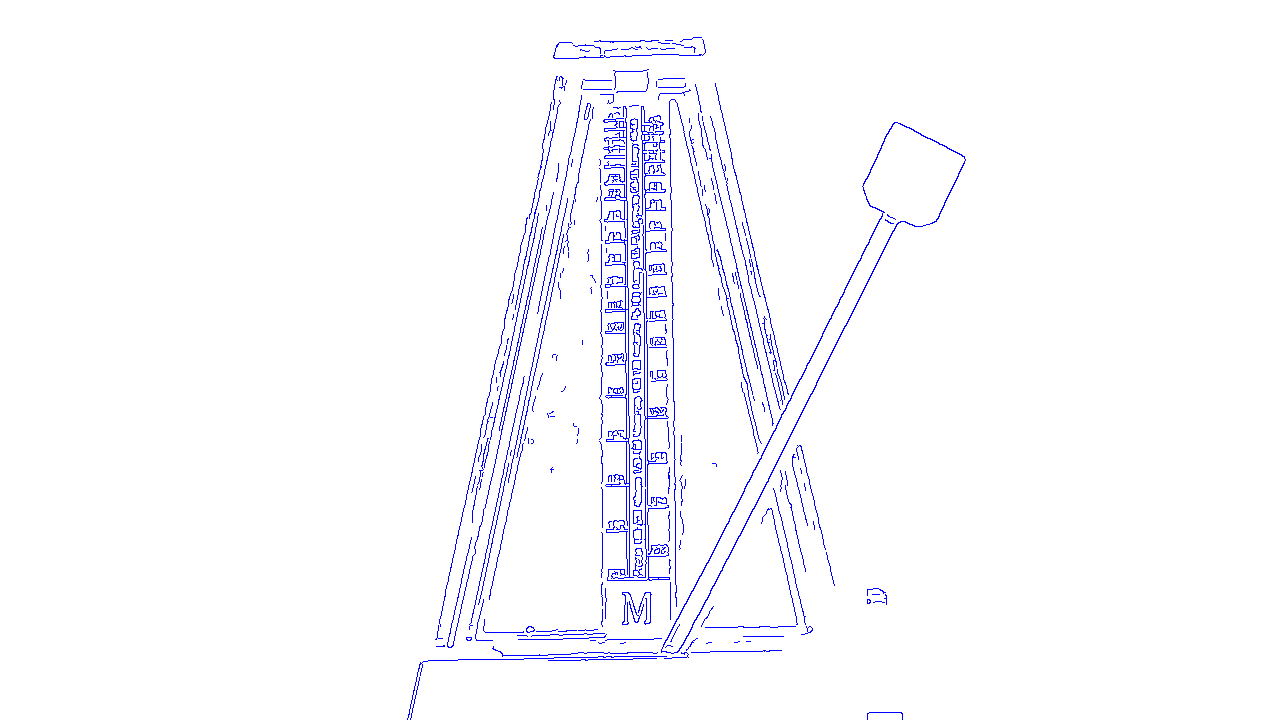}
    } 
    \subfigure[Static Edges (T = 3)]{  			 
        \includegraphics[width=0.31\linewidth]{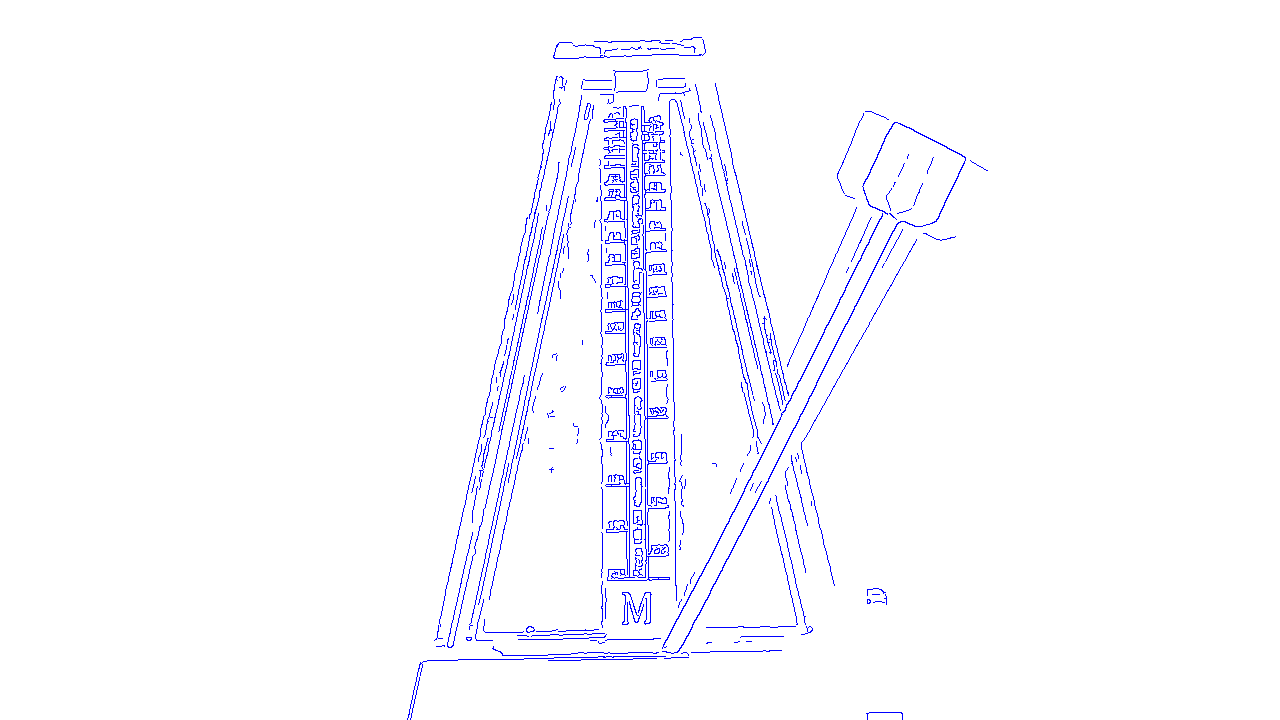}
    }
    \subfigure[Static Edges (T = 5)]{  			 
        \includegraphics[width=0.31\linewidth]{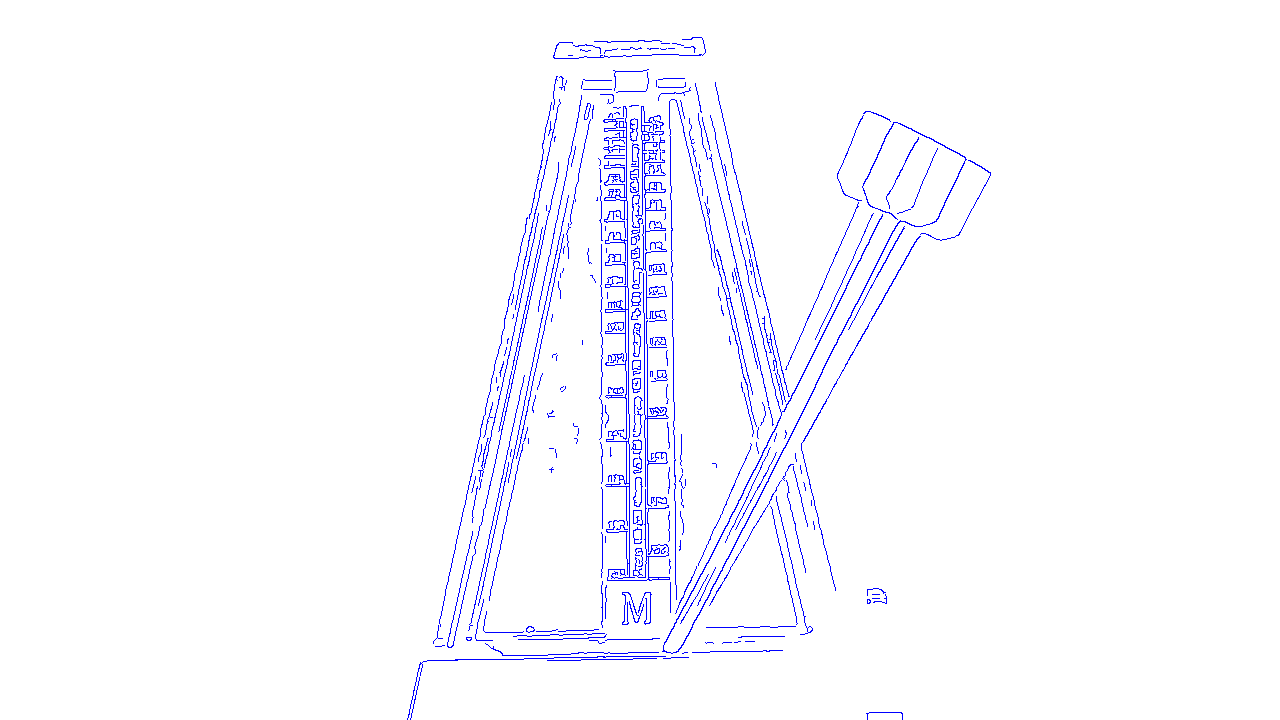}
    } 
    \subfigure[Image Edges (T = 1)]{  			 
        \includegraphics[width=0.31\linewidth]{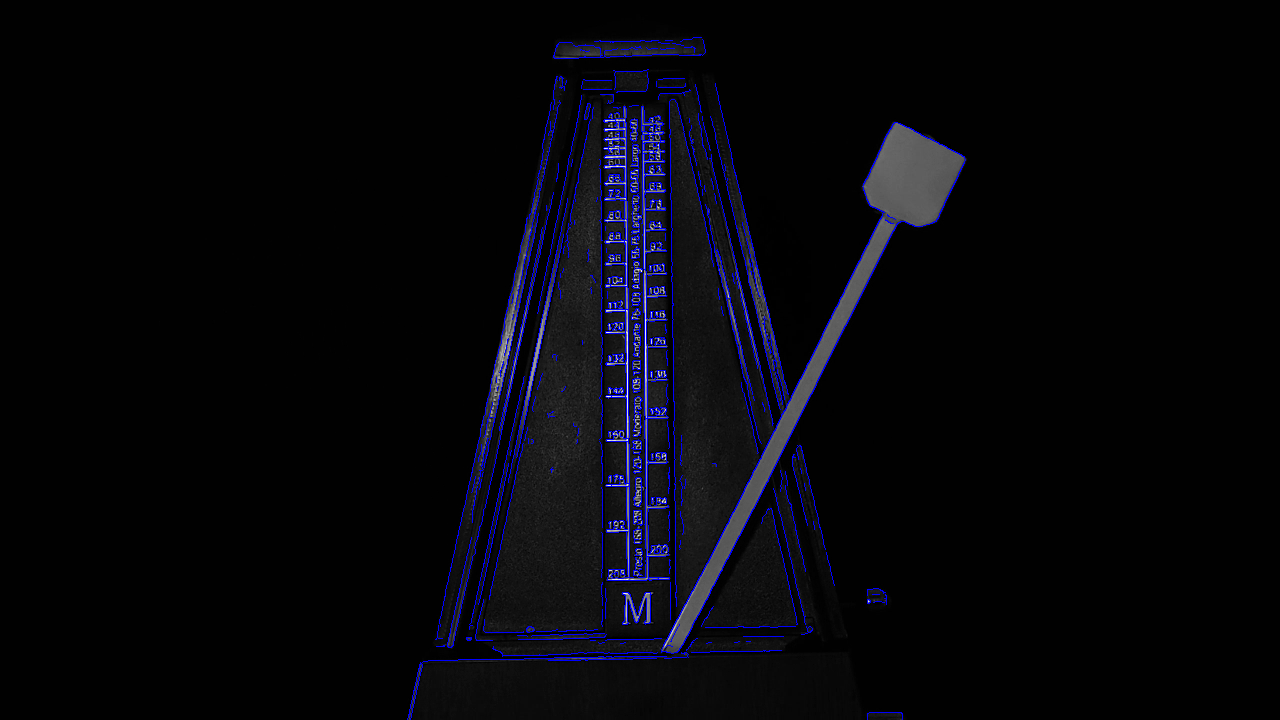}
    } 
    \subfigure[Kinetic Edges (T = 3)]{  			 
        \includegraphics[width=0.31\linewidth]{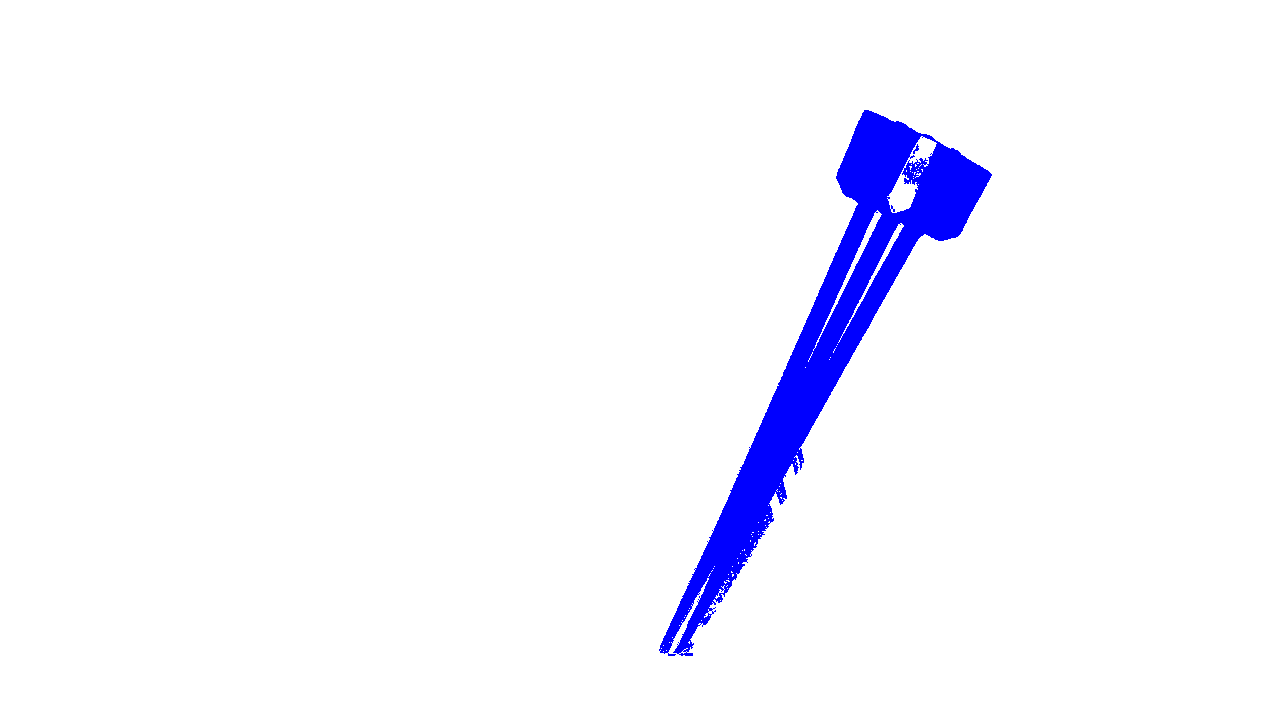}
    }
    \subfigure[Kinetic Edges (T = 5)]{  			 
        \includegraphics[width=0.31\linewidth]{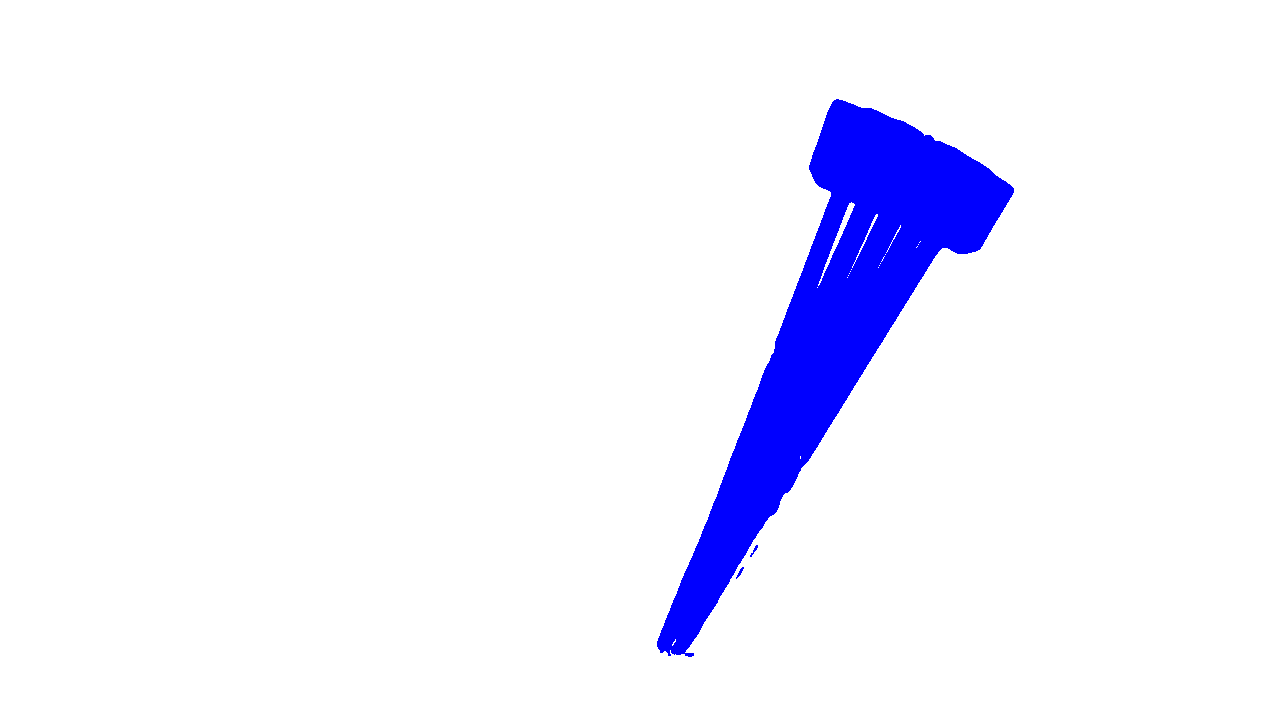}
    } 
    \caption{3D Edge detection results: (a) Input frame sequence, where T represents the number of effective frames; (b-d) merged visualization via HSV space; (e-g) static edges; (h) static edges of the constant frame sequence are equivalent to image edges; (i-j) kinetic edges.}  
    \label{fig:example3D-zhongbai} 
\end{figure}

Let us start to visualize and analyze the 3D edge under an ideal condition. We designed a scenario where a pendulum constantly swings left and right in front of a plain black background. The surface of the pendulum is treated to avoid specular reflection and maintain consistent surface brightness. Additionally, we used a floodlighting system to avoid the interference of shadows. Since the pendulum moves fast in the close shot, we slow down the motion of the swinging pendulum by copying frames while maintaining a frame sequence length of $15$ that is used for calculating TGD, in which the number of effective frames is $5$ since we create $3$ copies for each frame. This setup allows for an idealistic visualization of 3D edges and analysis of the detection algorithms. Figure~\ref{fig:example3D-zhongbai} shows that both the kinetic and static edge locations and orientations are all obtained with 3D TGD edge detection algorithms. Note that when only one effective frame is available, the input sequence is constant, then TGD-based 3D edge detection is equivalent to the corresponding image edge detection on the center frame (Figure~\ref{fig:example3D-zhongbai}.h). 

With moving objects in the input frame sequence, the TGD-based algorithm detects the moving parts effectively, generating the same static edge results as those in the constant frame sequence(Figure~\ref{fig:example3D-zhongbai}.e-g). As the number of effective frames increases, we could detect kinetic edges(Figure~\ref{fig:example3D-zhongbai}.i-j), and some static edges that are not present in the intermediate frame could be identified. This is because moving objects in different positions in each frame cause physical edges to overlap in the final results, producing identified static edges. Occluded physical edges that are partially exposed may also overlap in the final results, producing detected static edges.

\begin{figure}[htbp]    	
    \centering   
    \subfigure[Intermediate frame]{  			 
        \includegraphics[width=0.31\linewidth]{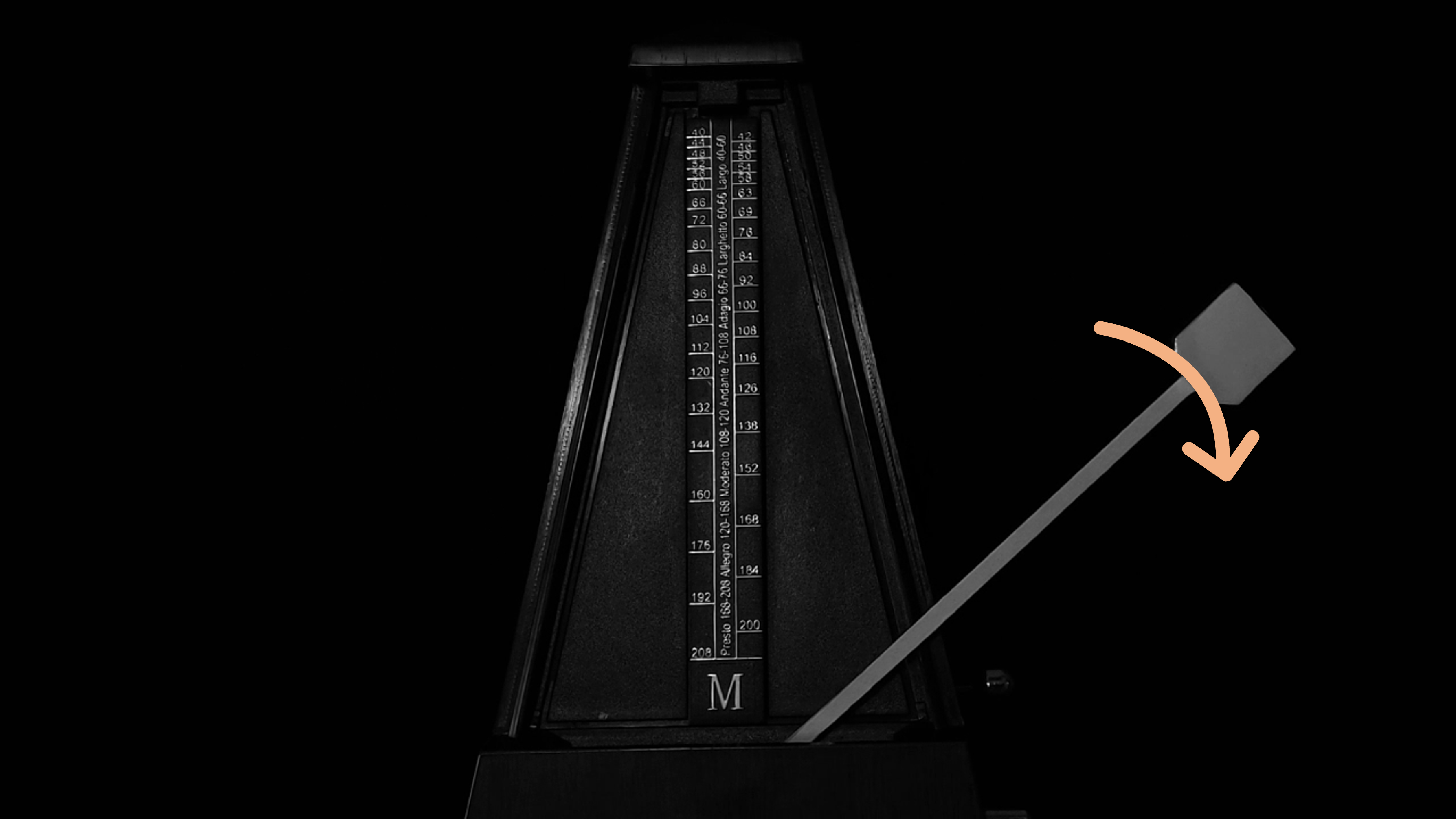}
    }  	
    \subfigure[Intermediate frame]{  			 
        \includegraphics[width=0.31\linewidth]{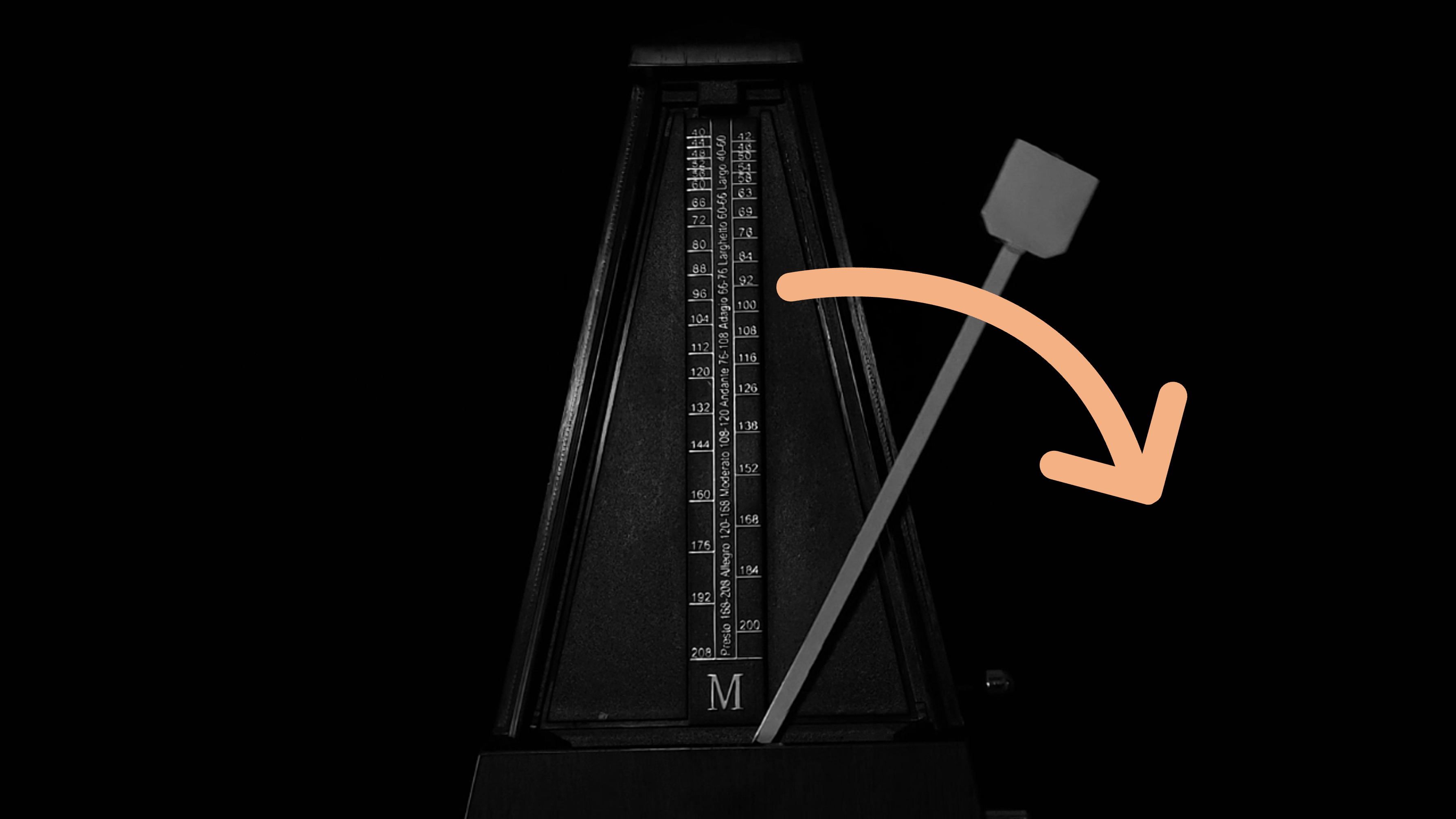}
    }
    \subfigure[Intermediate frame]{  			 
        \includegraphics[width=0.31\linewidth]{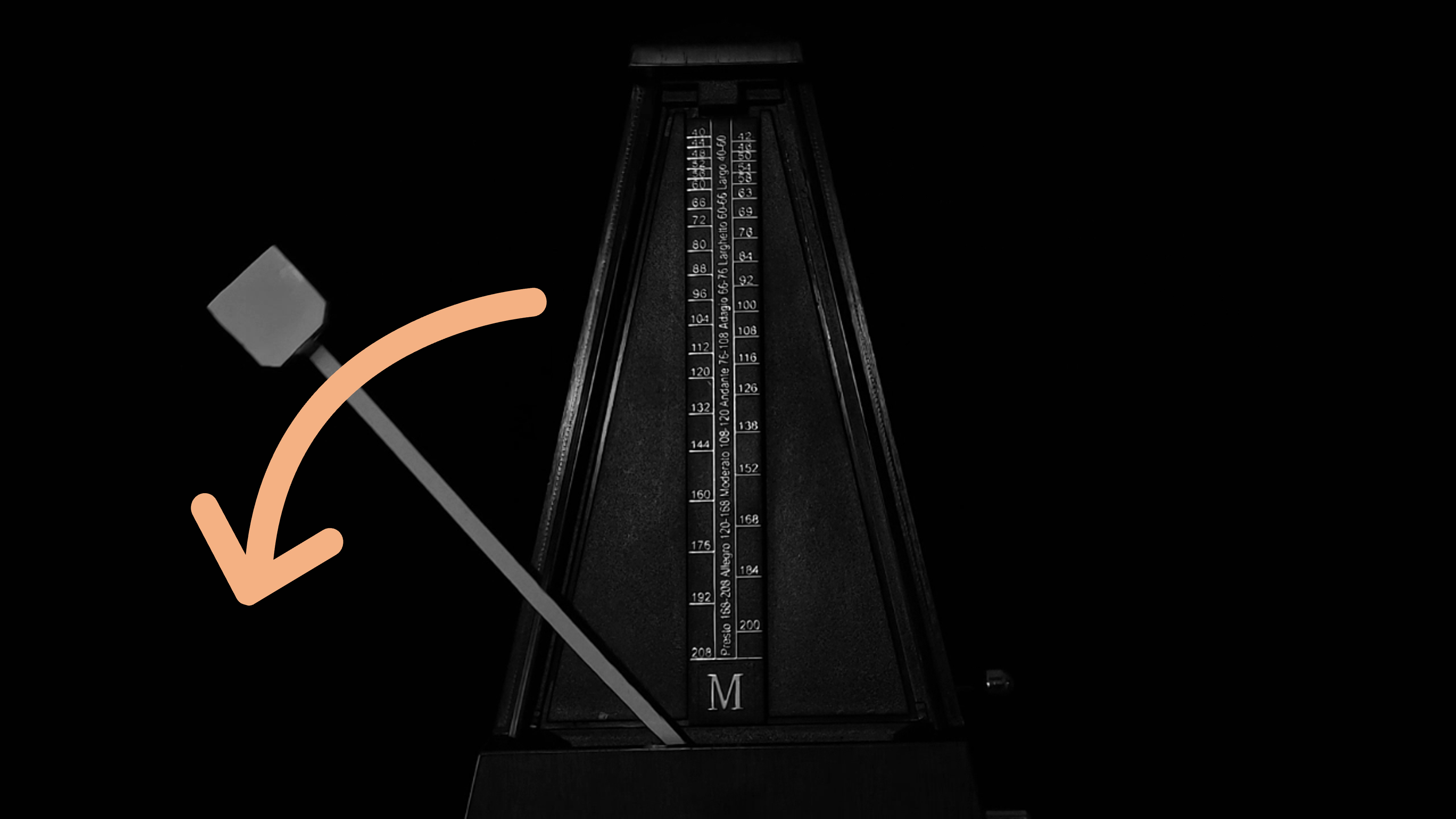}
    } 
    \subfigure[$dt$]{  			 
        \includegraphics[width=0.31\linewidth]{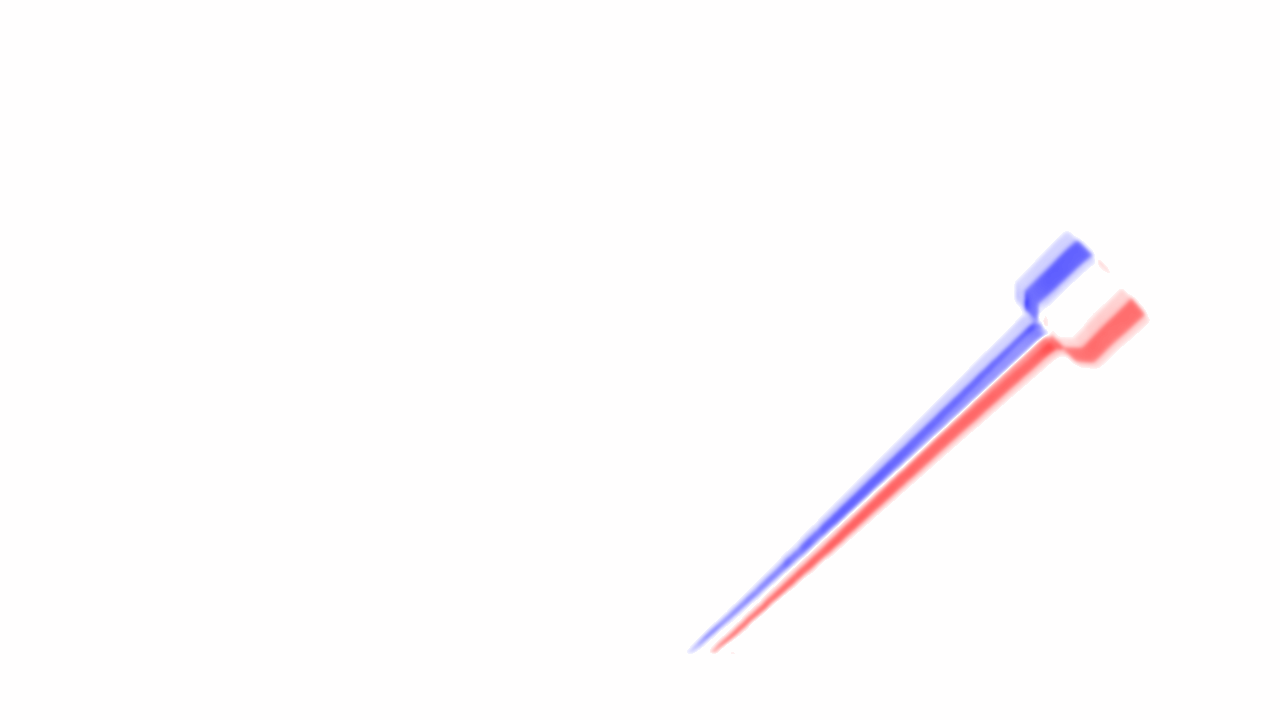}
    }
    \subfigure[$dt$]{  			 
        \includegraphics[width=0.31\linewidth]{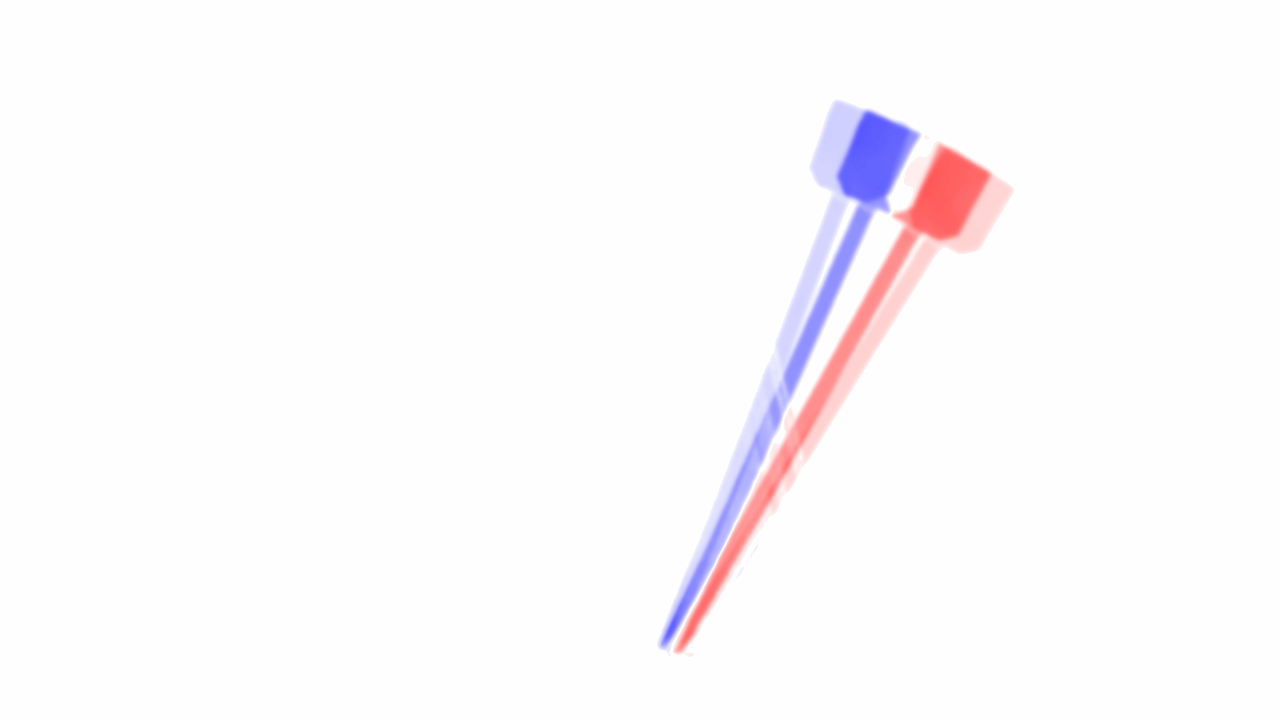}
    } 
    \subfigure[$dt$]{  			 
        \includegraphics[width=0.31\linewidth]{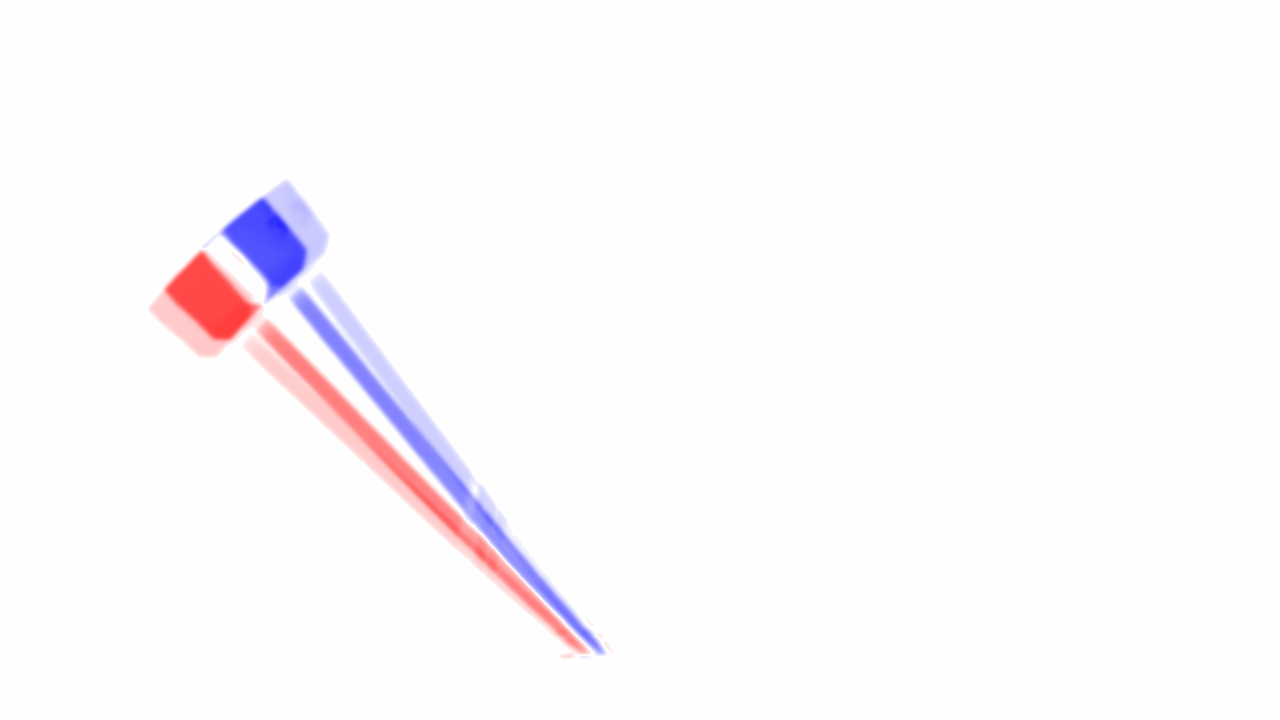}
    }
    \subfigure[$d^2 t$]{  			 
        \includegraphics[width=0.31\linewidth]{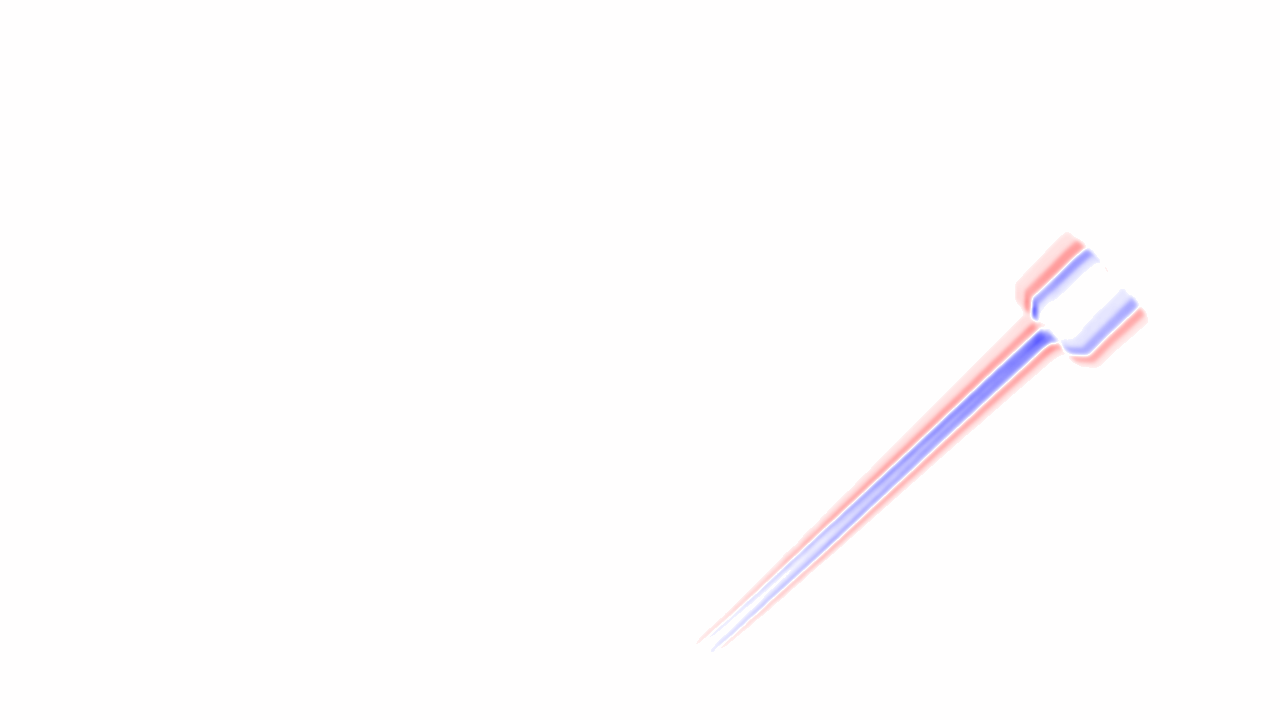}
    } 
    \subfigure[$d^2 t$]{  			 
        \includegraphics[width=0.31\linewidth]{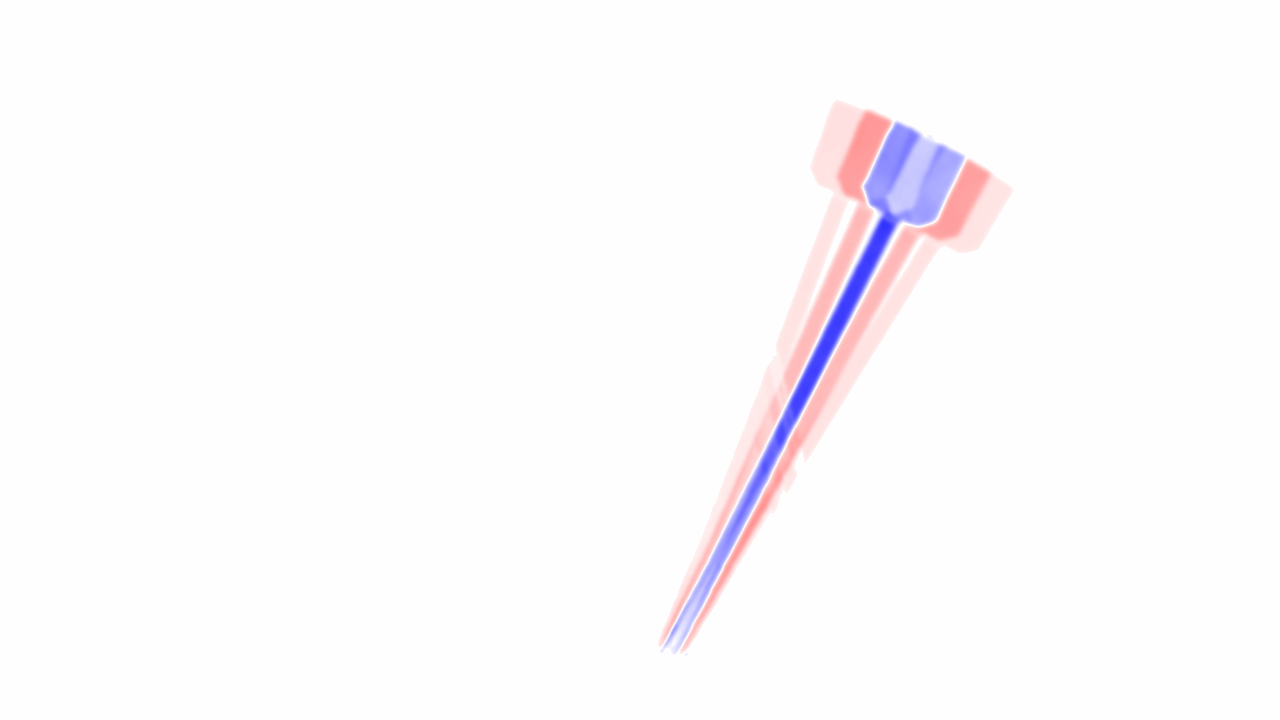}
    } 
    \subfigure[$d^2 t$]{  			 
        \includegraphics[width=0.31\linewidth]{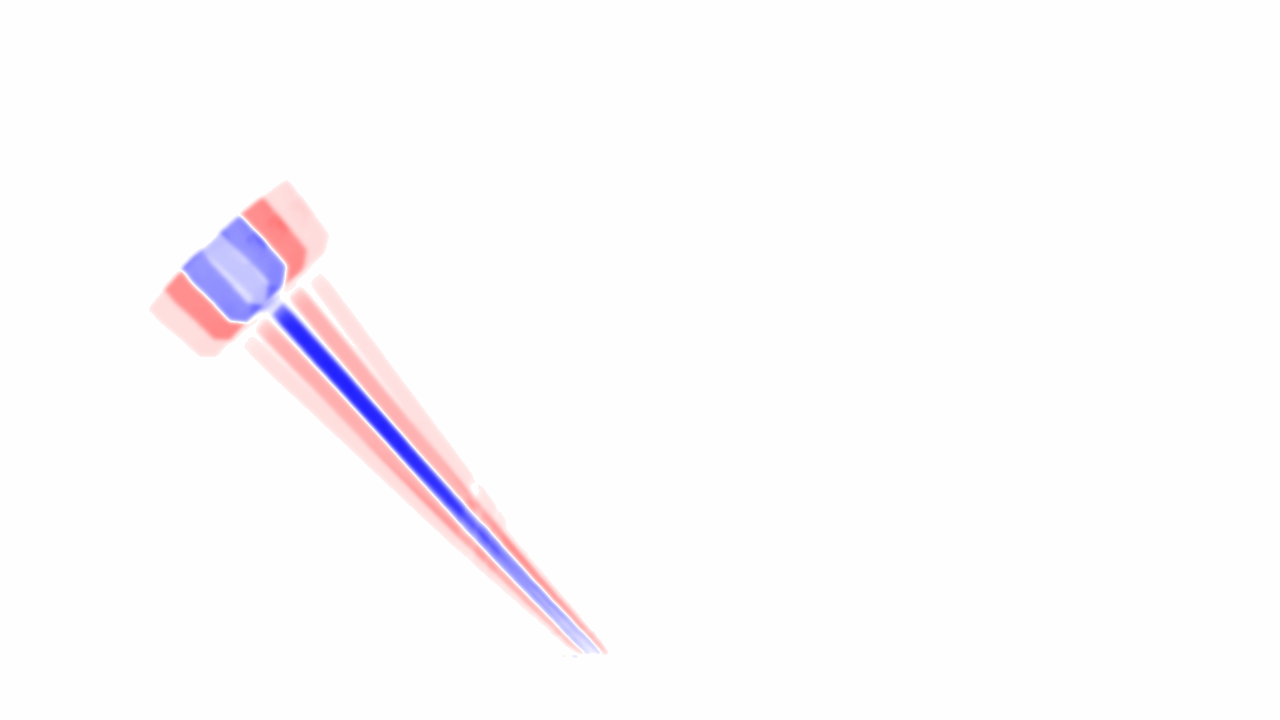}
    } 
    \caption{The correlation results of motion direction and partial TGD.}  
    \label{fig:example3D-motion} 
\end{figure}

Moreover, TGD results along the time axis can provide a deeper understanding of motion, beyond detecting the brightness fluctuation at a particular location to ascertain if a moving object is passing through. Here we assume that the moving object is brighter than the background. As shown in Figure~\ref{fig:example3D-motion}, we select three cases, with each comprising five effective frames. The orange arrows indicate the pendulum's moving direction. The first-order TGD detects moving objects in previous frames, producing negative responses (blue), and moving objects in forthcoming frames, producing positive responses (red) (Figure~\ref{fig:example3D-motion}.d-f). In contrast, for the second-order TGD, moving objects within the intermediate frame generate negative responses (blue), while moving objects in other frames generate positive responses (red)(Figure~\ref{fig:example3D-motion}.g-i). Further the frames away from the intermediate frame, the more the intensity of the response reduces\footnote{This is caused by the Monotonic Constraint of TGD.}. Combining this information would further enable the inference of motion direction, which is crucial in various downstream applications. 

\begin{figure}[ht]    	
    \centering    	
    \subfigure[Input Frame Sequence (T = 5)]{  			 
        \includegraphics[width=\linewidth]{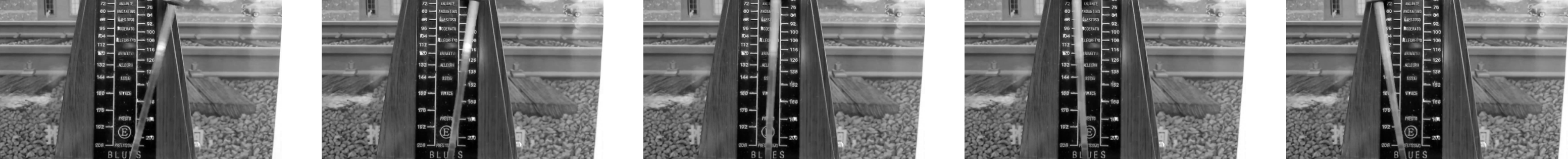}
    }
    \subfigure[$dx$]{  			 
        \includegraphics[width=0.23\linewidth]{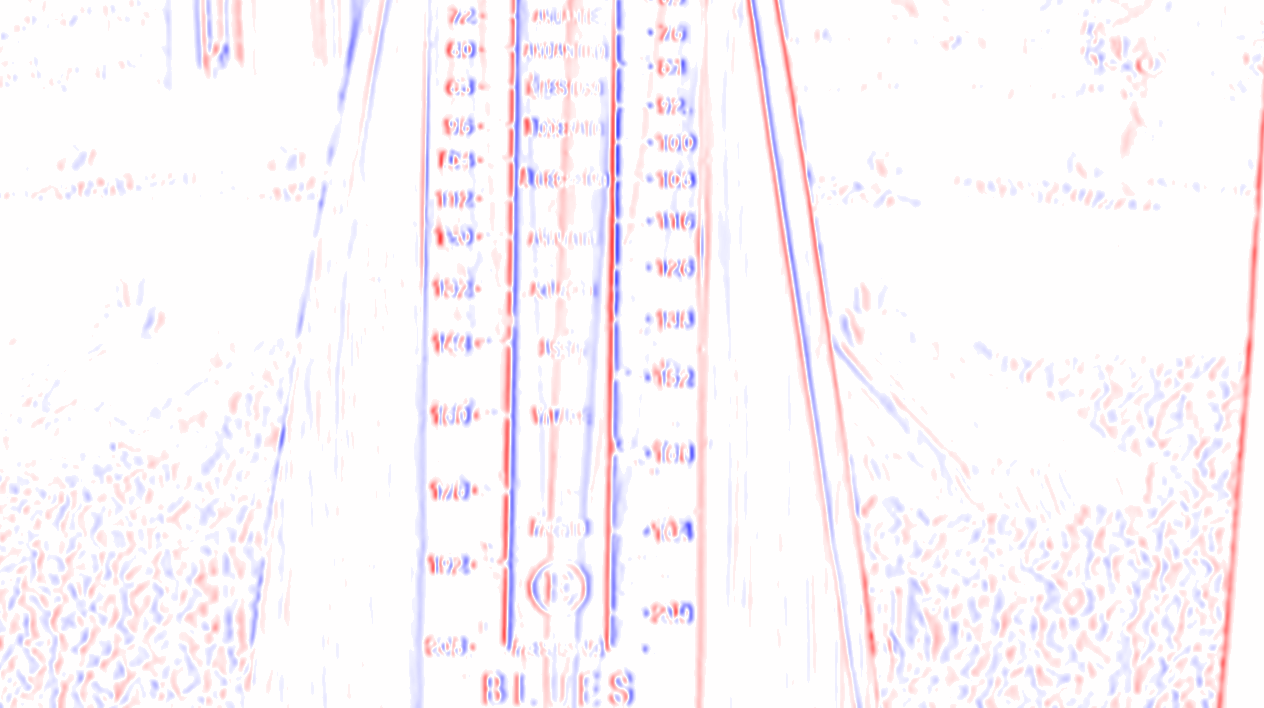}
    } 
    \subfigure[$dy$]{  			 
        \includegraphics[width=0.23\linewidth]{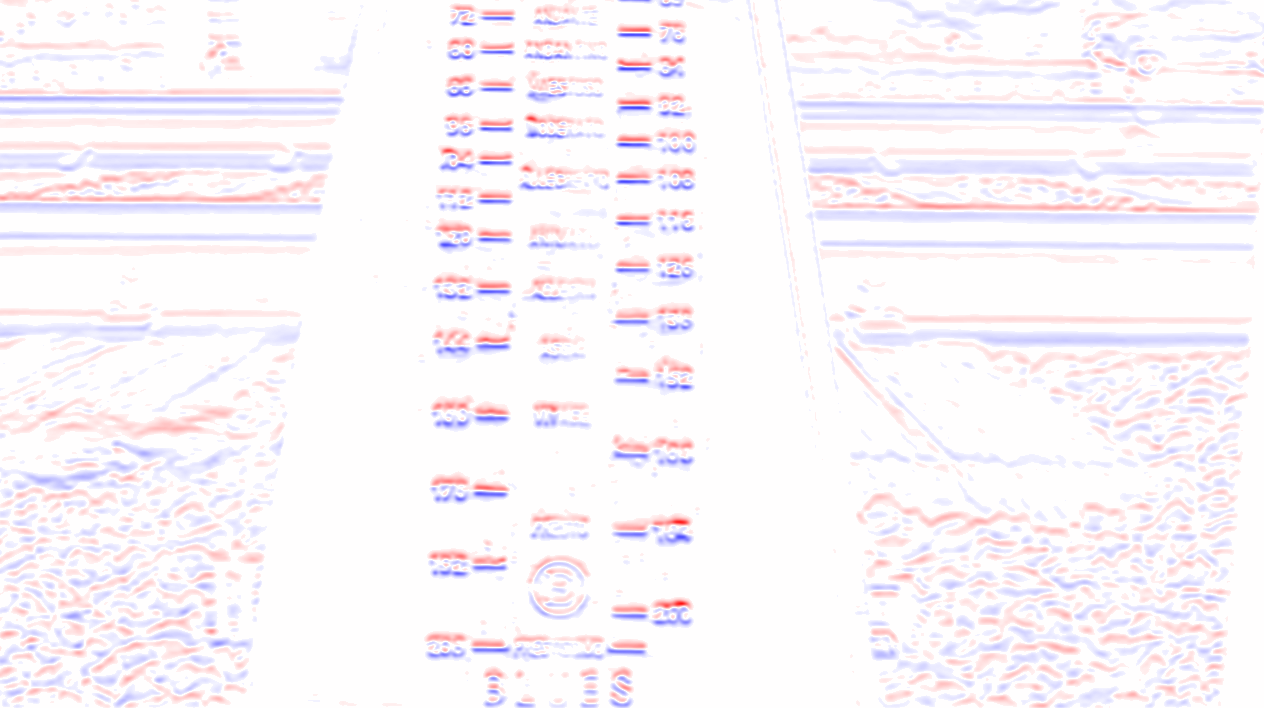}
    }
    \subfigure[$dt$]{  			 
        \includegraphics[width=0.23\linewidth]{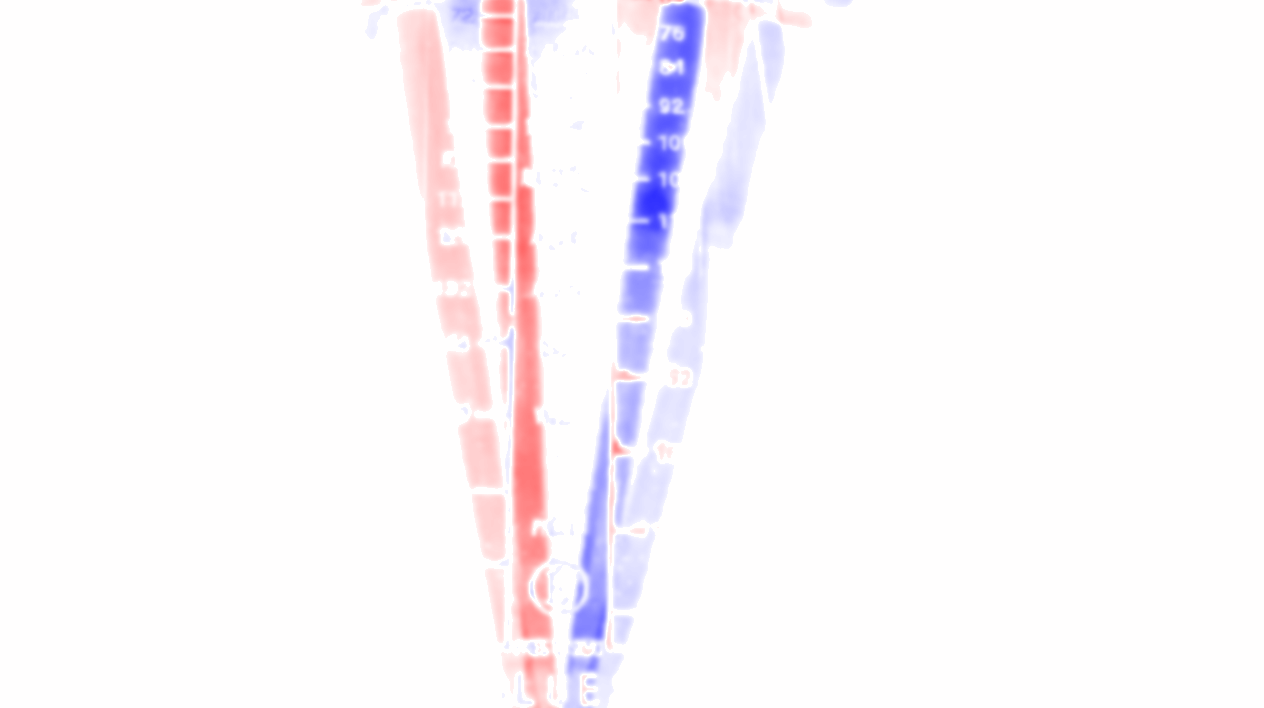}
    } 
    \subfigure[$d^2 t$]{  			 
        \includegraphics[width=0.23\linewidth]{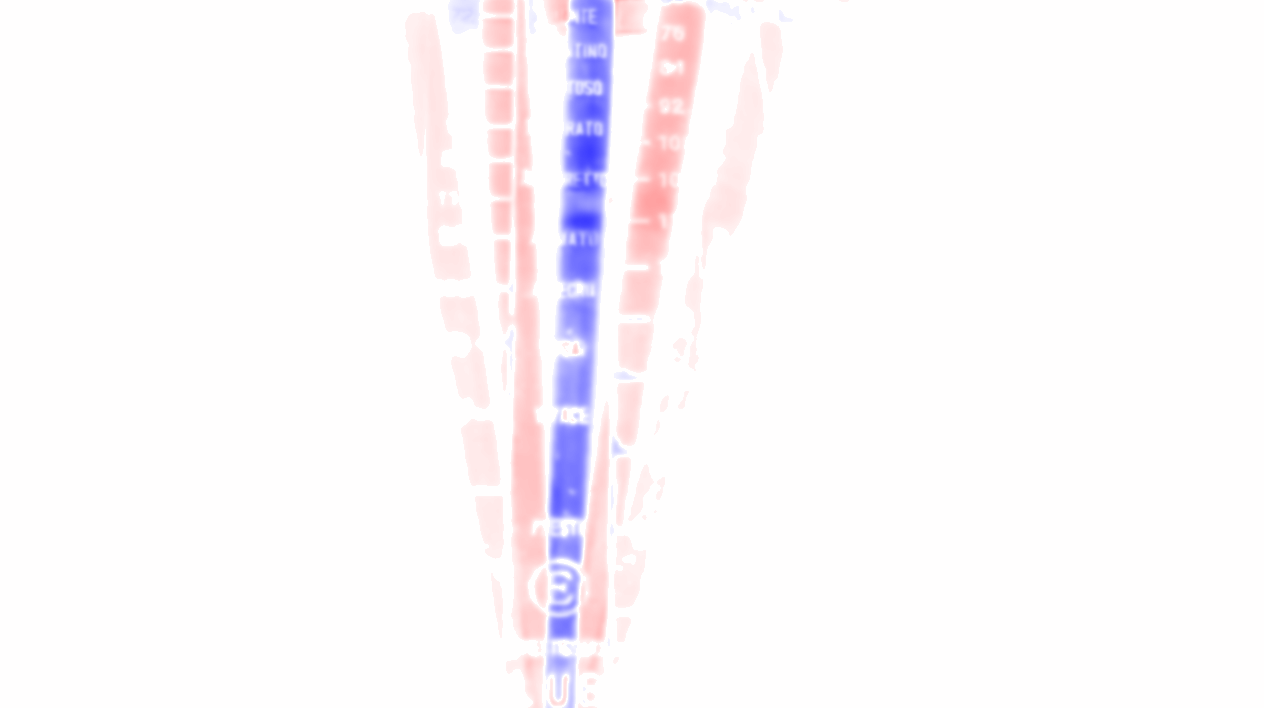}
    } 
    \subfigure[Static Edges]{  			 
        \includegraphics[width=0.48\linewidth]{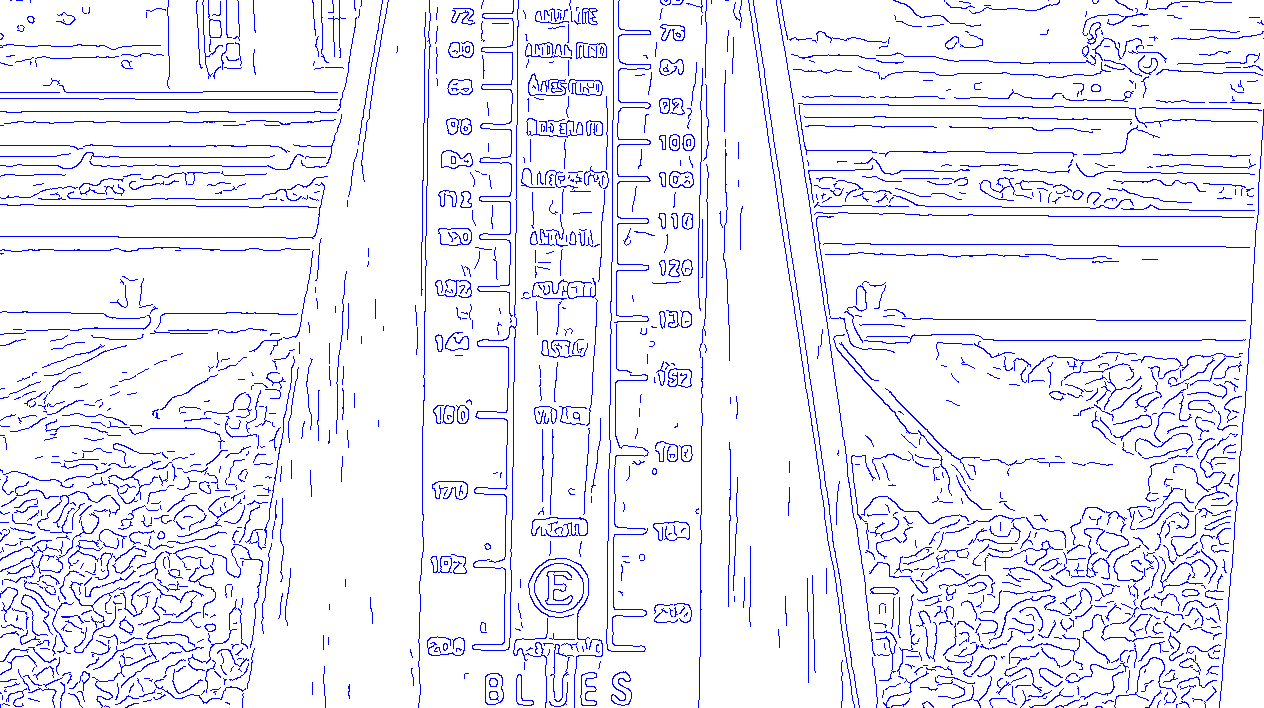}
    } 
    \subfigure[Motion Edges]{  			 
        \includegraphics[width=0.48\linewidth]{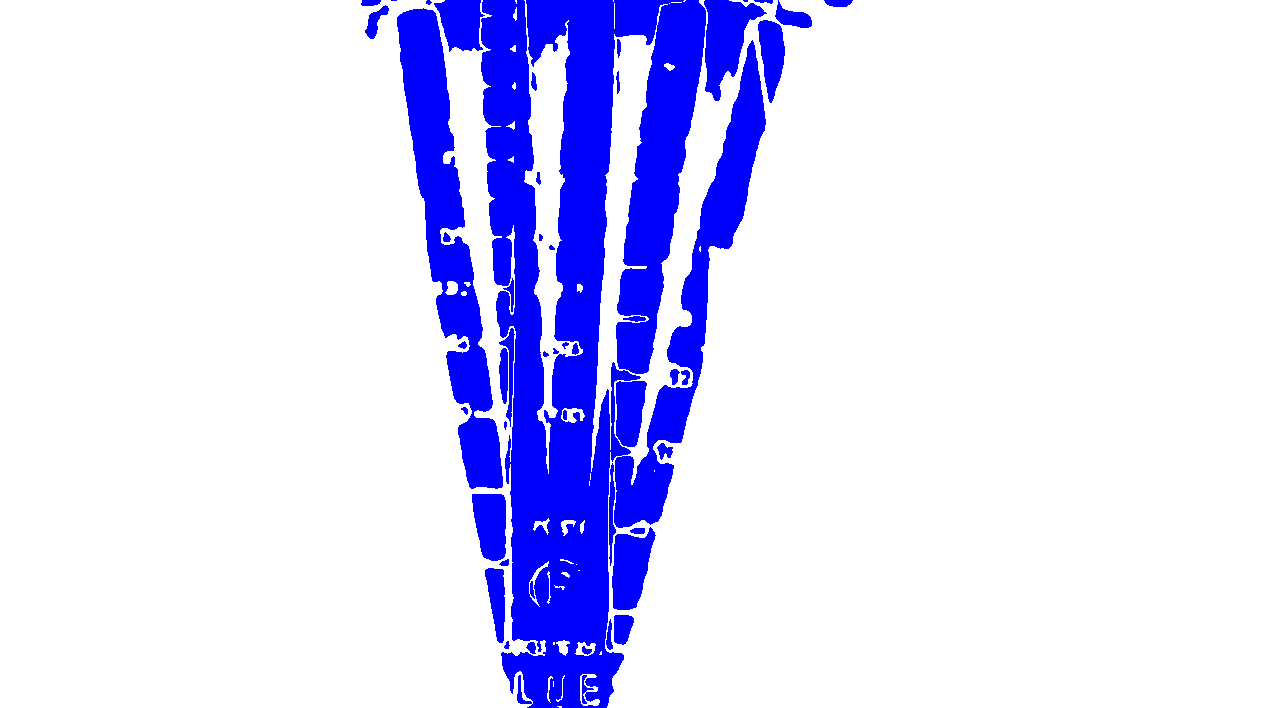}
    }
    \subfigure[Merge]{  			 
        \includegraphics[width=0.8\linewidth]{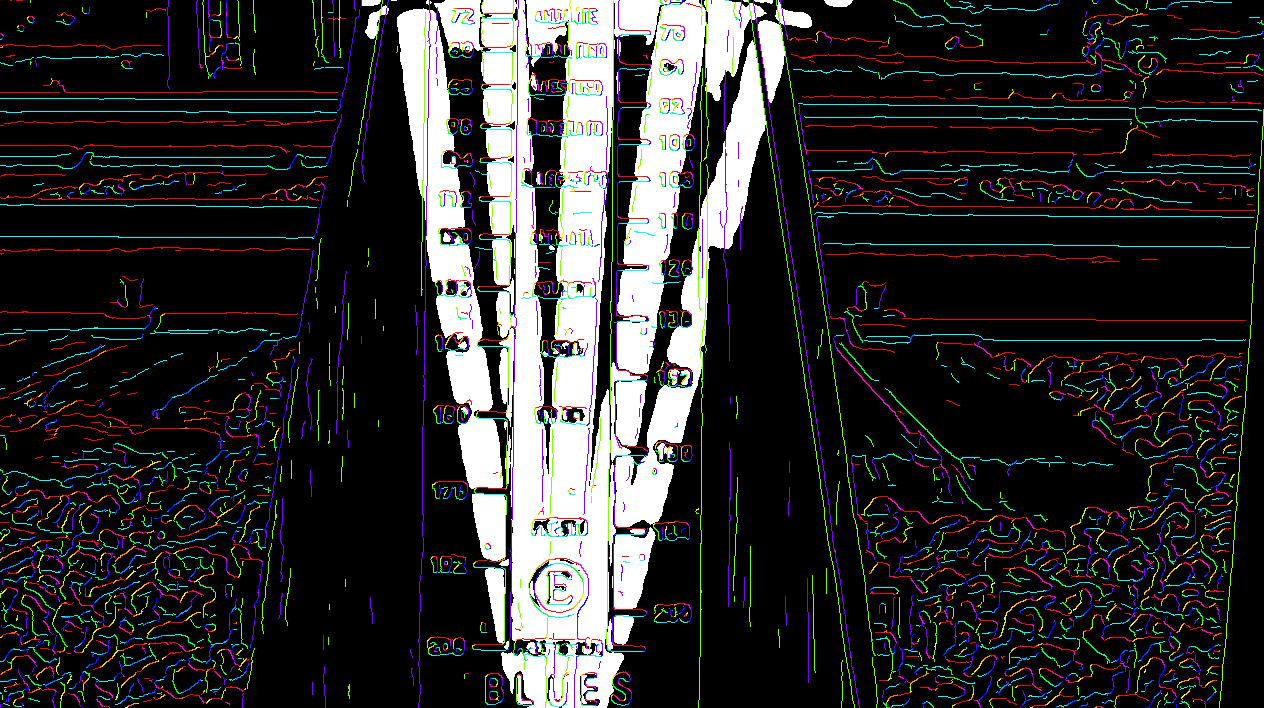}
    } 
    \caption{3D Edge detection results: (a) Input frame sequence, where T represents the number of effective frames; (b-e) first- and second-order TGD; (f) static edges;  (g) kinetic edges; (h) merged visualization via HSV space.} 
    \label{fig:example3D-pen_main} 
\end{figure}

\begin{figure}[htbp]    	
    \centering
    \subfigure[Kinetic edges of shadows]{  			 
        \includegraphics[width=\linewidth]{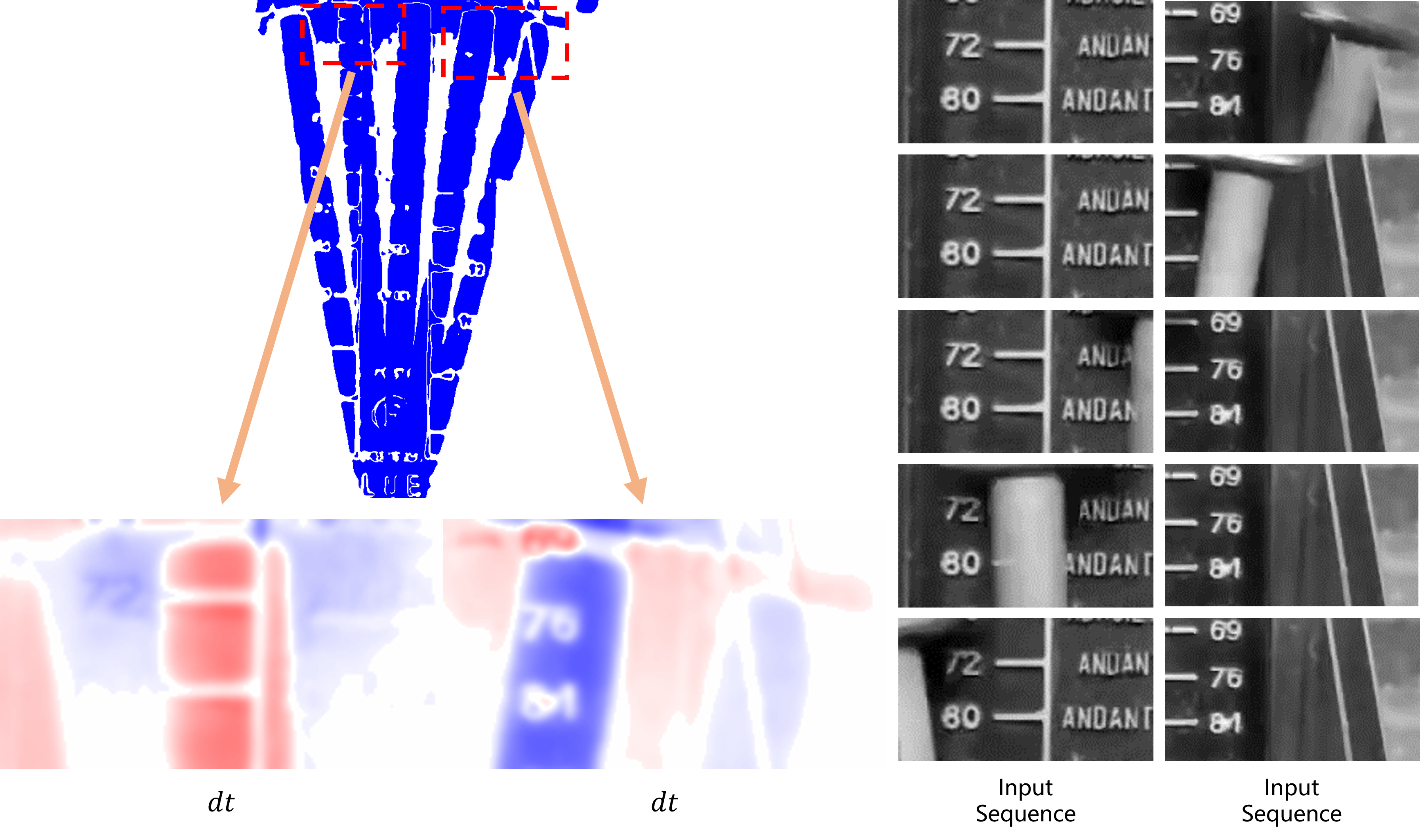}
    } 
    \subfigure[Static edges of moving objects and obscured objects]{  			 
        \includegraphics[width=\linewidth]{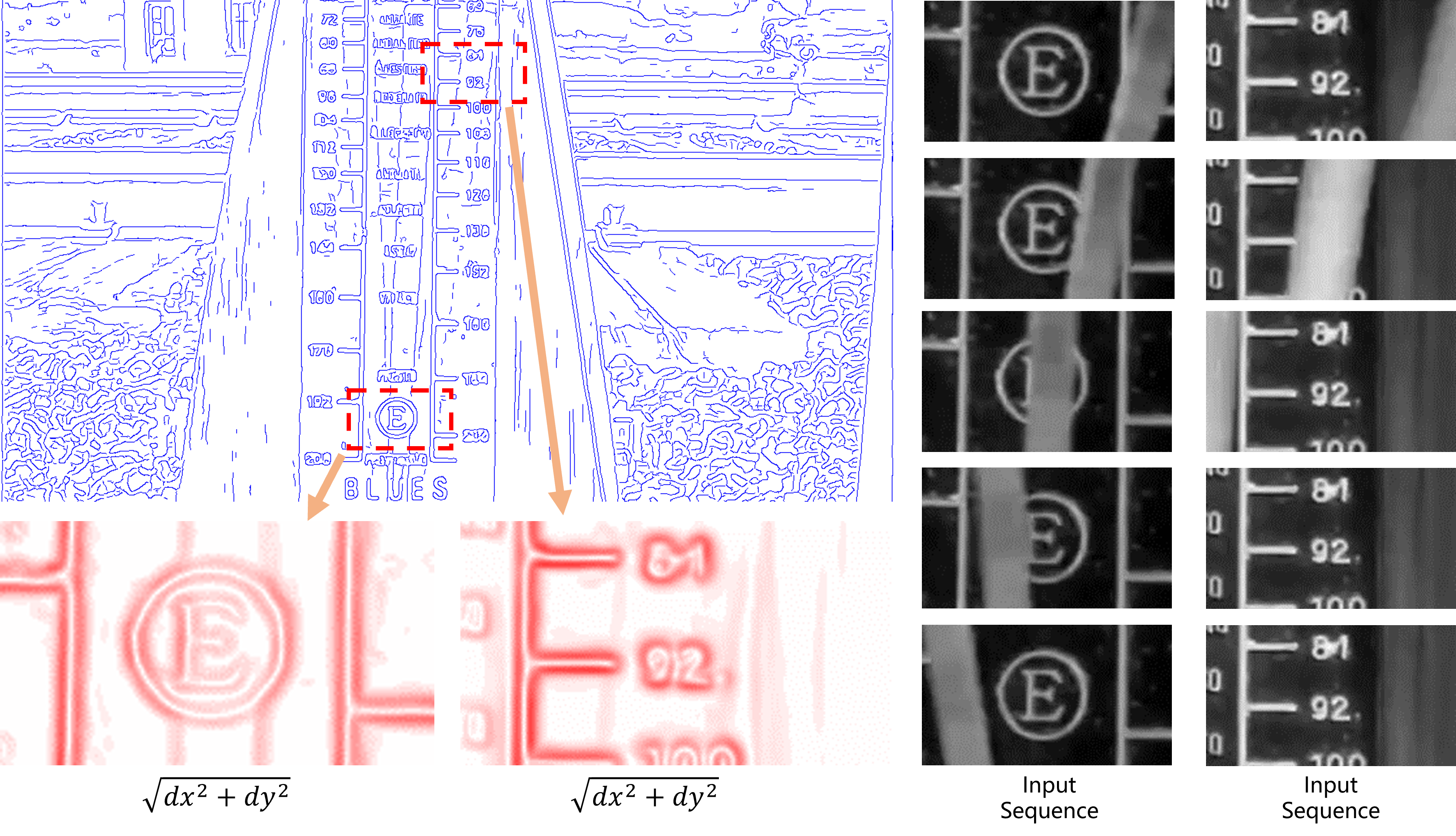}
    }
    \caption{Special case description of the detected edge.}  
    \label{fig:example3D-analysis} 
\end{figure}

We further designed a comparative experiment, which is a pendulum swinging before a clutter background under sunshine, for demonstrating the 3D edge detection in a complicate scenario (Figure~\ref{fig:example3D-pen_main}). The pendulum's swinging movement produces a periodic response on the time-direction TGD with the background transitioning from dim to light, then back to dim. Additionally, the periodic shadow produces a modification on the background color, transitioning on the saturation of the colors. As a result, the pendulum and the shadow engender opposing positive and negative responses, which are represented as red and blue in the TGD response. As observed in the local screenshots of input frames (second and third columns of Figure~\ref{fig:example3D-pen_main}.a), the movement of the pendulum produces a shadow on the background. This results in a variation of the brightness values at that particular spatial location over time, which is detected and marked as kinetic edges due to a variation in brightness over time ( Figure~\ref{fig:example3D-analysis}.a). 

Figure~\ref{fig:example3D-analysis}.b displays the static edges behind the moving object. The TGD responses in the $x$- and $y$-directions of the input frame sequence are identical to the weighted summation of the TGD responses of the individual frames. For moving objects that exist at different locations in different frames, their TGD responses at those locations result in a weak superposition effect on the final outcome. The further away from the intermediate frames, the weaker the superposition effect becomes. Likewise, for the background that is occluded and then reappears, its TGD responses are superimposed on the final response. This exposes itself in the identification of static edges (The letter “E”). It is conceivable that the TGD responses are accommodating to the occluded background and might be helpful in the future for detecting motion-obscured video objects and segmenting activities.

\begin{figure}[htbp]    	
    \centering    	
    \subfigure[Input Frame Sequence (T = 5)]{  			 
        \includegraphics[width=\linewidth]{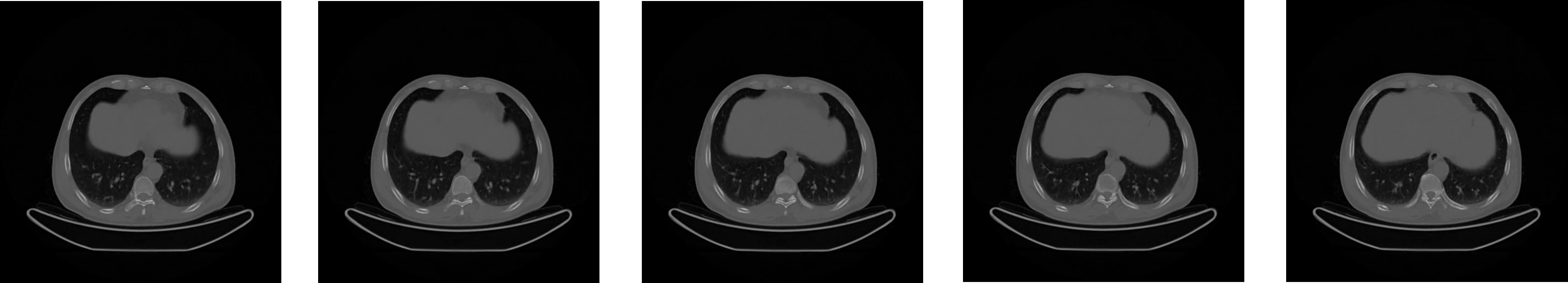}
    }
    \subfigure[$dx$]{  			 
        \includegraphics[width=0.23\linewidth]{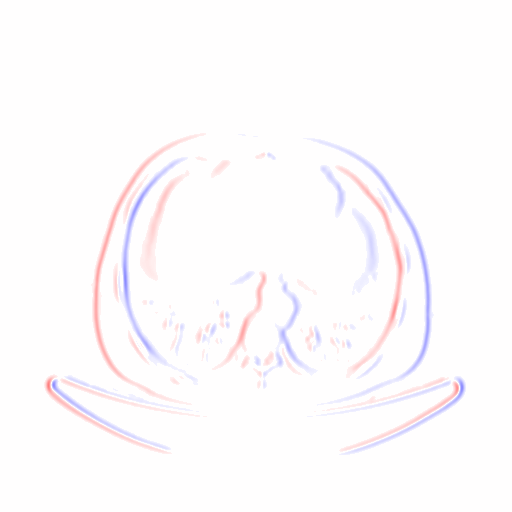}
    } 
    \subfigure[$dy$]{  			 
        \includegraphics[width=0.23\linewidth]{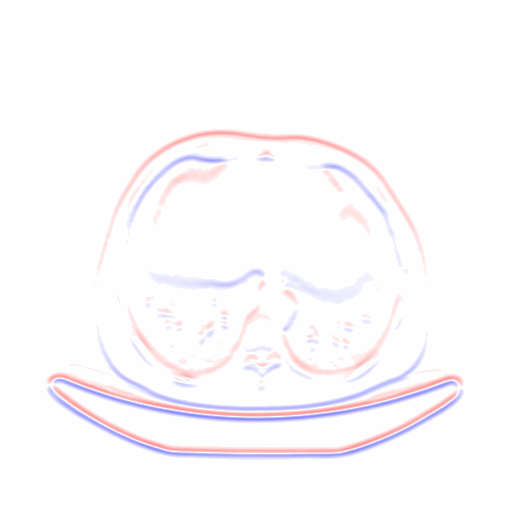}
    }
    \subfigure[$dz$]{  			 
        \includegraphics[width=0.23\linewidth]{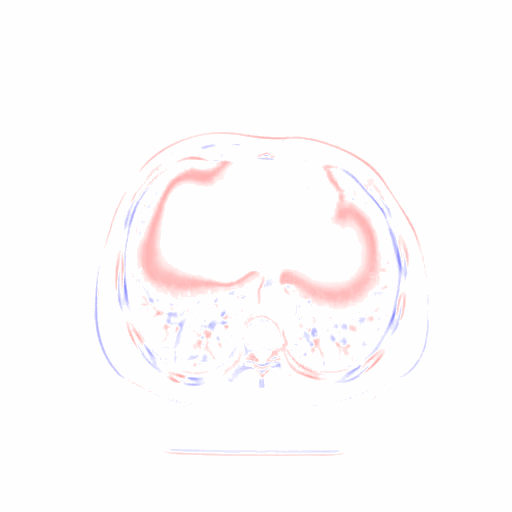}
    } 
    \subfigure[$d^2 z$]{  			 
        \includegraphics[width=0.23\linewidth]{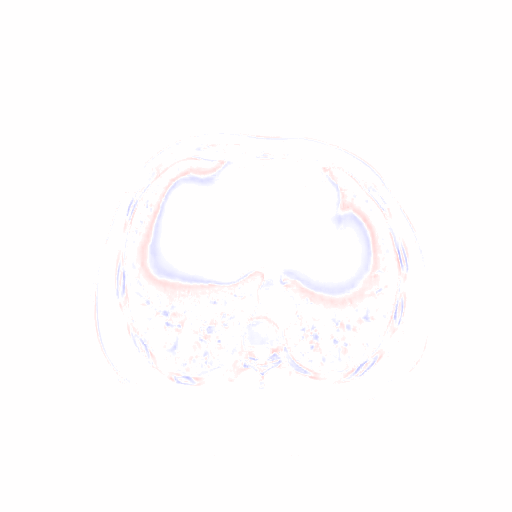}
    } 
    \subfigure[Static Edges]{  			 
        \includegraphics[width=0.31\linewidth]{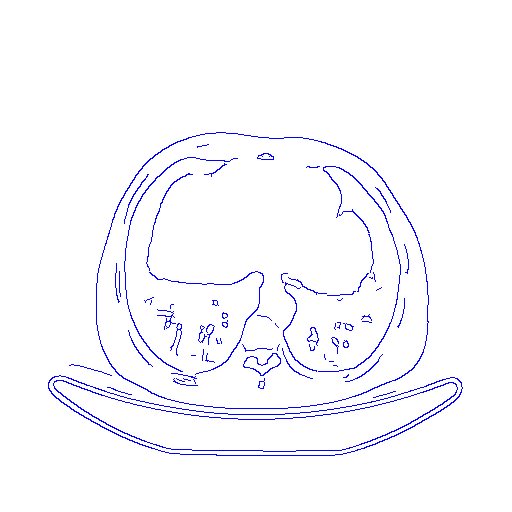}
    } 
    \subfigure[Kinetic Edges]{  			 
        \includegraphics[width=0.31\linewidth]{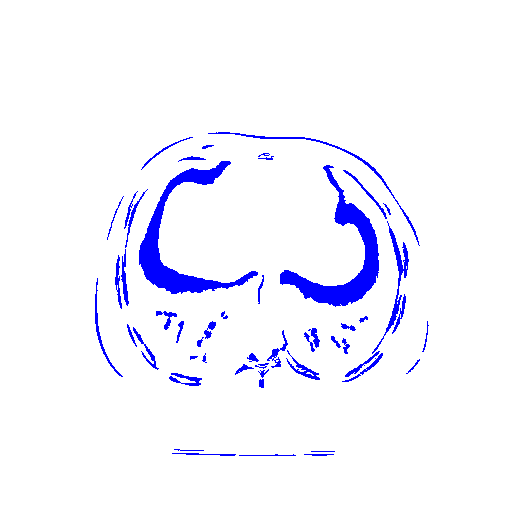}
    }
    \subfigure[Merge]{  			 
        \includegraphics[width=0.31\linewidth]{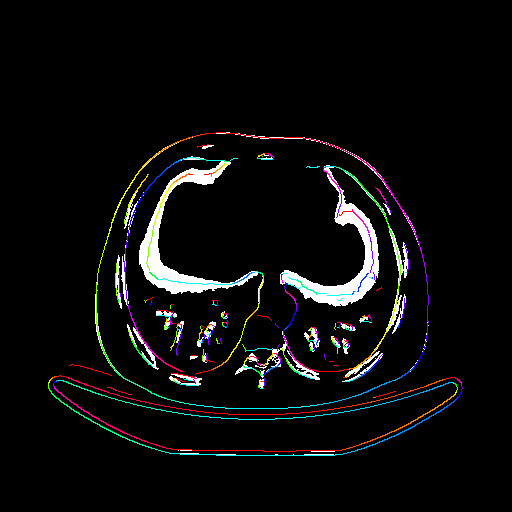}
    } 
    \caption{3D Edge detection results: (a) Input frame sequence, where T represents the number of effective frames; (b-e) first- and second-order TGD; (f) static edges;  (g) kinetic edges; (h) merged visualization via HSV space.}  
    \label{fig:example3D-CT} 
\end{figure}

Finally, we deploy our algorithm to the $xyz$ space rather than a video. Detection is performed on consecutive computerized tomography (CT) frames. Figure~\ref{fig:example3D-CT} demonstrates that the location as well as the changes in the organs are well detected. In the CT segment used for the experiment, the human exterior skin remains essentially unchanged, while the chest cavity volume changes (white area in Figure~\ref{fig:example3D-CT}.h). The larger the white area, the greater the range of change and the faster the rate of change. Moreover, both the thoracic and lumbar spine show slight variations along the $z$-axis. It may be possible to deploy TGD to track changes in normal patients and use this information to identify abnormalities. Characterizing changes in CT scans may provide potential benefits for medical diagnosis, which requires further collaborative research within cross disciplines.

\clearpage
\newpage

\section{Conclusion and Outlook}

We demonstrate the exceptional performance of TGD operators through signal processing experiments, especially operators constructed by the orthogonal method that combines good performance and fast computation. We introduce the concept of TGD-based smoothness to the discrete sequence, which is inherited from the derivative based smoothness of continuous functions. Building upon this smoothness principles of the first- and second-order TGD, we actively implement a single-layer one-dimensional convolution network and successfully reduce signal noise in the experiments. Likewise, two-dimensional TGD operators exhibit excellent performance in edge detection and noise suppression during image processing. Furthermore, we introduce the static and kinetic edges into three-dimensional TGD operators based 3D edge detection. 3D TGD operators are capable of performing spatio-temporal coordinated analysis and can serve as a pioneering method in the field of numerical analysis.

We anticipate that our colleagues in the fields of numerical analysis, signal processing, and computer vision will develop superior performing difference operators based on Tao General Difference, leading to greater advancement in numerical analysis.

\clearpage
\newpage

\bibliographystyle{unsrt}  
\bibliography{references}

\end{document}